%% file: main.tex
\documentclass[10pt,twocolumn,letterpaper]{article}

\usepackage[pagenumbers]{metafiles/cvpr} 
\usepackage{multirow}
\usepackage{makecell}

\definecolor{cvprblue}{rgb}{0.21,0.49,0.74}
\usepackage[pagebackref,breaklinks,colorlinks,allcolors=cvprblue]{hyperref}
\usepackage{utfsym}
\def\paperID{357}
\def\confName{3DV\xspace}
\def\confYear{2027\xspace}

\title{TV-SGS: Gaussian Splatting with Geometric Information Propagation\\ via Tensor Voting under sparse views}

\author{Harish N Sathishchandra\\
Stevens Institute of Technology\\
Hoboken, New Jersey, USA\\
{\tt\small hsathish@stevens.edu}\\
\and
Philippos Mordohai\\
Stevens Institute of Technology\\
Hoboken, New Jersey, USA\\
{\tt\small pmordoha@stevens.edu}\\
}

\begin{document}
\maketitle

\input{sec/abstract}    
\input{sec/introduction}
\input{sec/related}
\input{sec/method}

\input{sec/experiments}

\input{sec/conclusion}
{
    \small
    \bibliographystyle{ieeenat_fullname}
    \bibliography{bib/main, bib/all_nvs, bib/tv}
}
\clearpage
\maketitlesupplementary
\appendix
\setcounter{figure}{0}
\setcounter{table}{0}
\setcounter{equation}{0}
\renewcommand\thefigure{S.\arabic{figure}}  
\renewcommand\thetable{S.\arabic{table}}
\renewcommand\thesection{S.\arabic{section}}
\renewcommand\theequation{S.\arabic{equation}}

In this supplementary material, we present additional implementation details not included in the main paper in Section \ref{sec_supp:impl_details}.
Next, we provide additional quantitative and qualitative results, including the per-scene DTU Chamfer distances evaluated on the extracted mesh, geometric evaluations performed directly on splat centers on both datasets, as well as an evaluation using different initializations in Sections~\ref{sec_supp:add_quant_results} and \ref{sec_supp:add_qual_results}.

Finally, we provide quantitative and qualitative results on TV-SGS integration with additional state of the art \textit{dense-view} GS backbones such as PGSR, RaDe-GS and 2DGS, optimized on dense training views, in Section~\ref{sec_supp:dense_views}.

\input{sec_supp/implementation_details}
\input{figs_supp/dtu_small_large_overlap_example}
\input{sec_supp/add_quant_results}
\input{sec_supp/add_qual_results}
\clearpage
\input{sec_supp/dense_views}

\end{document}

%% file: sec/abstract.tex
\begin{abstract}
Gaussian Splatting has been effective in inferring scene representations that excel in novel view synthesis. Multiple splats cooperate seamlessly to synthesize the pixels of novel views and are jointly optimized even though they only affect each other indirectly, via pixels they project to in common. We present an approach that enables direct communication among splats to enhance the geometric structures they form in 3D. This is accomplished by Tensor Voting, which was originally designed to infer structures from noisy inputs and has been adapted here to provide supervision during test-time optimization, leading to more accurate scene geometry. We introduce a new class of 3D losses that do not rely on rendering and can be combined with essentially all losses previously reported in the literature. Our 3D losses are especially effective when the input views are sparse and geometric regularization is essential due to limited supervision from the images. Our method is easy to integrate with a diverse set of backbones, and our experiments on the DTU and Tanks-and-Temples datasets demonstrate that TV-SGS improves the geometry of the outputs compared to the backbone, while maintaining or improving rendering quality.  
\end{abstract}

%% file: sec/introduction.tex
\section{Introduction}\label{sec:intro}

For many years, 3D reconstruction was almost exclusively addressed by Multi-View Stereo (MVS) \cite{stathopoulou2023survey,wang2026learning_MVS_survey} which aims at generating high-fidelity geometric representations of the scene, based on some measure of photoconsistency. The computation of photoconsistency, however, is the only time MVS considers image appearance, since it often treats attaching texture-maps to the surfaces as a post-processing step, typically omitted from the papers. Counter-intuitively, despite the quality of the inferred geometry, visualizations generated from the output of MVS are typically unsatisfactory, even when appearance is modeled thoroughly~\cite{shan2013visualturing}. 

Breakthroughs in convincing view synthesis came a few years ago with the emergence of Neural Radiance Fields (NeRF) \cite{Mildenhall_2020_NeRF}, which utilize implicit representations of geometry and appearance, and a little later with the adoption of explicit representations, in the form of 3D Gaussian Splatting (3DGS) \cite{Kerbl_2023_3DGS}. Even though these methods attain substantially lower geometric accuracy than MVS, they are able to synthesize compelling images from novel viewpoints and have now become the leading paradigm in 3D reconstruction \cite{fei2024survey,tewari2022advances,xie2022neural}.\footnote{We will refer to both implicit and explicit flavors as Radiance Fields (RF), to the method of Kerbl et al. \cite{Kerbl_2023_3DGS} as 3DGS and to Gaussian splatting methods in general as GS.}

\input{figs/tnt_compare_barn}

Several authors have attempted to close the gap between MVS and RF \cite{Charatan_2024_pixelSplat,chen_2024_mvsplat,chen2021mvsnerf,liu2024mvsgaussian,wu2025sparse2dgs} and strike a better balance between geometry and appearance, but there still exists a dichotomy between methods that generate geometrically precise 3D representations and those that achieve compelling view synthesis. Our work aims to bridge the gap adopting an explicit RF representation and improving its geometric fidelity while at least maintaining the quality of its renderings.

The NeRF and GS literature has introduced a variety of losses, almost all of which are viewpoint-dependent; they are computed by contrasting rendered predictions, of color, depth, surface normal etc., made by the RF and some reference quantity, such as image colors or surface normals, available or rendered on the image plane.
A consequence of these formulations is that splats influence the evolution of each other only when they overlap during rendering. They cooperate via $\alpha$-blending to synthesize part of an image, depthmap or normal map, and receive gradients from the corresponding loss via backpropagation. 

In this paper, we take a different approach and present \textbf{TV-SGS}, the first GS framework that supports geometric information propagation among primitives in 3D. This is accomplished by a new formulation of \textit{Tensor Voting (TV)} \cite{gerard_book} designed to be compatible with Gaussian splats. Tensor Voting enables us to estimate the tangent and normal subspaces at the centroid of each Gaussian splat and to progressively refine the inferred structures.
Our most important technical contribution, in terms of GS, is \textbf{a new class of 3D losses} that operate on the Gaussian primitives without requiring rendering. 
We also introduce a method for inferring the most likely surface point along a given direction (the normal of a splat).
These surface points then provide supervision in a loss that encourages the splats to move towards the most likely surfaces. The end result is a scene representation with improved geometry (see Fig.~\ref{fig:tnt_compare_barn}) that maintains or improves the novel view synthesis capabilities of the backbone without the new losses. 

Tensor Voting is well suited for explicit representations with irregular density, like splats, for several reasons. First, all operations take place in local neighborhoods, and thus can be deployed on the GPU. Second, it is robust to outliers. Third, with our proposed per-primitive scale selection and neighborhood whitening, it can adapt to data of varying density. Fourth, it operates in 3D and affects all splats in the same optimization step, as opposed to standard GS processing in which only splats in a reference image with non-zero transmittance are optimized at each step. 

The effects of our new 3D losses are more pronounced under sparse-view settings, where the task of inferring accurate scene representations is hindered due to limited supervision from a small set of images. Under these conditions, TV-SGS is able to enforce constraints such as surface smoothness on the primitives directly in 3D without going through a rendering process.
TV-SGS is flexible; we have successfully integrated it with three backbones,
PGSR \cite{Chen_2024_PGSR}, FatesGS \cite {huang2025fatesgs} and VGGS \cite{xiang2026vggs},  and tested it in sparse-view settings, as well as additional backbones in dense-view settings (shown in the Supplement). The effectiveness of TV-SGS combined with a diverse set of backbones demonstrates that our approach can enhance various types of GS optimization: PGSR which uses MVS-based supervision and is designed for dense views without using any pre-trained priors, FatesGS \cite{huang2025fatesgs} and VGGS \cite{xiang2026vggs} both of which are designed for sparse views and use monocular and multi-view pre-trained depth priors respectively.
In summary, our contributions are:
\begin{itemize}
    \item A novel Gaussian splatting framework for sparse views with direct propagation of geometric information from primitive to primitive.
    \item Two novel 3D losses, namely the TV normal loss and TV position loss, to regularize the positions and orientations of the splats directly in 3D.
    \item A formulation of Tensor Voting that complements Gaussian splatting by encoding, propagating and aggregating geometric structure information. Our formulation includes a novel technique using virtual receivers to efficiently infer the most likely surface location along a 3D line segment, as well as a technique for whitening the voting neighborhoods.
    \item Integration of TV-SGS with several different backbones showing the effectiveness and flexibility of TV-SGS.
    \item An open-source implementation to be released upon acceptance.
\end{itemize}

%% file: figs/tnt_compare_barn.tex
\newcommand\cellwt{0.32\linewidth} 
\newcommand\hgapt{0.5pt}             
\newcommand\vgapt{0.5pt}             
\newcommand\rotvert{0.85cm}
\begin{figure}[t!]
\centering
\setlength{\tabcolsep}{\hgapt}

\begin{tabular}{@{}lccc@{}}
{} & PGSR & VGGS & FatesGS \\
\rotatebox{90}{Backbones} & 
\includegraphics[width=\cellwt]{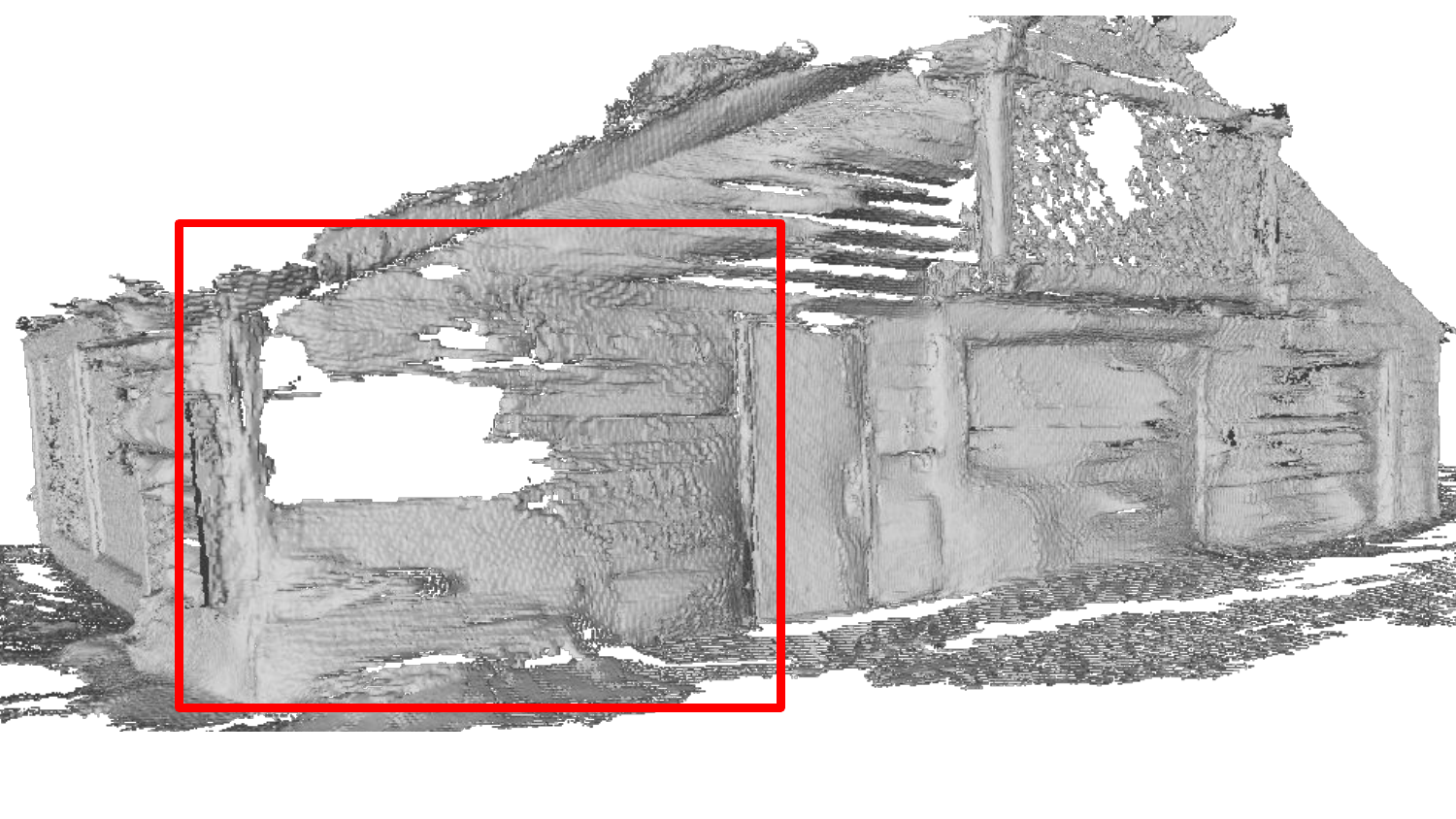} &
\includegraphics[width=\cellwt]{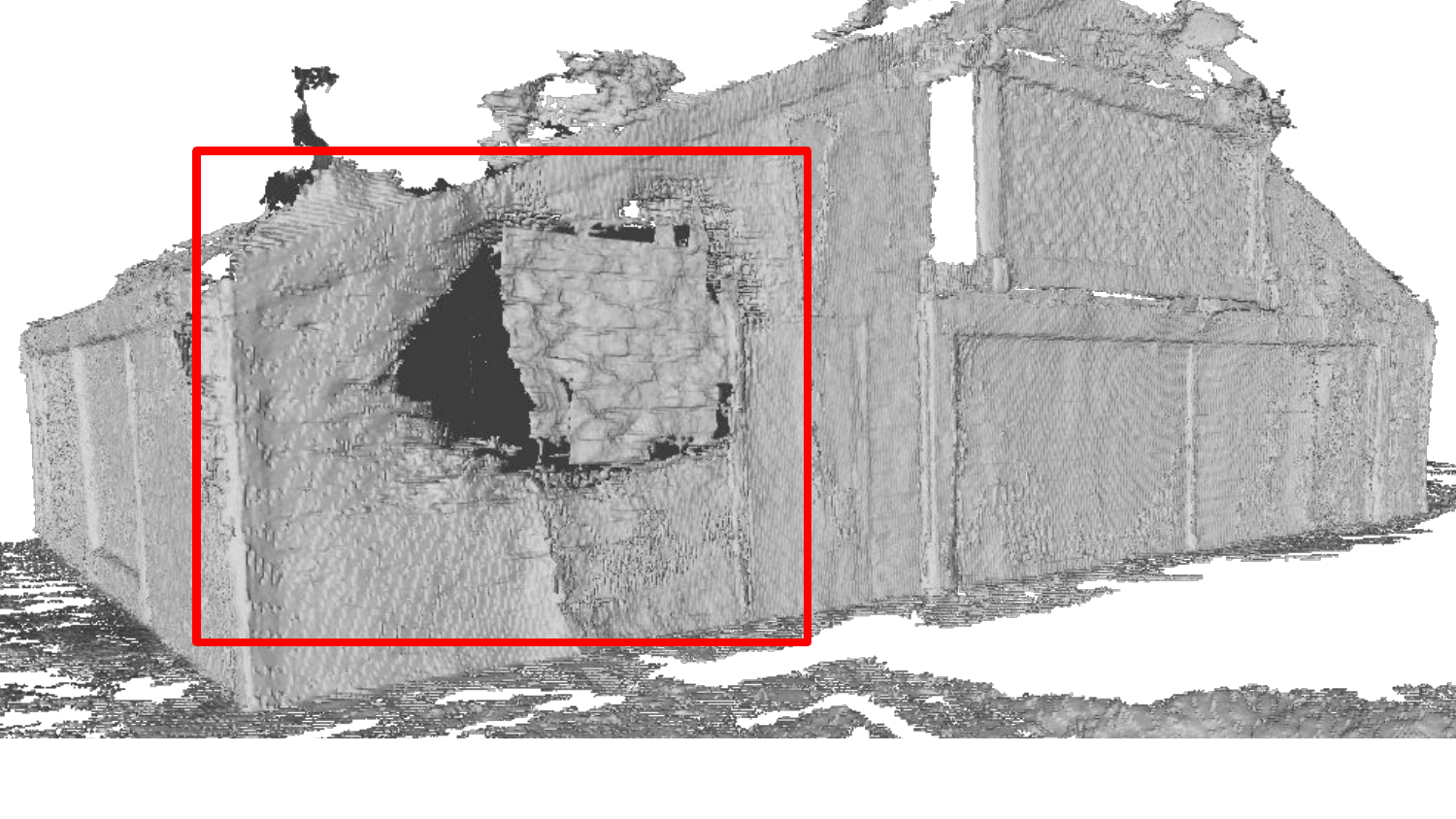} &
\includegraphics[width=\cellwt]{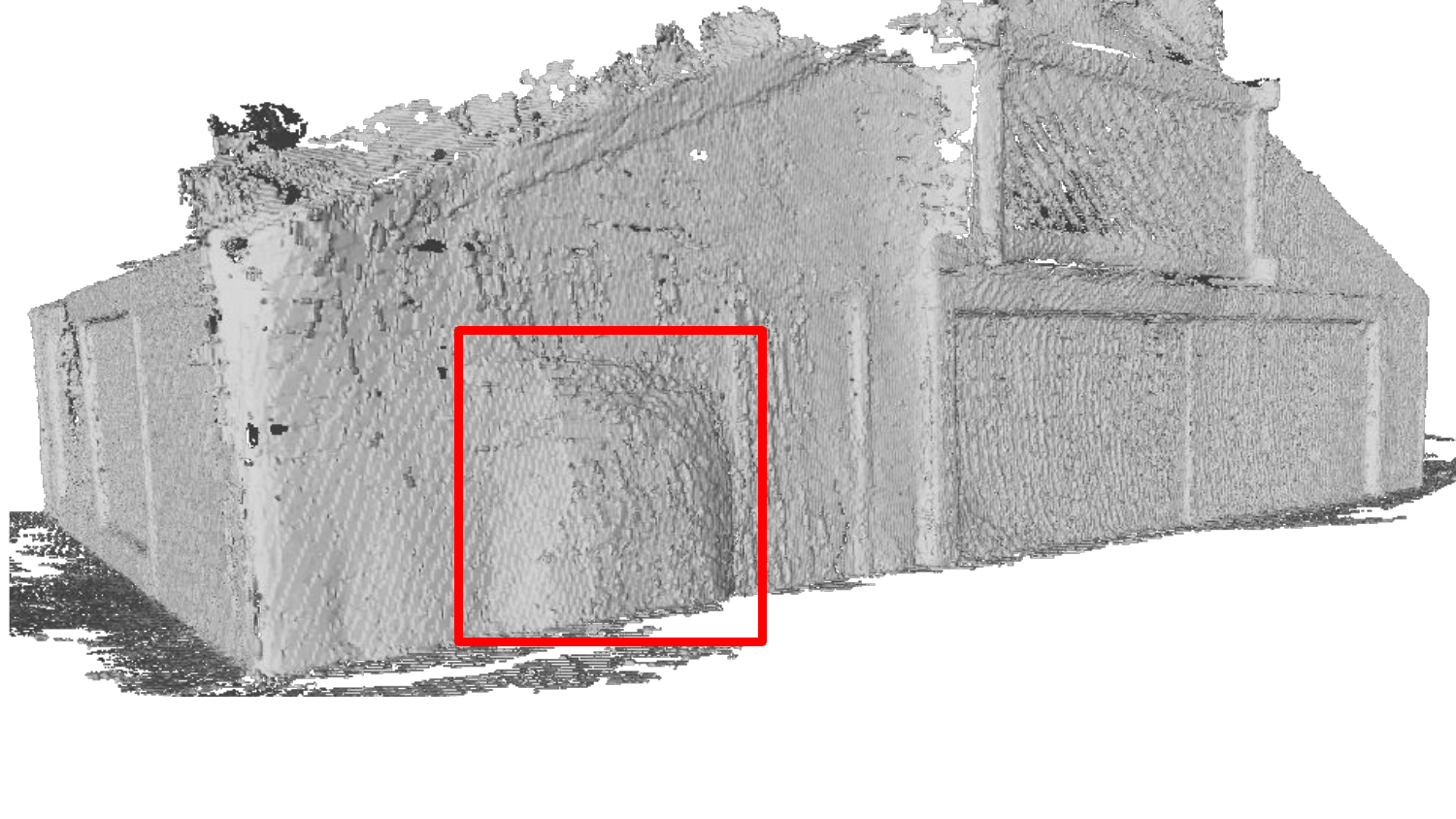} \\[\vgapt]
\rotatebox{90}{TV-SGS} & 
\includegraphics[width=\cellwt]{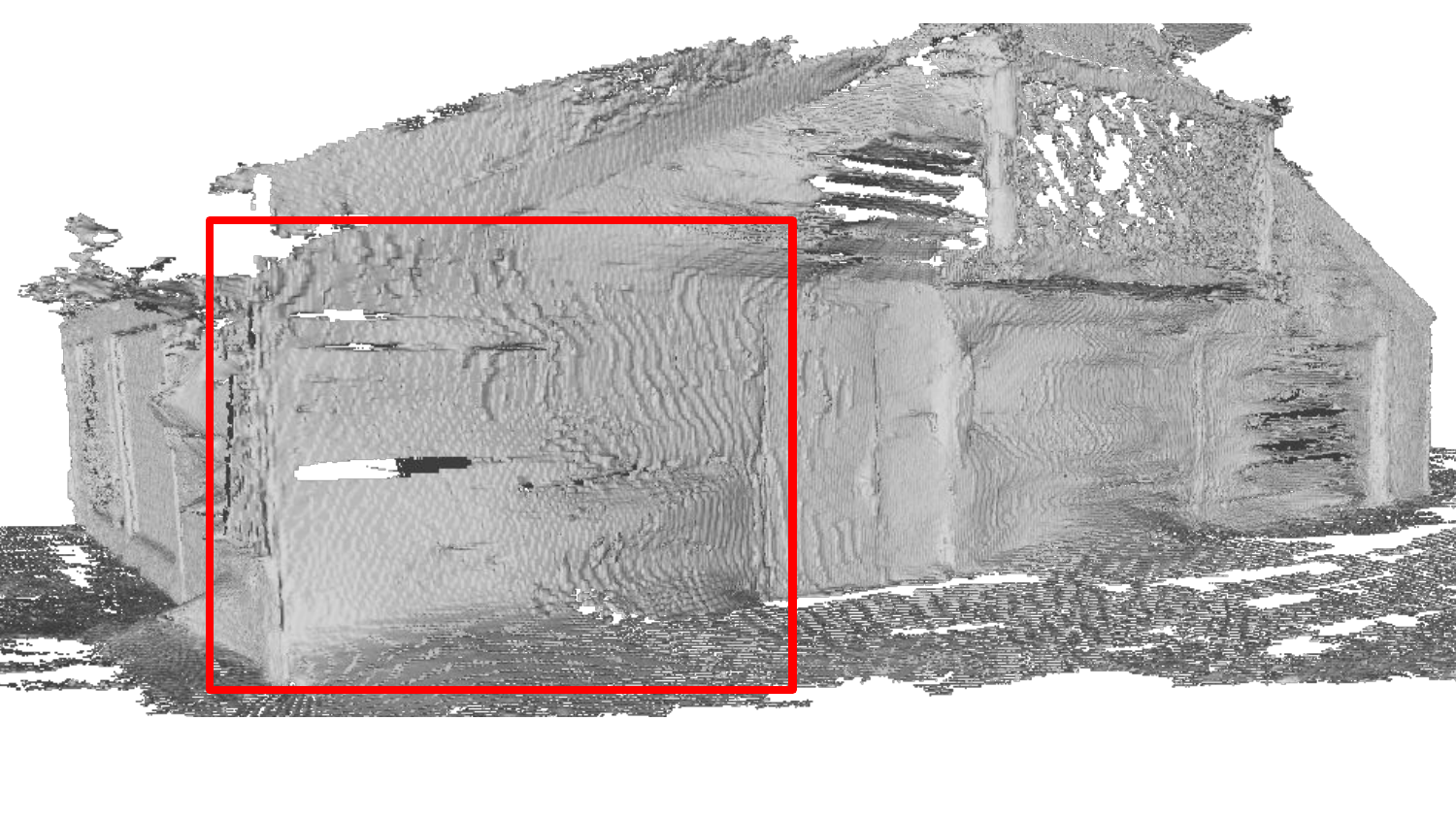} &
\includegraphics[width=\cellwt]{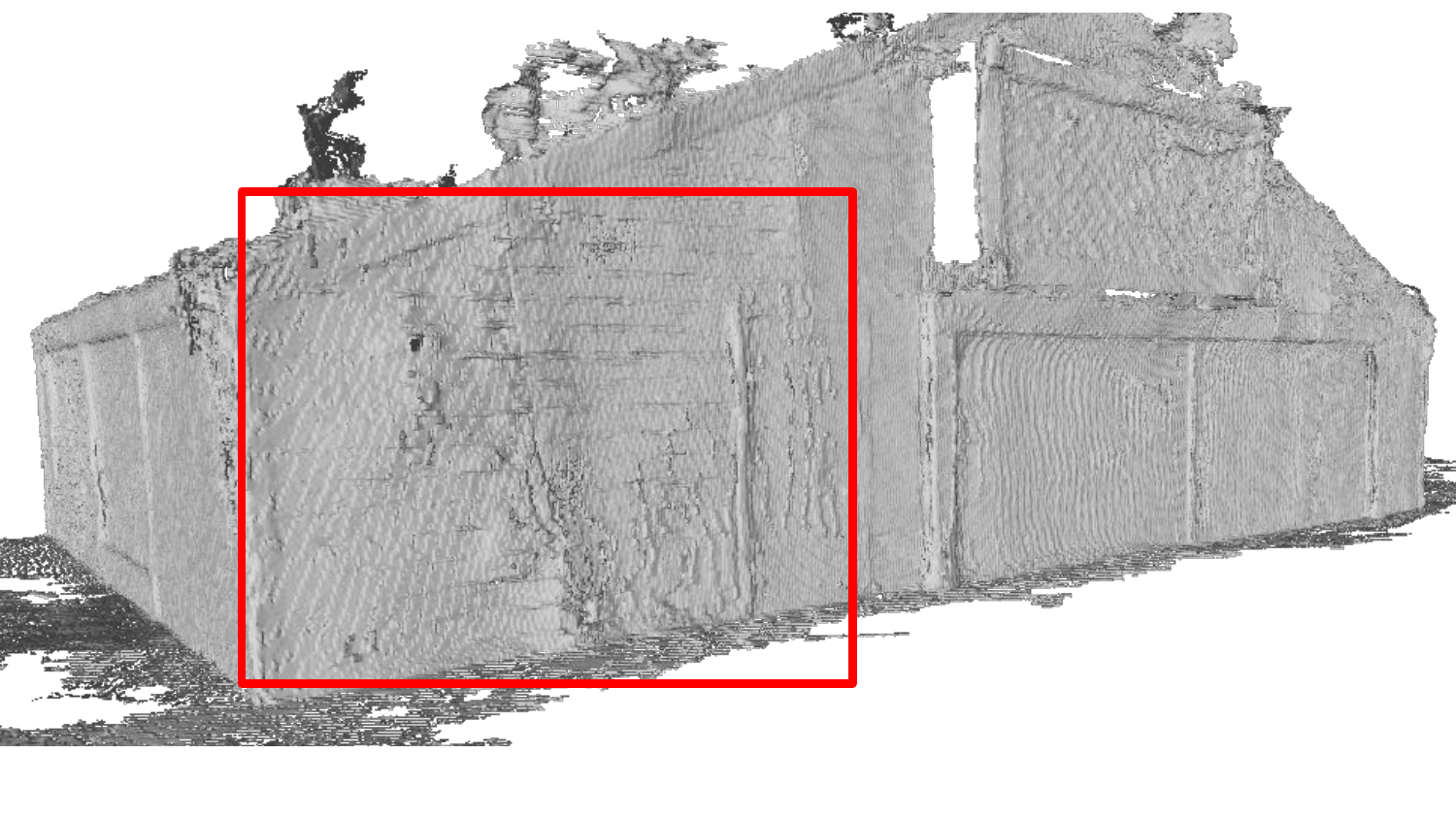} &
\includegraphics[width=\cellwt]{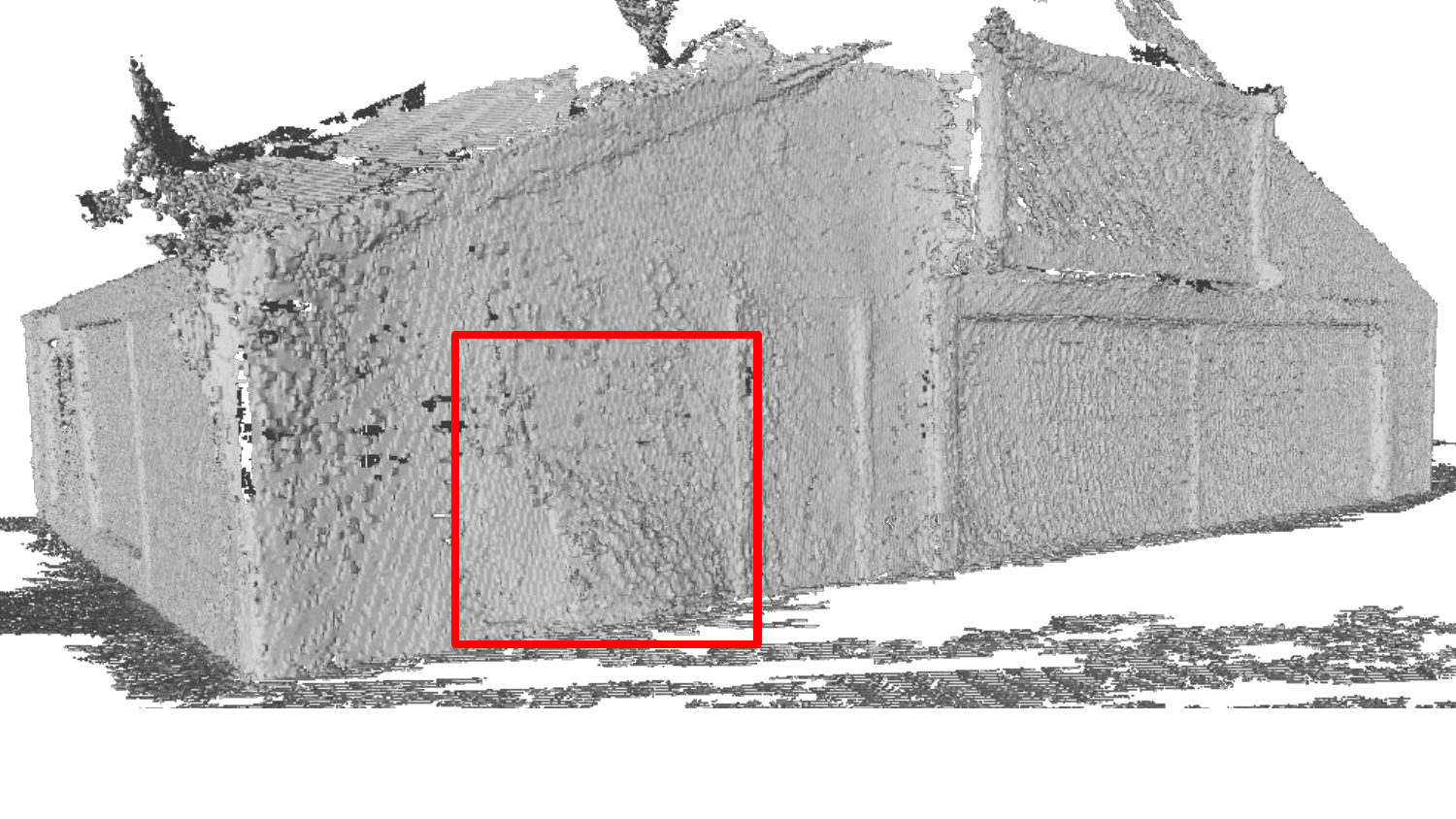}
\end{tabular}

\caption{Qualitative comparison of surface reconstruction on the Tanks \& Temples dataset \cite{Knapitsch_2017_Tanks} between the PGSR, VGGS and FatesGS backbones before and after integration with TV-SGS. Top: results by the backbones. Bottom: results by the respective TV-SGS implementations.}
\label{fig:tnt_compare_barn}
\end{figure}

%% file: sec/related.tex
\section{Related Work} \label{sec:related}

In this section, we review GS approaches that emphasize geometric accuracy in addition to novel view synthesis. We refer readers to surveys for a broader perspective on GS \citep{bao_2025_3d,dalal_2024_gaussian,luo_2024_review,wu_2024_recent}. While volumetric rendering, implicit \cite{Mildenhall_2020_NeRF} or explicit \citep{Kerbl_2023_3DGS}, is not limited to solid surfaces, true volumetric elements of a scene, such as smoke or fog, are typically transient. The permanent elements, for which geometric accuracy is important, are surfaces. 
Therefore, this section focuses on surface geometry in the context of GS, and sparse-view GS in particular.

Several authors have modified the representation to emphasize surface reconstruction. 
2D Gaussian Splatting (2DGS) \citep{Huang_2024_2DGS} converts the 3D Gaussian ellipsoids to flat 2D ellipses by setting their minimum eigenvalue to 0. 
SuGaR \citep{Guedon_2024_SuGaR} favors opaque and flat 3D Gaussians regularized by a loss enforcing signed distance function (SDF) consistency. 
Gaussian Opacity Fields (GOF) \citep{Yu_2024_Gaussian} extract surfaces directly as a level set of the opacity in 3D.
RaDe-GS \cite{Zhang_2026_RaDe} shows that under the local affine approximation the ray-Gaussian intersections of a pixel bundle are coplanar, thus giving a closed form planar depth and a per splat normal that can be rasterized efficiently.
PGSR \citep{Chen_2024_PGSR} relies on a planar representation and a multi-view loss that penalizes splat photo-inconsistencies on pairs of images.
CarGS~\cite{shen2025evolving} is similar but focused on efficient rendering.
3D-Half-Gaussian Splatting (3D-HGS) \citep{li_2024_3d} enables the representation of planar surfaces and hard edges by attaching a splitting plane to each Gaussian,
that separates free from occupied space.
GeoGaussian \cite{li2024geogaussian} initializes thin Gaussians aligned with smooth surfaces inferred from the input SfM point cloud, and ensures Gaussians generated during densification remain on these surfaces.

Other approaches integrate additional geometric cues. Monocular surface normal estimates by pre-trained networks are used in Gaussian Surfels \citep{dai_2024_high}, which also segments the foreground from the background, VCR-GauS \cite{chen_2024_vcr}, which integrates an uncertainty-aware normal regularizer, and GSrec \cite{wu2024surface}, which estimates the SDF locally.
StruGS \cite{Pang_2025_StruGS} considers multi-view guidance to focus on structurally salient regions and remove floaters.

All of the previously mentioned methods are designed for dense view coverage and usually suffer under sparse input views, producing noisy and incomplete surfaces. Although there has been significant interest in GS under sparse views constraint, most of the approaches focus on novel view synthesis ignoring geometry \cite{chung2024depth, kumar2024few, li2024dngaussian, park2025dropgaussian, xu2025dropoutgs, zhu2024fsgs, chen_2024_mvsplat, han2024binocular, zhang2024cor}. Among the few works that focus on geometry, the most prominent ones are FatesGS \cite{huang2025fatesgs}, which uses monocular depthmaps to enforce consistent depth ranking within local image patches, along with cross view feature consistency for absolute depths. VGGS \cite{xiang2026vggs} uses multi-view consistent anchor pixels to align depthmap priors to the underlying surface and then propagates the depths from anchors to areas with unreliable depth estimates. Both these methods, however, rely on rendered image space supervision, which is limited  due both to the small number of images and to the small overlaps among them. On the other hand, Sparse2DGS \cite{wu2025sparse2dgs} relies on 
MVS-derived geometric features and on directly optimizing points sampled from the 2D Gaussian disks.
However, these sampled 3D points are still projected into image space for supervision based on cross-view image feature matching, which is poorly constrained under the sparse-view setting.

Similar to TV-SGS, FeatureGS \cite{jager2025featuregs} investigates the use of features derived from the scatter matrix of the neighborhood of each splat as additional losses in 3DGS. Properties such as planarity, `omnivariance’ and `eigenentropy’ are tested but the results are not competitive yet. In our terminology, FeatureGS encodes only colinearity without considering proximity.

%% file: sec/method.tex
\section{Method}
\label{sec:method}

\input{figs/overview}

In this section, we present our GS framework (TV-SGS), which infers structural information and enforces constraints to the primitives in 3D.
TV-SGS enables direct communication between the 3D splats via a voting process that recovers the underlying geometric structures they represent. See Fig.~\ref{fig:overview} for an overview.

\subsection{Representation}
\label{subsec:rep}
Like most Gaussian splatting methods, we represent the scene by a set of Gaussian splats. 
Each splat $g _i\in \mathcal{G}$ encompasses: 
\begin{itemize}
    \item A center (mean) $\boldsymbol{\mu} \in \mathbb{R}^3$.
    \item A covariance matrix $\boldsymbol{\Sigma} \in \mathbb{R}^{3 \times 3}$, represented by a rotation matrix $\boldsymbol{R} \in SO(3)$ and a scaling matrix $\boldsymbol{S} \in \mathbb{R}^{3 \times 3}$. The rotation matrix is implemented as a quaternion $\boldsymbol{q} \in \mathbb{R}^4$ for compactness.
    \item An opacity value $\alpha$.
    \item 48 spherical harmonics (SH) coefficients for representing color.
    \item A second-order tensor $\boldsymbol{T} \in \mathbb{R}^{3 \times 3}$ representing the aggregated influence from neighboring splats onto splat $g_i$. This can be further decomposed into: eigenvalues $\boldsymbol{\lambda_i,}$ with $i \in [1, 3]$, and a matrix of eigenvectors $\boldsymbol{E} \in SO(3)$ which serves as a local coordinate system.
    \item The predicted Gaussian center $\boldsymbol{\mu'} \in \mathbb{R}^3$ with maximum surface saliency after Tensor Voting.
\end{itemize}

Note that both $\boldsymbol{\Sigma}$ and $\bf{T}$ are positive semi-definite symmetric matrices. When they represent a local surface, their respective eigenvectors corresponding to the minimum eigenvalue represent the local surface normal. Their other two eigenvectors represent the tangents to the local surface. The eigenvalues, however, represent different properties of the neighborhood. The eigenvalues of $\boldsymbol{\Sigma}$ are proportional to the sparsity of the neighborhood along the corresponding eigenvector, while the eigenvalues of $\bf{T}$ are proportional to the support a splat received from its neighbors along that direction. In other words, the eigenvalues in the tangent directions represent sparsity in $\boldsymbol{\Sigma}$ and density in $\bf{T}$. (See also Section~\ref{sec:conclusion}.)

\subsection{Tensor Voting}
\label{subsec:tv}
This section describes how geometric information of each splat can be encoded and propagated to other splats within its neighborhood using Tensor Voting.
The use of a voting process for structure inference from sparse and noisy data was presented in \cite{gerard_book}. Note that in the work of Medioni et al. \cite{gerard_book,mordohai2006tensor} voting is performed by the normals of the underlying geometric structures. Here, to align the tensors used for voting and the covariance matrices of the splats, we modify the Tensor Voting framework so that voting is done by \textit{tangents of the underlying geometric structures}.

Tensor Voting (TV) in TV-SGS is applied in 3D, with primitives represented by 3D second-order, symmetric, positive semi-definite tensors. Information about the type of geometric structure, surface, curve or junction, is encoded in the eigenvalues of the tensor, and information about its orientation in the eigenvectors. Eigenvalue gaps (differences between adjacent eigenvalues) represent the \textit{saliency} of the corresponding structure type, with the eigenvectors representing the local tangent and normal subspaces. For example, an ideal surface has a tensor with one non-zero eigenvalue gap between the second and third eigenvalue, a 2D tangent subspace spanned by the first two eigenvectors and a surface normal represented by the third eigenvector.
Tensor voting propagates each point's preferred orientations within its neighborhood subject to constraints such as proximity, co-linearity and co-curvilinearity. The scale of voting, $\tau$, determines the attenuation of vote strength with distance and curvature. (See~(\ref{eq:decay_func})).

A symmetric, positive semi-definite tensor can be decomposed as:
\begin{align}
\textbf{T} &= \sum_i \lambda_i \mathbf{\hat{e}_i\hat{e}^T_i} \nonumber \\
&= (\lambda_1 - \lambda_2)\mathbf{\hat{e}_1\hat{e}^T_1} \nonumber \\ 
 &+ (\lambda_2 - \lambda_3)(\mathbf{\hat{e}_1\hat{e}^T_1} + \mathbf{\hat{e}_2\hat{e}^T_2}) \nonumber \\ 
 &+ \lambda_3(\mathbf{\hat{e}_1\hat{e}^T_1} + \mathbf{\hat{e}_2\hat{e}^T_2} + \mathbf{\hat{e}_3\hat{e}^T_3}) 
\label{eq:tensor_decomp}
\end{align}
The first term in (\ref{eq:tensor_decomp}) corresponds to an elongated ellipsoid, defined as the \textit{stick tensor}, which indicates a curve (or a surface intersection) with $\mathbf{\hat{e}_1}$ as its tangent and $(\lambda_1 - \lambda_2)$ as its curve saliency. 
The second term corresponds to a disk shaped ellipsoid, defined as the \textit{plate tensor} that indicates an elementary surface with $\mathbf{\hat{e}_3}$ as its normal  and $(\lambda_2 - \lambda_3)$ as its surface saliency. The third term corresponds to a sphere, defined as the \textit{ball tensor}, a structure that has no preference in orientation, such as a junction or curve intersection, with $\lambda_3$ as its saliency.

The fundamental operation in Tensor Voting is casting a vote from a voter to a receiver within the voter's neighborhood. The vote has the orientation the receiver would have if both the voter and the receiver belonged to the same perceptual structure. \textit{The receiver's tensor does not affect the vote.} The voter is first decomposed into its elementary tensors according to (\ref{eq:tensor_decomp}) before voting. Since our focus is on surfaces, only the plate components cast votes. We have found experimentally that the other votes do not affect the output meaningfully. The exception is the first iteration, in which ball votes are cast by isotropic tensors (whose only non-zero component is the ball.)

We still begin the description with the vote cast by a stick tensor, since the other votes can be derived from it.
The stick tensor propagates preference for a smooth continuation to the receiver in the form of a circular arc passing through the receiver and the voter with the stick as its tangent (or a linear segment, if the receiver is collinear with the voter's stick tensor). Therefore, stick voting occurs on a plane defined by the stick tensor, as the x-axis, and the vector $\mathbf{d}$ connecting the voter and receiver. (If they are collinear, any plane containing them is valid.) For convenience, we set the xz-plane as the voting plane.
The strength of this vote, $V$, decays with respect to the length and curvature of the connecting arc, encoding a preference for proximity and smoothness, as shown below:
\begin{align}
    &V(\boldsymbol{d},\theta)  = e^{-(\frac{s^2 +
     c\kappa^2}{\tau^2})}, \; 
     \text{where }
      s =\frac{\theta ||\boldsymbol{d}||}{\sin(\theta)}, \;  \kappa =\frac{2\sin(\theta)}{||\boldsymbol{d}||}
    \label{eq:decay_func}
\end{align} 
where, $\mathbf{d}$ is the vector connecting voter and receiver, $\theta$ is the angle between $\mathbf{d}$ and the voting stick, $s$ is the length of the circular arc that is tangent to the voter and passes through the receiver, $\kappa$ is its curvature, and $c$ is constant shown in the Supplement. The vote itself is a stick tensor with the two smallest eigenvalues equal to 0, its largest eigenvalue equal to $(\lambda_1-\lambda_2)V(\boldsymbol{d},\theta)$ and the corresponding eigenvector specified in (\ref{eq:oriented_votes}) and Fig. \ref{fig:oriented_votes}(left). 
\begin{align}
\textbf{S}(\boldsymbol{d},\theta) & = \left[
\begin{array}{c}
  cos(2\theta) \\
  0 \\
  sin(2\theta) \\
\end{array}
\right][cos(2\theta) \hspace{0.1in} 0 \hspace{0.1in} sin(2\theta)]
\label{eq:oriented_votes}
\end{align}

Plate and ball tensors can be thought of as having multiple tangent orientations, spanning 2D or 3D  subspaces, respectively. A plate, therefore, represents a surface characterized by tangents that span a 2D subspace and are all orthogonal to the surface normal. 
In the original TV framework, plate and ball tensors voted by integrating stick votes that spanned the voting subspace (normal in original TV, tangent here) followed by normalization. 
The resulting votes had a dominant stick component consistent with the vector connecting the voter and receiver as well as non-zero saliency in their ball and plate components. 
Aggregating such votes can give rise to any structure type at the receiver. Here, we accelerate voting from plate and ball tensors by \emph{casting the vote with the maximum strength} from their tangent space, bypassing casting and integrating multiple votes.

\input{figs/stick_tensor}

\subsection{Whitening of Voting Neighborhoods}
\label{subsec:aniso_nei}

So far, we have presented a generic Tensor Voting implementation which assumes that points are sampled from the underlying structures randomly. This, however, is not true in Gaussian splatting where splats cover the surfaces non-uniformly, guided by view synthesis criteria. The tangent planes of splats may become anisotropic guided by the frequency content of the images. For example, in an image region with stripes, splats would become elongated along the stripes, and thus would be spaced further apart along that direction and more densely along the orthogonal direction that exhibits higher texture frequencies. Applying Tensor Voting on primitives placed this way would lead to artifacts because the denser direction would appear as more salient after voting.

To mitigate this effect, we apply a whitening transformation to the space around each voter. The distortion we wish to undo is precisely captured by the voter's covariance matrix $\boldsymbol{\Sigma_i}$. Therefore, we use the Mahalanobis distance according to the voter's $\boldsymbol{\Sigma_i}$ to identify its nearest neighbors in the whitened space. The distance between voter $i$ and receiver $j$ is:
\begin{align}
\left\Vert \bf{d_{ij}^{\prime}} \right\Vert= \sqrt{\boldsymbol{d}_{ij}^T\boldsymbol{\Sigma}_{i}^{-1}\boldsymbol{d_{ij}}}
\label{eq:maha_d}
\end{align}
where $\boldsymbol{d_{ij}}$ is the vector from voter to receiver. As a result, a voter that does not belong to the structure represented by the receiver will appear to be far away in the whitened space even if it is close in the world space.  
Voting is performed as above, after transforming $\mathbf{d}_{ij}$ and the voter's tangent, $\mathbf{\hat{t}_{i}}$, to the whitened space.
\begin{align}
&\bf{d_{ij}^{\prime}} = \bf{M}_i\bf{d_{ij}} \text{, where } \bf{M}_i = \bf{\Sigma}_{i}^{-\frac{1}{2}} \nonumber \\ 
&\mathbf{\hat{t}_{i}^\prime} = \mathbf{M_i}\mathbf{\hat{t}_{i}} \nonumber \\ 
&\bf{S}_{ij} = \mathbf{M_i^{-1}}\mathbf{S_{ij}^{\prime}}\mathbf{M_i^{-T}}
\label{eq:vote_maha_transform}
\end{align}
where $\mathbf{S_{ij}^{\prime}}$ is the vote in the whitened space and $\mathbf{S_{ij}}$ is the vote in the original space. Votes from all voters can be aggregated at the receiver after being transformed to the original world coordinate system.

\subsection{Tensor Voting within Gaussian Splatting}
\label{subsec:TV-SGS}

Tensor Voting is applied to infer geometric information at each Gaussian primitive in two ways: first by estimating the tangent and normal subspaces at the center of each splat, and second by estimating the most likely surface position along the normal of each splat (see Section~\ref{subsec:virtual_receivers}). Tensor voting is treated as a non-differentiable module and the estimates it generates are used as constraints in the loss. 

We start the voting process with a round of ball voting from the tensors that are initialized with three equal eigenvalues. Since we are interested in surfaces, we only cast plate votes and ignore the other components of the tensor in all subsequent voting rounds after initialization. Experimentally, we found that casting stick and ball votes did not contribute much to surface estimation.

The voting process is applied on the centers of the splats, $\boldsymbol{\mu}_i$, which act both as voters and receivers. This entails the following steps:
\begin{enumerate}[label=\arabic*., leftmargin=*, labelindent=\parindent]
    \item Find the $2k$ nearest neighbors of $\boldsymbol{\mu}_i$ using Euclidean distances and binning on the GPU.
    \item Determine the scale of voting, $\tau_i$, for each voter by taking the median distance from its $k$ nearest neighbors after whitening. (See Section \ref{sec:Exp}.)
    \item Cast plate votes from each voter (all $\boldsymbol{\mu}_i$) to all receivers in its neighborhood and accumulate votes at the receivers via tensor addition. 
    \item Apply eigen-decomposition to the accumulated votes to obtain surface tangents and saliencies.
\end{enumerate}

The estimates of normals and positions generated by Tensor Voting provide a means for view-independent regularization of the geometry that leads to direct gradients to each splat, as opposed to view-based optimization that leads to gradients that have to be backpropagated through the $\alpha$-blending operations.

\subsection{Virtual Receivers}
\label{subsec:virtual_receivers}
So far, we have presented the use of TV to provide estimates of surface orientation, which are used in a loss encouraging the splats to match the estimated orientations (see Section~\ref{subsec:loss}). However, it does not provide a mechanism for affecting the positions of the splats. In the GS literature, the positions of the splats are updated by backpropagating from losses that operate on rendered RGB, depths or normals. 
Multiple such viewpoint-based gradients are combined to update each splat, but the mechanism is convoluted since backpropagation takes place through transmittance and depends on the configuration of the training viewpoints. Moreover, splats with low opacity and splats with high opacity ``hiding" behind other opaque splats may not receive meaningful updates.

We propose a viewpoint-independent technique for encouraging the centers of the splats to move towards the most salient local surface. To accomplish this, we sample \textbf{virtual receivers} along the normal of each splat (see the Tensor Voting Position Alignment block in Fig.~\ref{fig:overview}). 
These virtual receivers collect votes, but do not cast any since that would corrupt the true surfaces. After voting, we choose the point with the highest surface saliency, $\lambda_2-\lambda_3$, among the original splat center and the virtual receivers, and add a loss encouraging the center of the splat to move towards it.
To reduce the impact of outliers in this step, we filter splats with noisy neighborhoods and those that receive weak votes. 
Details are presented in the Supplement.

\subsection{Losses}
\label{subsec:loss}

As discussed in Section \ref{subsec:TV-SGS}, we aim to regularize the splats directly in world space. We use the following two 3D losses on the splats' normals and positions:
\begin{align}
&\mathcal{L}_{tvn} = \sum_{i}1 - \mathbf{|\hat{n}}_s^{\intercal}\mathbf{\hat{n}}_{tv}| \nonumber \\
&\mathcal{L}_{tvp} = \sum_{i}\lVert \mathbf{p}_s - \mathbf{p}_{tv}\rVert_1 \nonumber \\
&\mathcal{L}_{tv} = w_{tvn}\mathcal{L}_{tvn} + w_{tvp}\mathcal{L}_{tvp}
\label{eq:3d_losses}
\end{align}
where, $\mathbf{\hat{n}_s}, \;\mathbf{\hat{n}_{tv}}, \;\mathbf{p_{s}}, \;\mathbf{p_{tv}}$ are the normals and positions of the splats and TV estimates, all in world space, and $\;w_{tvn}, \;w_{tvp}$ are the weights of the respective loss terms. Note that gradients only flow through $\mathbf{\hat{n}}_s$ and $\mathbf{p}_s$.

Regularizing the geometry only in world space tends to over-smooth the fine details of the scene, which are present in the images. Therefore, in addition to the above 3D losses we also use rendering-based ones inherited from the respective backbone methods. 
Our final loss is:
\begin{align}
\mathcal{L} = \mathcal{L}_{backbone} + \mathcal{L}_{tv}
\end{align}
where $\mathcal{L}_{backbone}$ represents all the losses inherited from the respective backbones, and $\mathcal{L}_{tv}$ represents the two TV losses in (\ref{eq:3d_losses}). Figure~\ref{fig:overview} shows the interactions among the components of TV-SGS.

%% file: figs/overview.tex
\begin{figure*}[t!]
  \centering
    \includegraphics[width=0.8\linewidth]{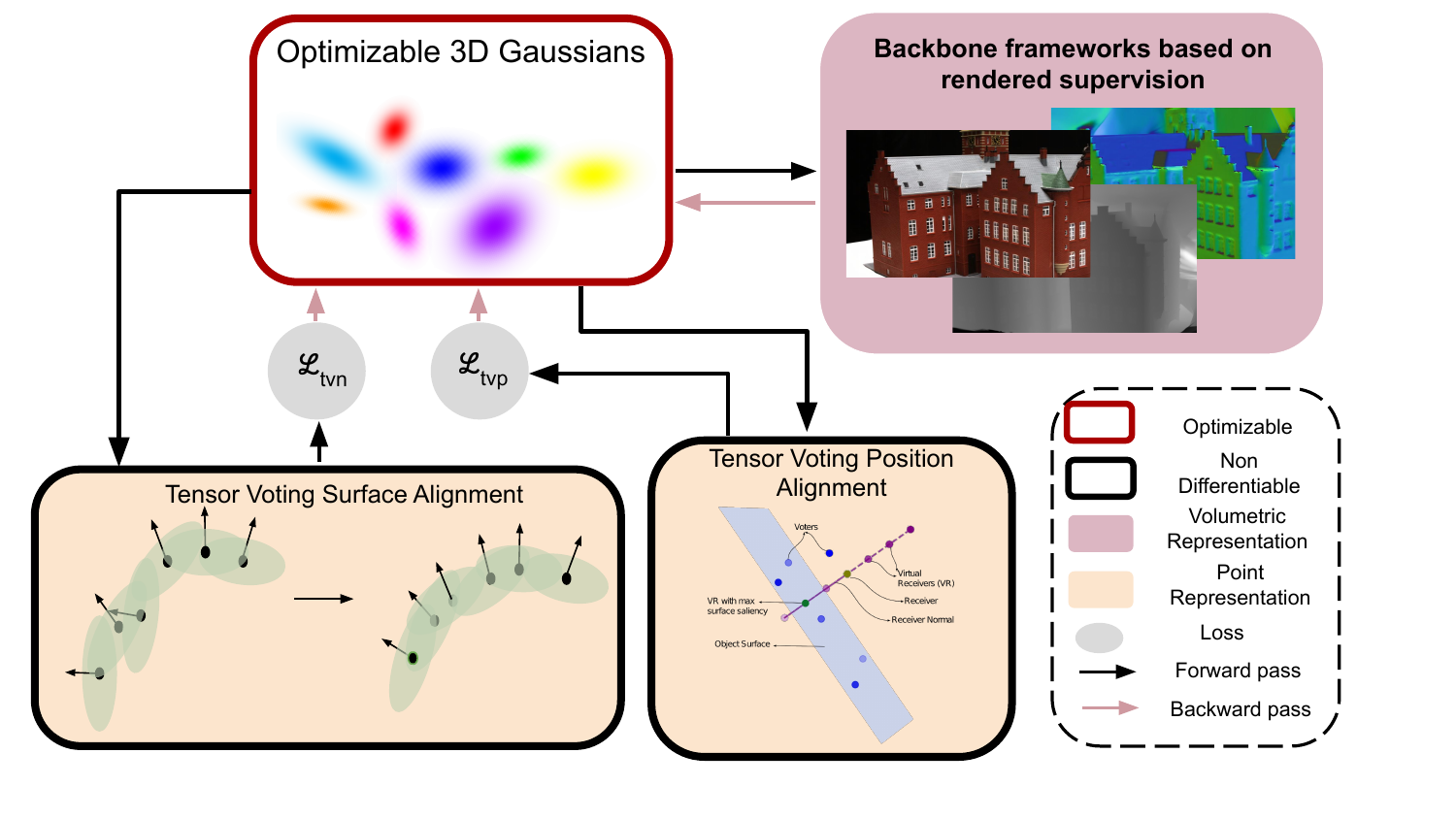}
    \caption{Overview of the proposed method. We infer and regularize geometry directly from the splats in world space, thereby influencing the geometry independently of opacity and scale. (This diagram applies for the integration of TV-SGS with any GS framework as the backbone.)}
  \label{fig:overview}
\end{figure*}

%% file: figs/stick_tensor.tex
\newcommand\hgaptt{0.5pt}             

\begin{figure}[t!]
\centering
\setlength{\tabcolsep}{\hgaptt}

\begin{tabular}{@{}cc@{}}
\includegraphics[width=0.42\linewidth]{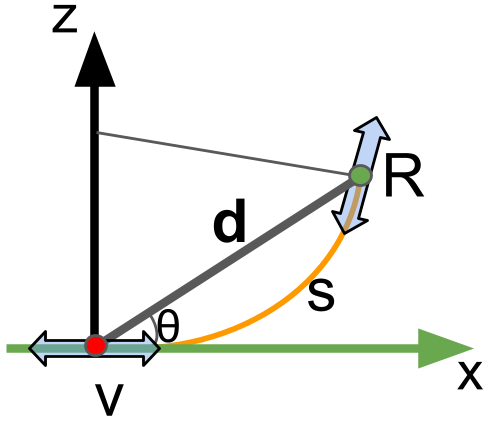} &
\includegraphics[width=0.52\linewidth]{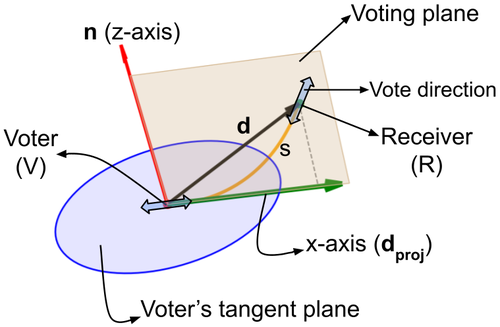} \\
\end{tabular}

\caption{Left: elementary stick votes are cast along the smooth circular arc connecting the voter and receiver in the voting plane. Right: a plate tensor casts a vote from a stick tensor contained in its 2D tangent space (the plate) that forms the smallest angle with $\boldsymbol{d}$. The angle may be 0 if $\boldsymbol{d}$ is in the tangent space. Otherwise, the projection of $\boldsymbol{d}$ in the tangent space is used to select  the voting stick. Equations (\ref{eq:decay_func}) and (\ref{eq:oriented_votes}) specify the vote.}
\label{fig:oriented_votes}
\end{figure}

%% file: sec/experiments.tex
\section{Experimental Results}
\label{sec:Exp}
\input{figs/fates_backbone_render_compare}

In this section, we present implementation details, followed by quantitative and qualitative results. More details can be found in the Supplement.

\subsection{Implementation Details and Data}

\paragraph{Implementation Details.}
We implement the core Tensor Voting module and neighborhood search with custom CUDA kernels. We have integrated TV-SGS with three backbones: FatesGS~\cite{huang2025fatesgs}, VGGS~\cite{xiang2026vggs} and PGSR~\cite{Chen_2024_PGSR}. We denote the corresponding implementation of TV-SGS by TV-SGS\textsubscript{F}, TV-SGS\textsubscript{V} and TV-SGS\textsubscript{P} respectively. In general, we use the losses of each backbone in addition to the ones in (\ref{eq:3d_losses}). Please see the Supplement for more details about the individual backbone losses. We set the weights of the backbone losses following the respective backbones, and $w_{tvn}, \; w_{tvp}$ to 0.5 and 1.0, respectively, for \textit{all} backbone implementations. We also keep all the training schedules the same as the backbones. In addition, results using all input training images for both datasets, and different state of the art dense view backbones PGSR \cite{Chen_2024_PGSR}, RaDeGS \cite{Zhang_2026_RaDe} and 2DGS \cite{Huang_2024_2DGS}, are shown in the Supplement. These results confirm the effectiveness of our method also in dense-view settings.

We start Tensor Voting in the first iteration  with the TV normal loss along with the TV position loss. We perform voting every 10 iterations of photometric optimization, and update each splat's neighbors every 100 iterations. We search for 200 neighbors in the original Euclidean space, and then select the 100 neighbors that are closest in the whitened space to receive votes. We also set the scale of voting $\tau_i$ equal to the median of the distances of these 100 neighbors in the whitened space.

We have used Mast3r \cite{duisterhof2025mast3r} to initialize the pointclouds for all backbones, keeping other backbone specific priors unchanged. We have also used pointclouds from MVSAnywhere \cite{izquierdo2025mvsanywhere} and COLMAP  \cite{schonberger2016structure} for initialization. Please see the Supplement for the evaluation using different initializations, and also for more implementation details.

\input{figs/dtu_compare_scan37}
\input{tab/tab_dtu_sparse_summary}

\paragraph{Datasets.}
We evaluate TV-SGS and the backbones on the Tanks and Temples (TnT) \cite{Knapitsch_2017_Tanks} and DTU \cite{Aanaes_2016_Large} datasets. 
We use the same scenes for testing as in previous works \cite{Huang_2024_2DGS,Yu_2024_Gaussian,Zhang_2026_RaDe}.
For geometry evaluation, we report the F1 scores for TnT and the average Chamfer distance (CD) for DTU using points sampled from the extracted mesh using TSDF depth fusion, as done in previous works \cite{Huang_2024_2DGS,Yu_2024_Gaussian,Zhang_2026_RaDe}. In addition, we also report the F1 and CD evaluated directly on the splat centers in the Supplement, to show the precision of the splat positions and the removal of floaters. We also report PSNR, SSIM, and LPIPS for TnT to evaluate NVS. 

\subsection{Quantitative and Qualitative Evaluation}
\label{subsec:Comparison}

\paragraph{Results on the DTU Dataset.}
We select 15 scenes and use views 23, 24 and 33 for \textit{3-view large-overlap reconstruction} (as in SparseNeuS \cite{long2022sparseneus})  and views 22, 25 and 28 for \textit{3-view small-overlap reconstruction} (as in PixelNeRF \cite{yu2021pixelnerf}), following previous work \cite{huang2025fatesgs, xiang2026vggs}. 
Examples of the different degrees of overlap are shown in the Supplement.

Table~\ref{tab:dtu_sparse_summary} shows the geometric evaluation on small and large-overlap 3-view selections for DTU, evaluated on the extracted mesh. The implementations of TV-SGS surpass the accuracy of their own backbones for both small and large overlap settings. Figures~\ref{fig:fates_backbone_render_compare} and \ref{fig:dtu_compare_scan37} show qualitative comparisons between the backbones and the respective TV-SGS implementations.
Please see the Supplement for the per scene CD for both DTU small and large overlap splits.

Among the backbones used to evaluate TV-SGS, PGSR is expected to perform poorly on both small and large-overlap setting since it is designed for dense views without any pre-trained priors. FatesGS is designed primarily for large overlap but the authors report metrics for small overlap as well. The authors of VGGS mainly focus on the small-overlap setting, and do not present results with large image overlaps.

\paragraph{Results on the TnT Dataset.}
We uniformly select 20 views for each scene, leading to very low overlaps especially in the larger scenes such as Courthouse. Table~\ref{tab:tnt_20_views} shows geometric results on TnT evaluated on the extracted mesh while Table~\ref{tab:sparse_tnt_nvs_20_views} summarizes the rendering statistics. Figures~\ref{fig:tnt_compare_barn} and \ref{fig:fates_backbone_render_compare} provide visual comparisons. Note that most of the sparse-view methods
exclude Courthouse and Meetingroom since they are too large for sparse view reconstruction. We have included these scenes here for completeness.

\input{tab/tab_tnt_20_views_m3_init}
\input{tab/tab_tnt_nvs}

TV-SGS is particularly well suited for complex scenes with less overlap on the surfaces due to its 3D losses that ensure all splats receive supervision.
TV-SGS outperforms all its backbones quantitatively by about 7\% in geometry while maintaining their rendering quality.

\input{tab/tab_TnT_ablations}

\paragraph{Ablation Study}
Here we take a closer look at how each component of TV-SGS\textsubscript{F} contributes to the final reconstruction quality. Results are averaged over the four smaller scenes of the TnT dataset after surface extraction using TSDF fusion. 

As shown in Table~\ref{tab:abl_tnt_20_views}, the TV normal loss clearly has the strongest effect in isolation while whitening improves upon the gains produced by either of the other losses. (Activating whitening by itself is pointless.) The TV position loss benefits from the activations of the normal loss, which leads to better placement of the virtual receivers. All three components together produce the best results. The effects of TV position loss are not captured completely by this ablation study. The TV position loss it helps with removing floaters, especially behind objects which have little effect on the metrics. Please see the Supplement where we visualize these effects of the TV position loss.

%% file: figs/fates_backbone_render_compare.tex
\renewcommand\cellwt{0.31\linewidth} 
\renewcommand\hgapt{0.5pt}             
\renewcommand\vgapt{0.5pt}             

\begin{figure}[b]
\centering
\setlength{\tabcolsep}{\hgapt}

\begin{tabular}{@{}lccc@{}}
{} & \small \makecell{TV-SGS\textsubscript{F}\\rendered RGB} & \small\makecell{FatesGS\\rendered normals}
    & \small\makecell{TV-SGS\textsubscript{F}\\rendered normals} \\
\rotatebox{90}{\hspace{-0.15cm} Caterpillar} &
\includegraphics[width=\cellwt]{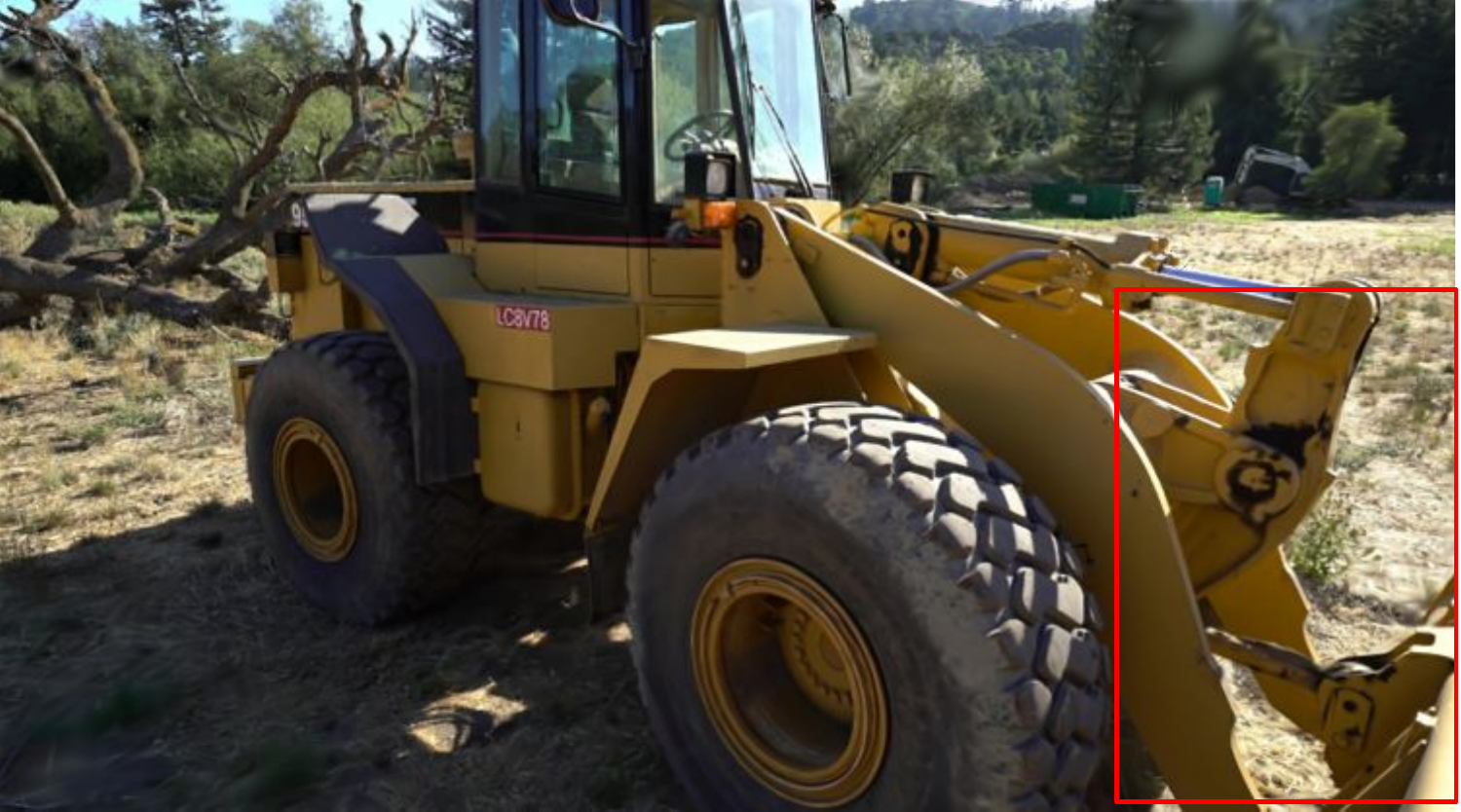} &
\includegraphics[width=\cellwt]{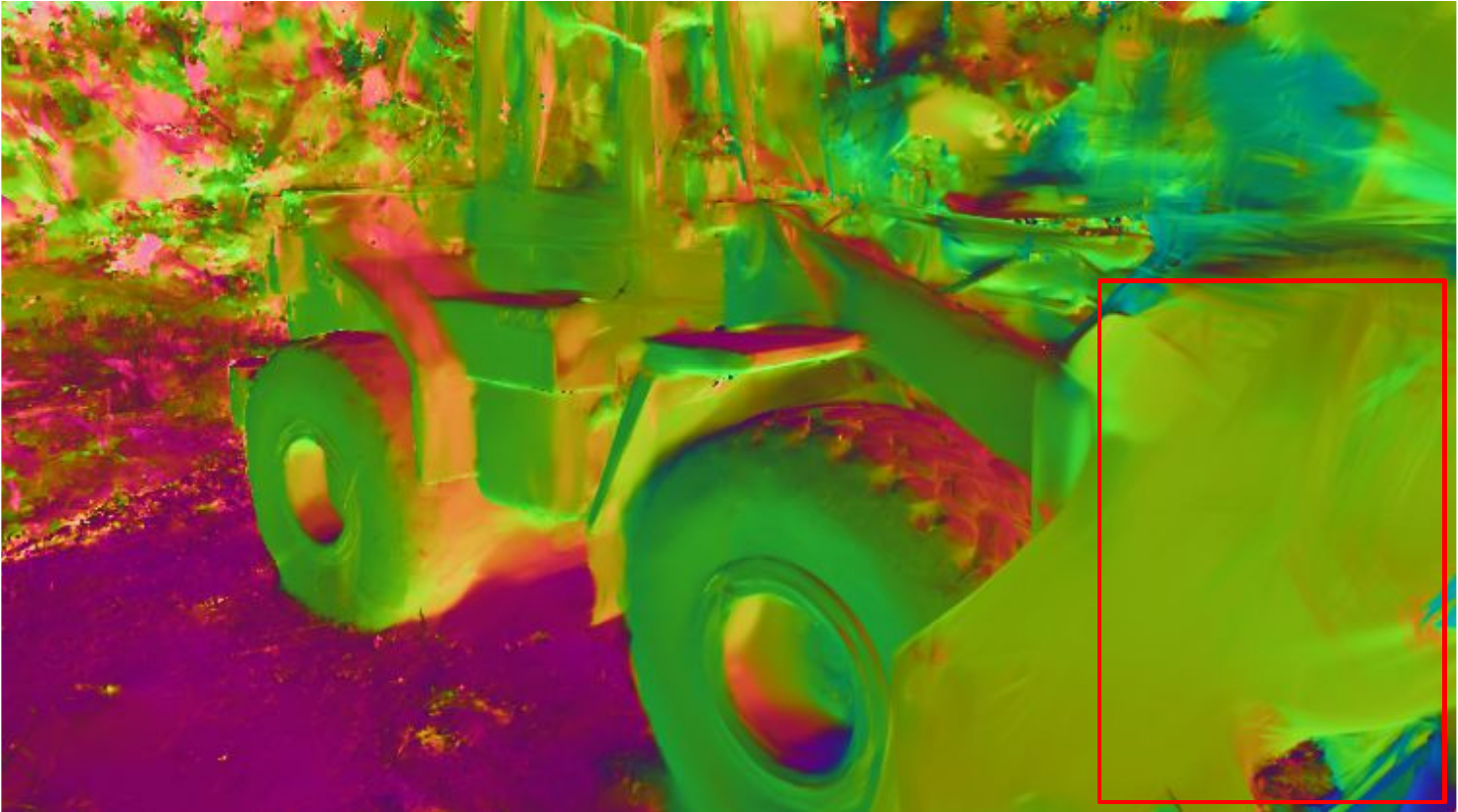} &
\includegraphics[width=\cellwt]{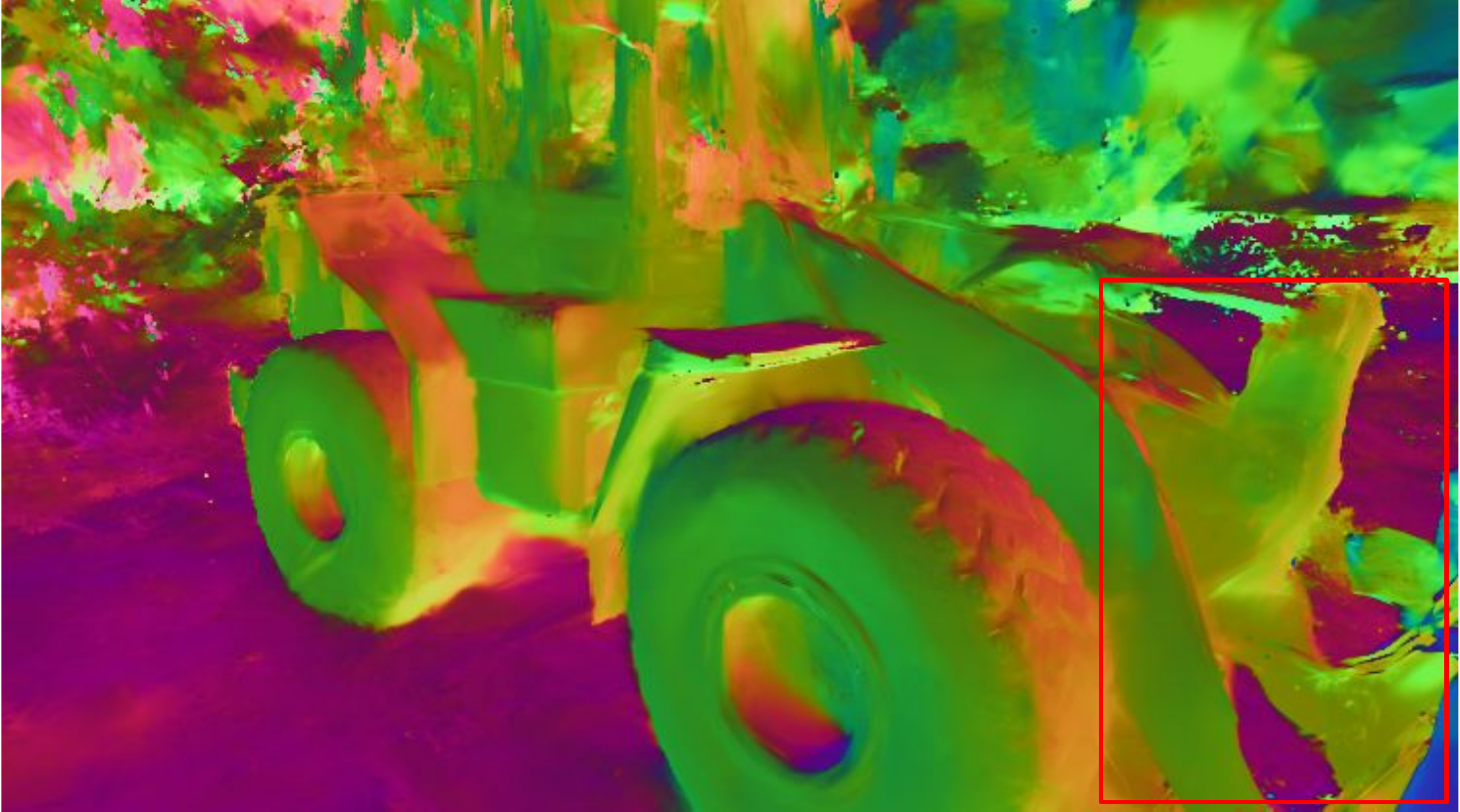} \\[\vgapt]
\rotatebox{90}{\hspace{0.35cm} Scan97} &
\includegraphics[width=\cellwt]{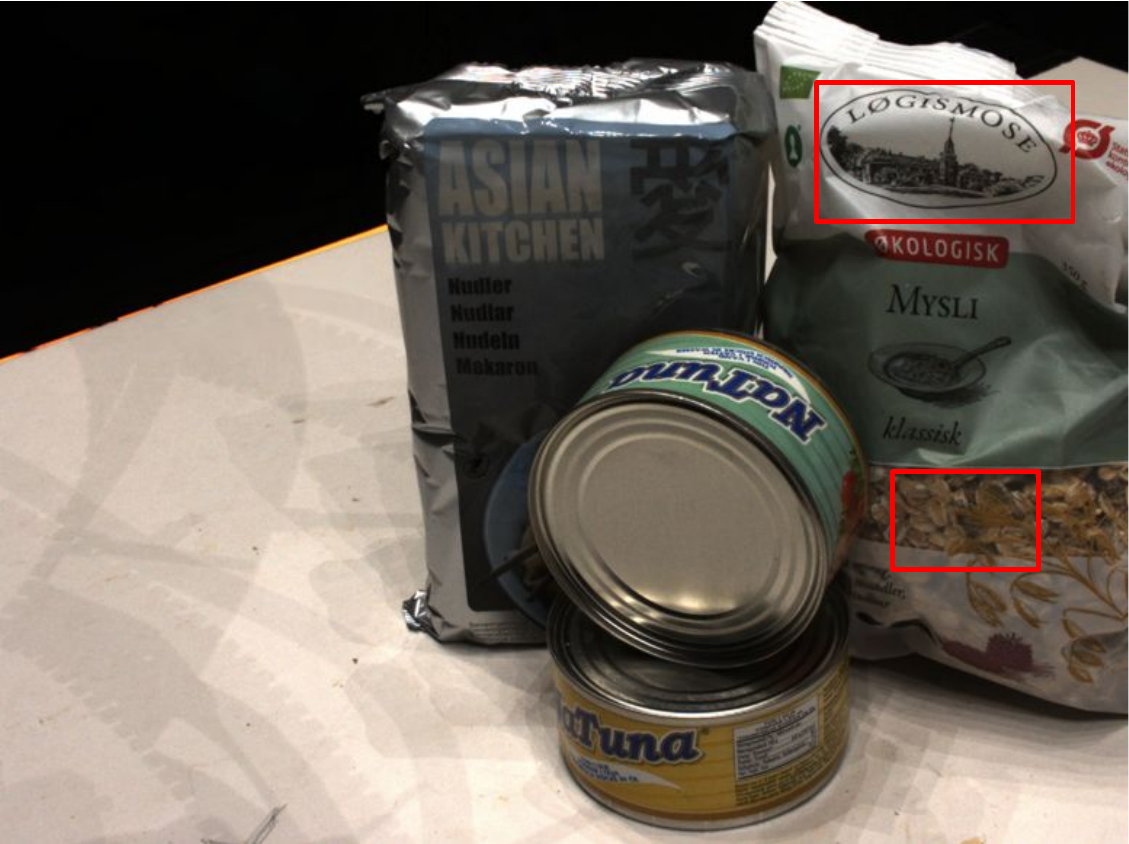} &
\includegraphics[width=\cellwt]{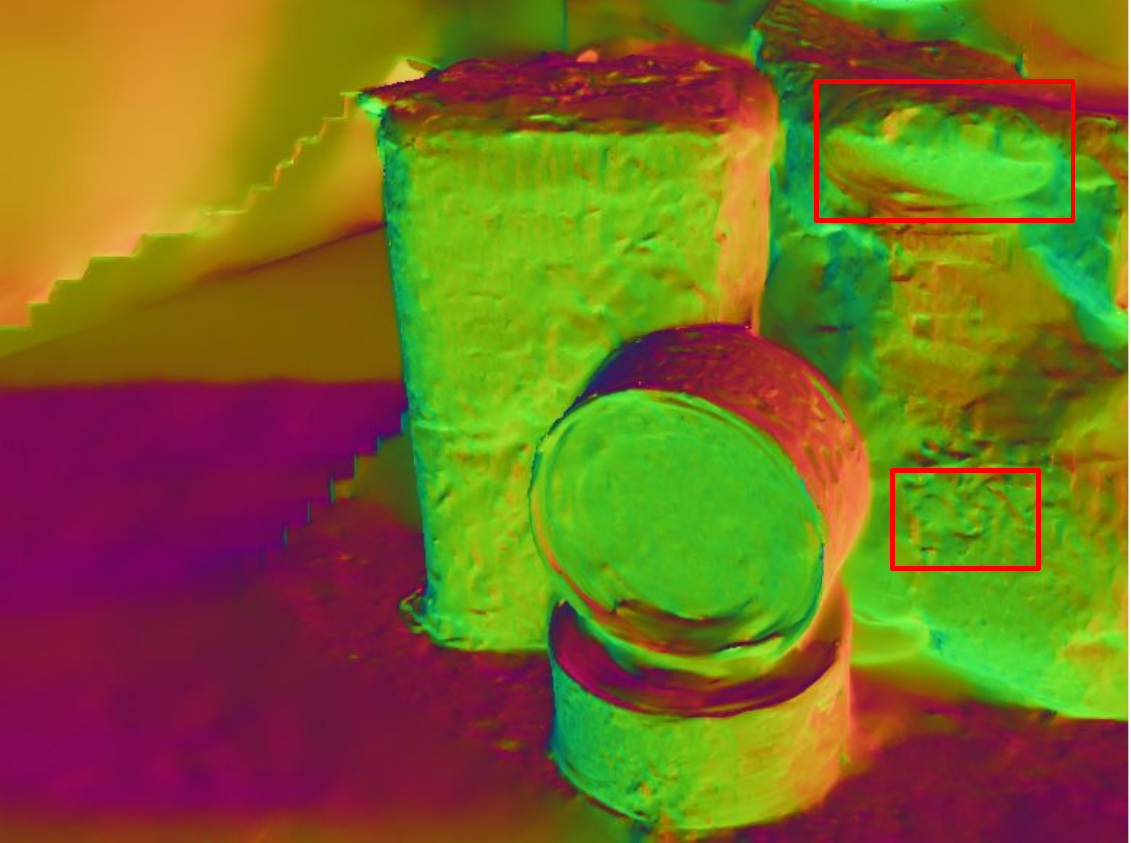} &
\includegraphics[width=\cellwt]{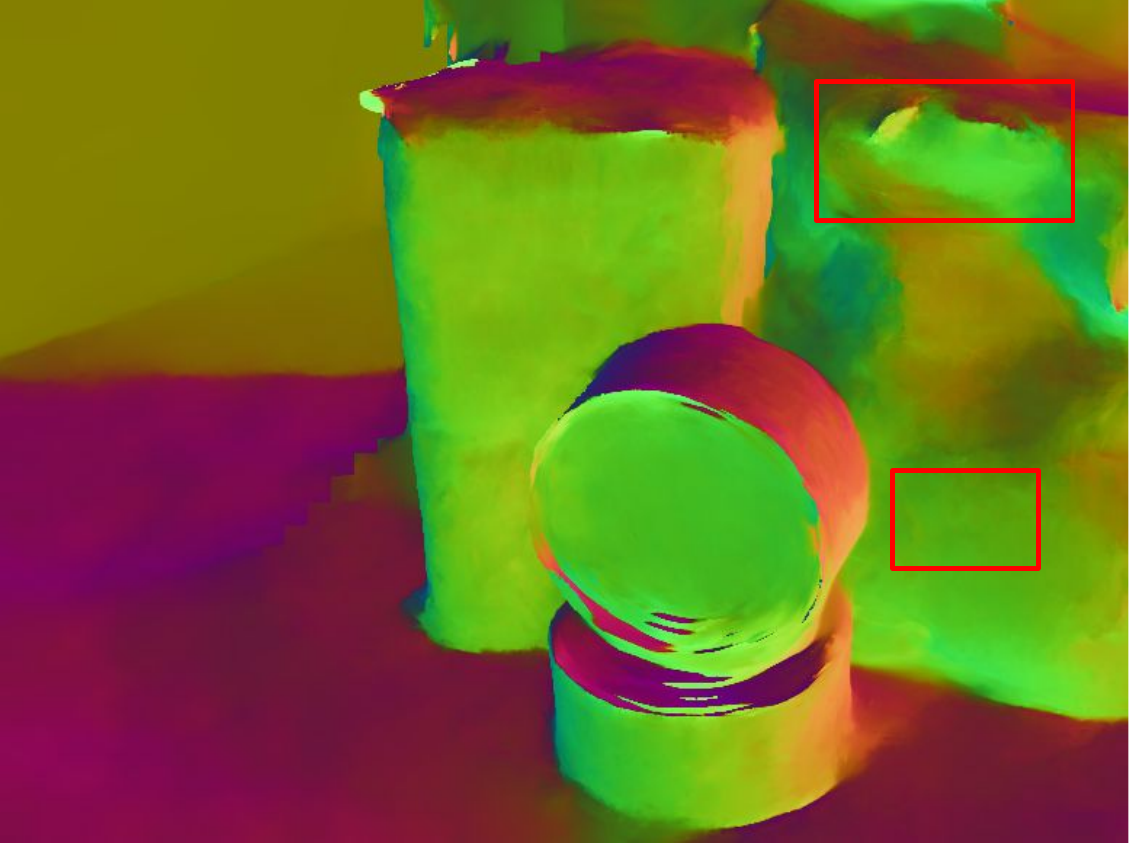}
\end{tabular}

\caption{Qualitative comparisions of the rendered normals between FatesGS and TV-SGS\textsubscript{F} on Caterpillar scene from the TnT dataset and Scan97 from the DTU dataset. Notice the absence of floaters in TV-SGS\textsubscript{F} rendered normals in Caterpillar. Also notice how texture influences the normals for FatesGS in Scan97.}
\label{fig:fates_backbone_render_compare}
\end{figure}

%% file: figs/dtu_compare_scan37.tex
\renewcommand\cellwt{0.32\linewidth} 
\renewcommand\hgapt{0.5pt}             
\renewcommand\vgapt{0.5pt}             

\begin{figure}[t!]
\centering
\setlength{\tabcolsep}{\hgapt}

\begin{tabular}{@{}lccc@{}}
{} & PGSR & VGGS & FatesGS \\
\rotatebox{90}{Backbones} &
\includegraphics[width=\cellwt]{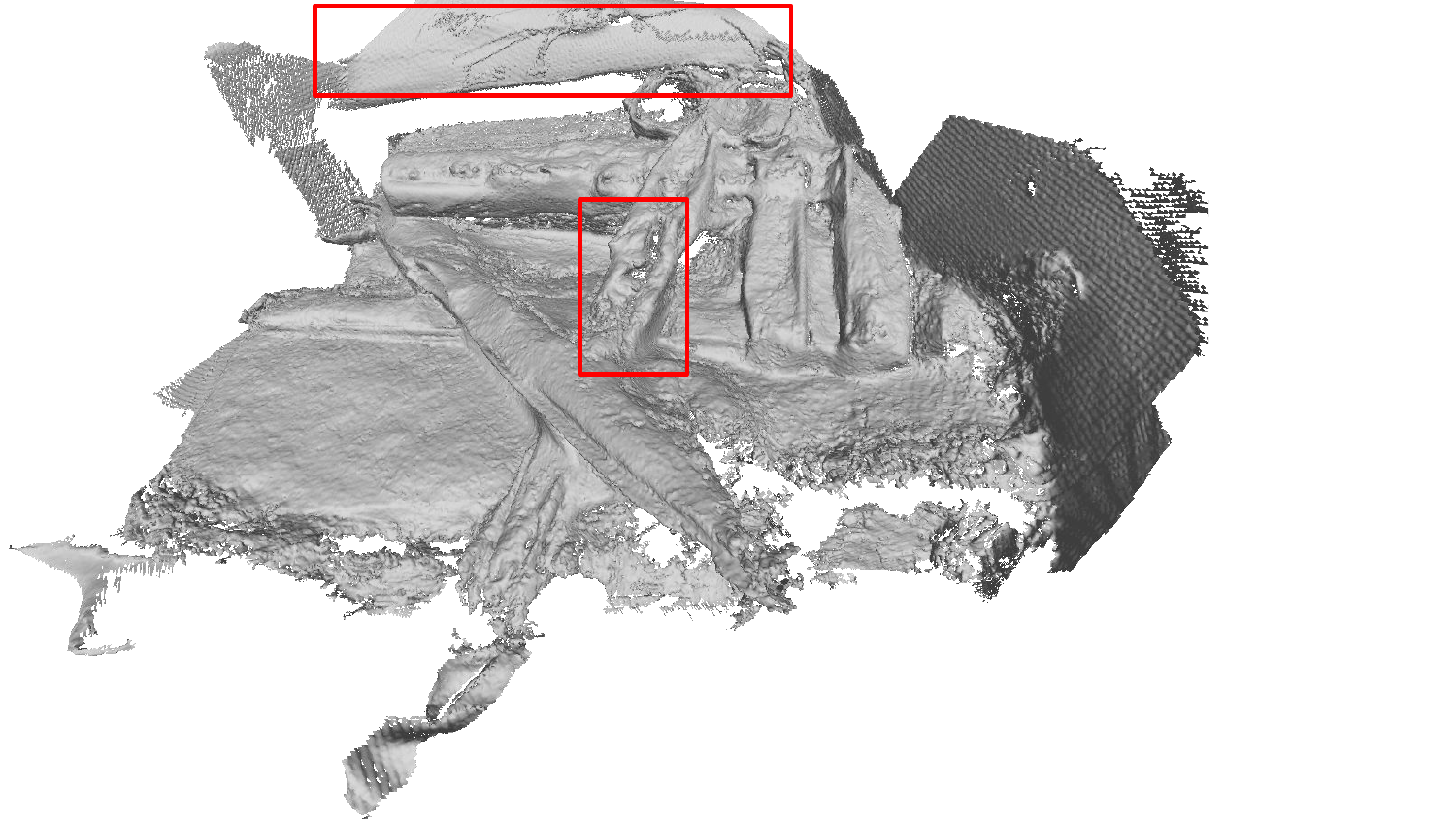} &
\includegraphics[width=\cellwt]{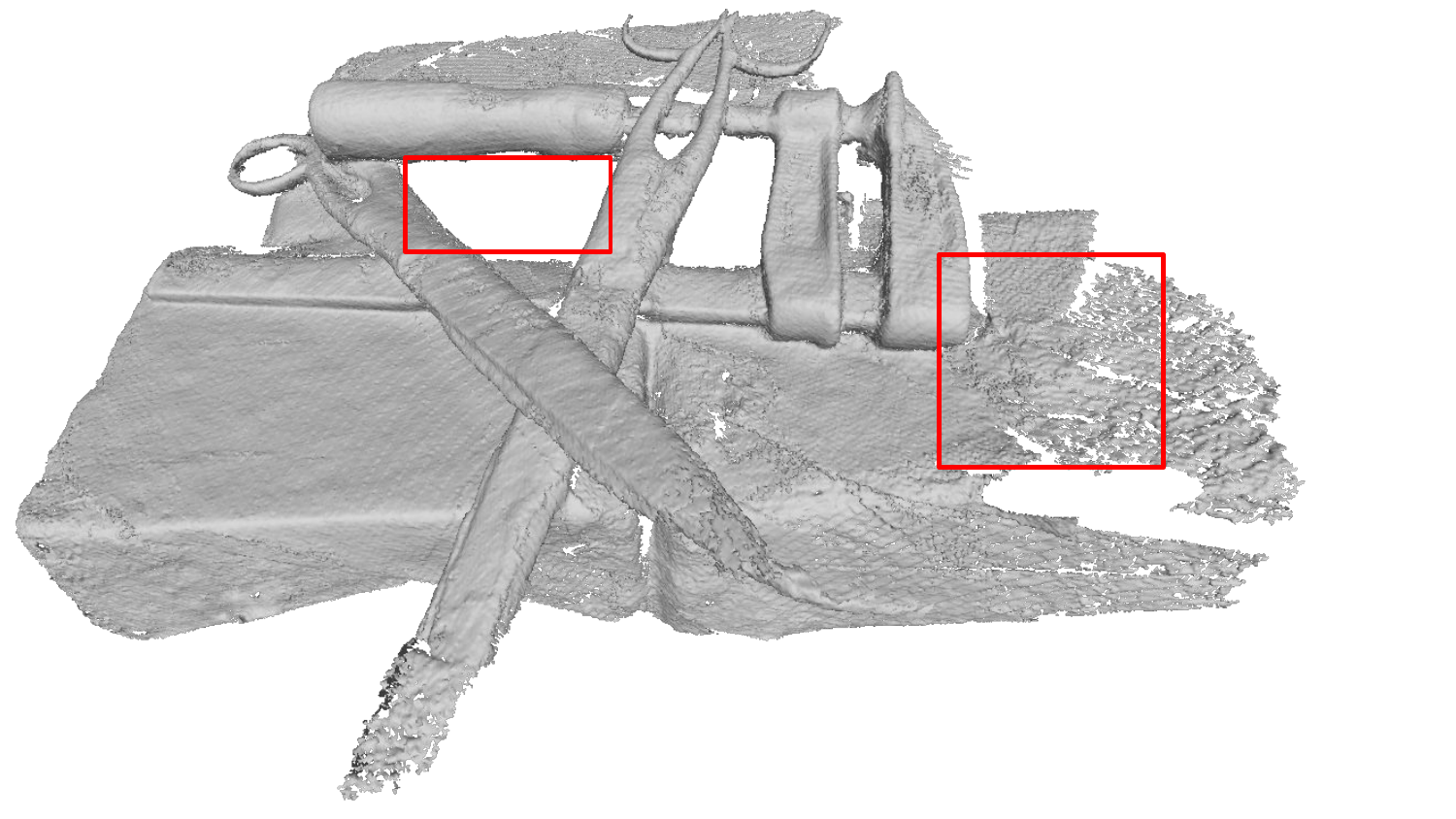} &
\includegraphics[width=\cellwt]{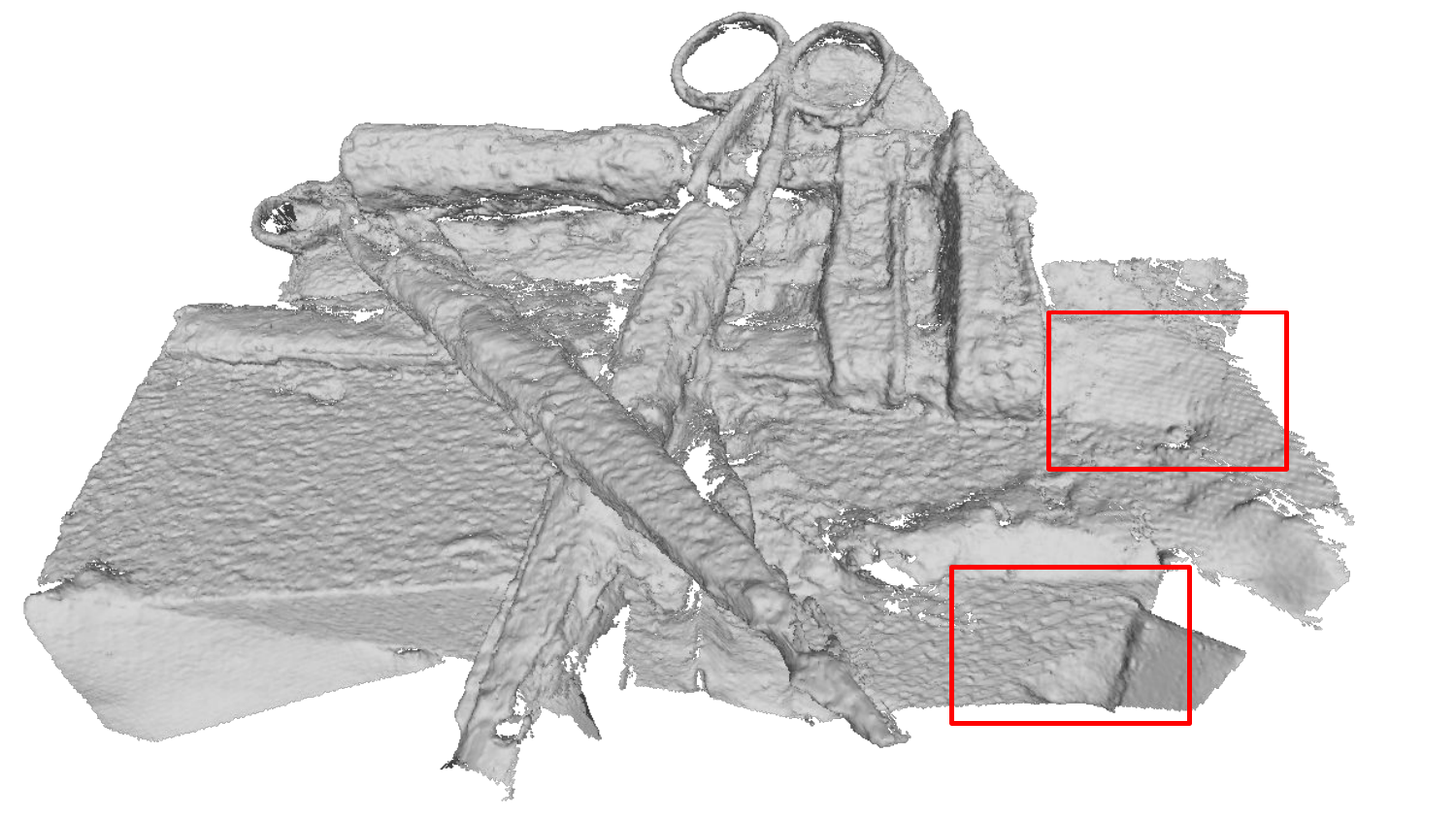} \\[\vgapt]
\rotatebox{90}{TV-SGS} &
\includegraphics[width=\cellwt]{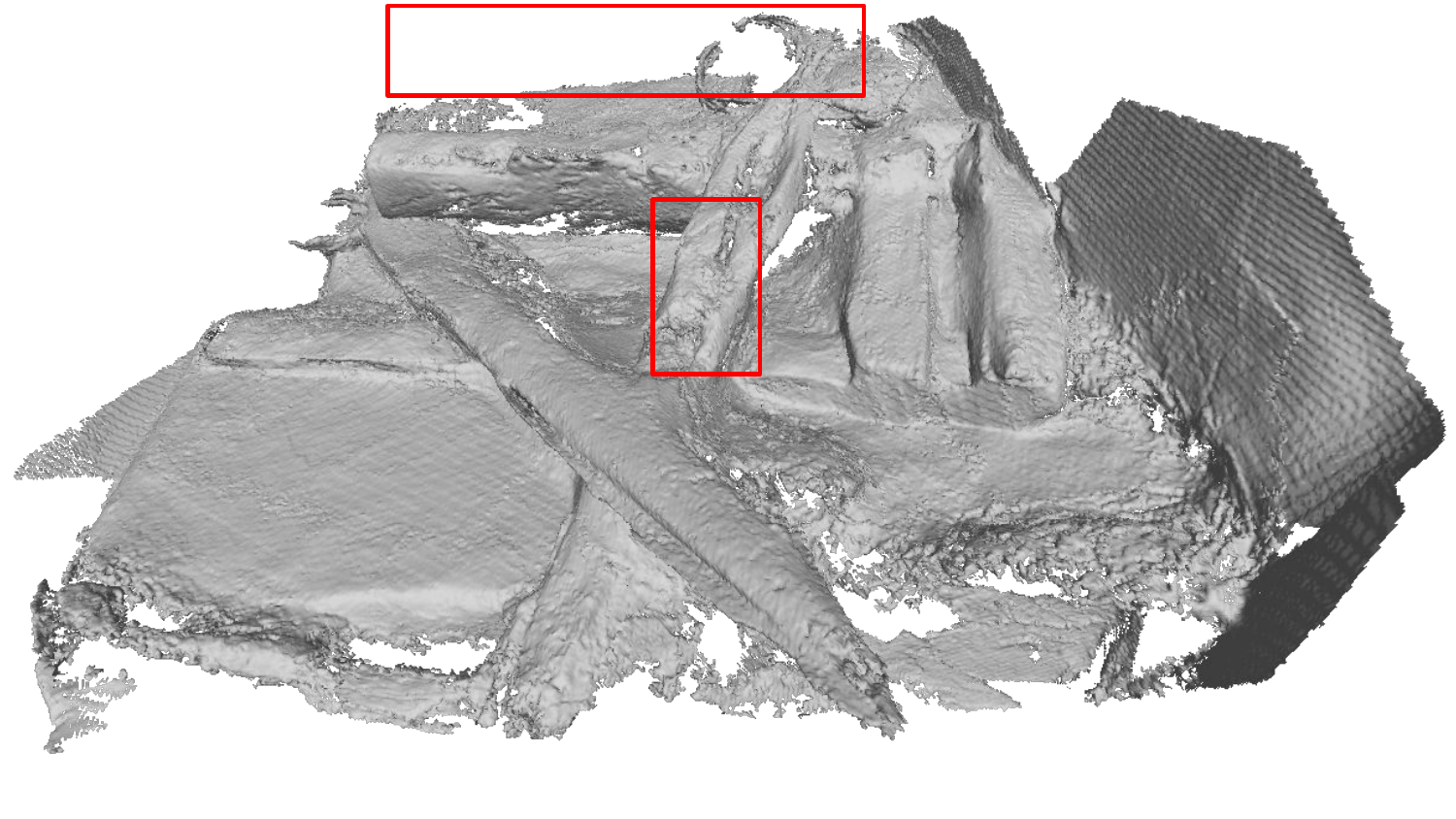} &
\includegraphics[width=\cellwt]{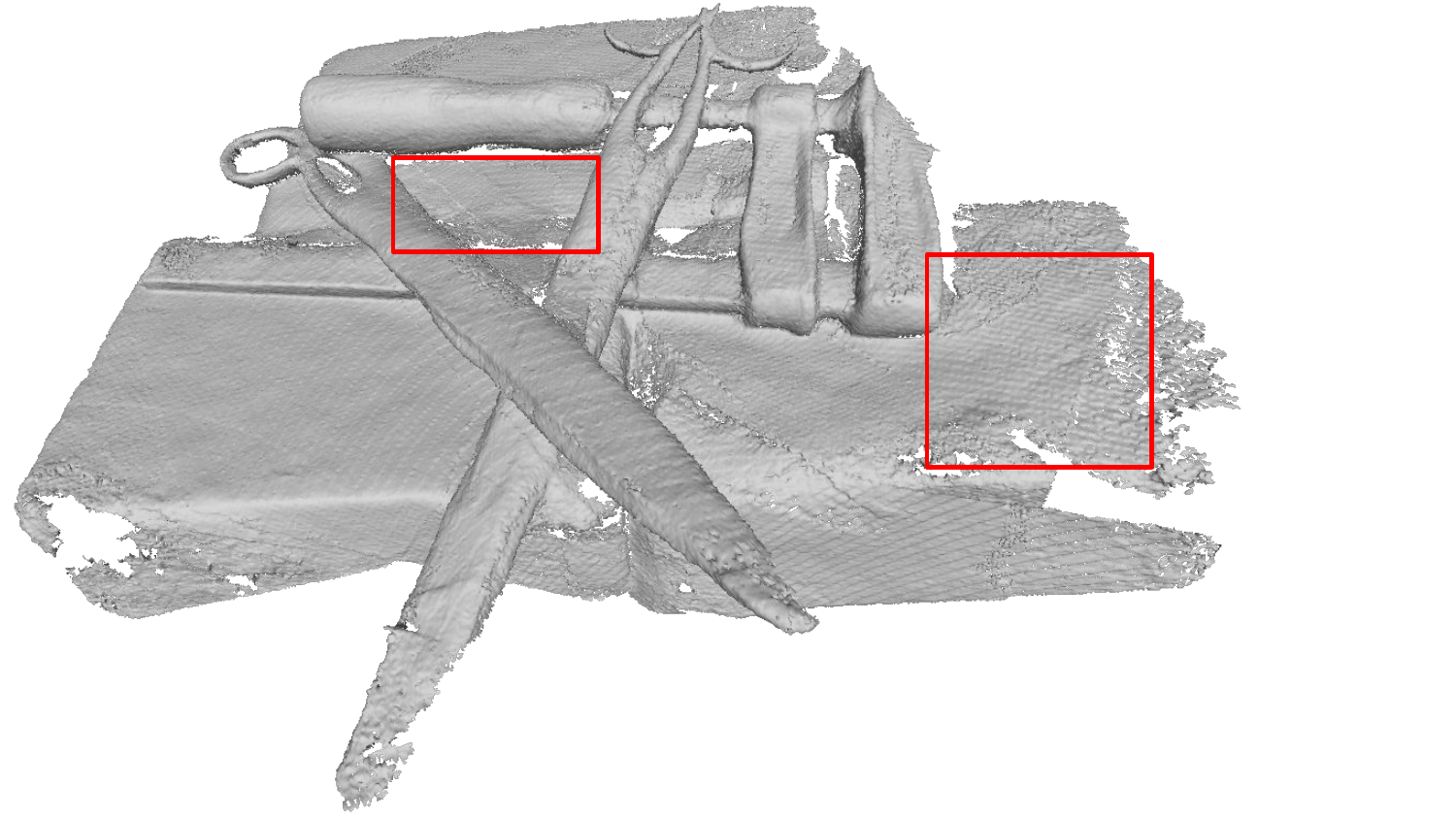} &
\includegraphics[width=\cellwt]{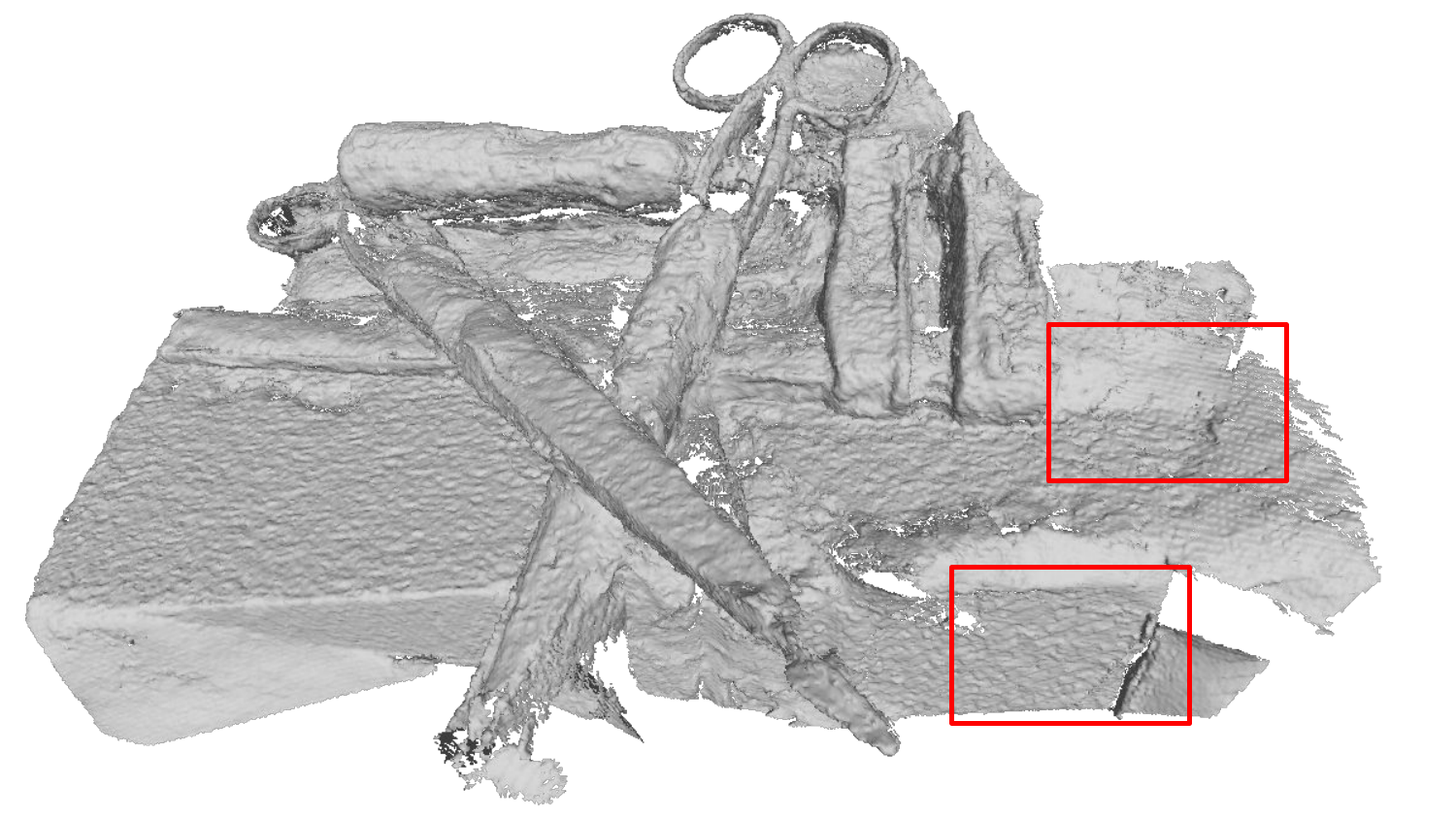}
\end{tabular}

\caption{Qualitative comparison of surface reconstruction on the DTU dataset between the PGSR, VGGS and FatesGS backbones before and after integration with TV-SGS. Top: results by the backbones. Bottom: results by the respective TV-SGS implementations.}
\label{fig:dtu_compare_scan37}
\end{figure}

%% file: tab/tab_dtu_sparse_summary.tex
\begin{table}[htbp]
  \centering
  \setlength{\tabcolsep}{3pt}
  \begin{tabular}{@{}l|c||c}
    {} & \textbf{Small Overlap} & \textbf{Large overlap} \\
    \midrule
    Methods & Mean CD $\downarrow$ & Mean CD $\downarrow$ \\
    \midrule
    $\mathrm{PGSR}$\cite{Chen_2024_PGSR} & 2.56 & 1.74 \\
   $\mathbf{TV-SGS}_{\mathbf{P}}$ & \textbf{2.18} & \textbf{1.60} \\
   \midrule
    $\mathrm{SparseNeus_{ft}}$ \cite{long2022sparseneus} & -- & 1.27 \\
    $\mathrm{VolRecon}$ \cite{Ren_2023_VolRecon} & -- & 1.38 \\
   $\mathrm{GenS_{ft}}$ \cite{peng2023gens} &  -- & 1.03 \\
    $\mathrm{ReTR}$ \cite{liang2023retr} & -- & 1.17 \\
    $\mathrm{UFORecon}$ \cite{na2024uforecon} & -- & 0.99 \\
    \midrule
   $\mathrm{MonoSDF}$ \cite{yu2022monosdf} & 1.86 & 2.93 \\
    $\mathrm{NeuSurf}$ \cite{huang2024neusurf} & 1.35 & 0.99 \\
    $\mathrm{SparseCraft}$ \cite{younes2024sparsecraft} & -- & 1.04 \\
    \midrule
    $\mathrm{Sparse2DGS}$ \cite{wu2025sparse2dgs} & -- & 1.13 \\
    $\mathrm{FatesGS}$ \cite{huang2025fatesgs} & 1.07 & 0.95 \\
    $\mathbf{TV-SGS}_{\mathbf{F}}$ & \textbf{1.03} & \textbf{0.92} \\
    $\mathrm{VGGS}$ \cite{xiang2026vggs} & 1.06 & 1.29 \\
    $\mathbf{TV-SGS}_{\mathbf{V}}$ & \textbf{1.00} & \textbf{1.20} \\
    \hline
  \end{tabular}
  \caption{Geometric evaluation on \textbf{DTU} using 3 views with \textbf{small} and \textbf{large} overlap  and Mast3r initialization. The methods are divided into four categories, from top to bottom: (1) dense-view reconstruction methods related to TV-SGS (but trained on 3 views), (2)  generalizable sparse-view reconstruction methods (3) implicit sparse-view inference time optimization methods, and (4) explicit sparse-view inference time optimization methods. We mark with boldface the best result between a backbone and its corresponding TV-SGS implementation. The values of all methods are taken from the respective papers, except the backbones used by TV-SGS, for which we executed the authors' code.}
  \label{tab:dtu_sparse_summary}
\end{table}

%% file: tab/tab_tnt_20_views_m3_init.tex
\begin{table}[t!]
  \centering
  \setlength{\tabcolsep}{3pt}
  \resizebox{\columnwidth}{!}{%
  \begin{tabular}{c|c||c||c}
    \hline
     & $\mathbf{P}$ / $\mathbf{R}$ / $\mathbf{F1}$ & $\mathbf{P}$ / $\mathbf{R}$ / $\mathbf{F1}$ & $\mathbf{P}$ / $\mathbf{R}$ / $\mathbf{F1}$\\
    \hline
    \textbf{Scenes} & $\mathbf{PGSR}$ & $\mathbf{FatesGS}$ & $\mathbf{VGGS}$ \\
    \hline
    Barn & 0.231 / 0.144 / 0.177 & 0.318 / 0.318 / 0.318 & 0.286 / 0.265 / 0.275 \\
    Caterpillar & 0.197 / 0.298 / 0.237 & 0.149 / 0.216 / 0.177 & 0.146 / \textbf{0.240} / \textbf{0.181} \\
    Courthouse & \textbf{0.122} / \textbf{0.001} / \textbf{0.003} & 0.030 / 0.005 / 0.009 & 0.027 / \textbf{0.009} / 0.014 \\
    Ignatius & \textbf{0.552} / \textbf{0.586} / \textbf{0.569} & \textbf{0.495} / 0.517 / 0.506 & 0.292 / 0.322 / 0.306 \\
    Meetingroom & 0.107 / 0.039 / 0.057 & 0.086 / 0.046 / 0.060 & \textbf{0.192} / 0.119 / 0.147 \\
    Truck & 0.422 / 0.484 / 0.451 & 0.424 / 0.453 / 0.438 & 0.314 / 0.362 / 0.336 \\
    \hline
    \textbf{Average} & \textbf{0.272} / 0.259 / 0.249 & 0.250 / 0.259 / 0.251 & 0.210 / 0.219 / 0.210 \\
    \hline
     & $\mathbf{TV-SGS}_{\mathbf{P}}$ & $\mathbf{TV-SGS}_{\mathbf{F}}$ & $\mathbf{TV-SGS}_{\mathbf{V}}$ \\
     \hline
    Barn & \textbf{0.260} / \textbf{0.206} / \textbf{0.230} & \textbf{0.332} / \textbf{0.341} / \textbf{0.336} & \textbf{0.320} / \textbf{0.305} / \textbf{0.312} \\
    Caterpillar & \textbf{0.211} / \textbf{0.311} / \textbf{0.251} & \textbf{0.171} / \textbf{0.262} / \textbf{0.207} & \textbf{0.147} / 0.238 / \textbf{0.181} \\
    Courthouse & 0.009 / 0.001 / 0.002 & \textbf{0.049} / \textbf{0.007} / \textbf{0.012} & \textbf{0.036} / \textbf{0.009} / \textbf{0.014} \\
    Ignatius & 0.549 / 0.578 / 0.563 & 0.491 / \textbf{0.522} / \textbf{0.506} & \textbf{0.322} / \textbf{0.342} / \textbf{0.332} \\
    Meetingroom & \textbf{0.133} / \textbf{0.053} / \textbf{0.075} & \textbf{0.093} / \textbf{0.050} / \textbf{0.065} & 0.189 / \textbf{0.122} / \textbf{0.149} \\
    Truck & \textbf{0.453} / \textbf{0.514} / \textbf{0.482} & \textbf{0.480} / \textbf{0.503} / \textbf{0.491} & \textbf{0.335} / \textbf{0.391} / \textbf{0.361} \\
    \hline
    \textbf{Average} & 0.269 / \textbf{0.277} / \textbf{0.267} & \textbf{0.269} / \textbf{0.281} / \textbf{0.270} & \textbf{0.225} / \textbf{0.235} / \textbf{0.225} \\
    \hline
  \end{tabular}
  }
  \caption{Quantitative results on geometry on the \textbf{TnT} dataset using 20 training views, using Mast3r pointcloud initialization. The meshes are extracted via TSDF depthmap fusion. P/R/F1 refers to precision, recall and F1 scores. TV-SGS improves significantly upon all three backbones, except for one metric.}
  \label{tab:tnt_20_views}
\end{table}

%% file: tab/tab_tnt_nvs.tex
\begin{table}[t!]
  \centering
  \setlength{\tabcolsep}{3pt}
  \resizebox{\columnwidth}{!}{%
  \begin{tabular}{c|c||c||c}
  \hline
    & $\mathbf{P / S / L}$ & $\mathbf{P / S / L}$ & $\mathbf{P / S / L}$ \\
     \hline
    \textbf{Scenes} & $\mathbf{PGSR}$ & $\mathbf{FatesGS}$ & $\mathbf{VGGS}$  \\
     \hline
    Barn & 15.095 / 0.574 / 0.384 & 16.788 / 0.670 / 0.299 & 16.898 / 0.661 / 0.324 \\
    Caterpillar & 15.235 / 0.484 / 0.402 & 15.444 / 0.499 / 0.381 & 15.520 / 0.531 / 0.377 \\
    Courthouse & 10.250 / 0.287 / 0.585 & 11.877 / 0.385 / 0.542 & 11.680 / 0.339 / 0.559 \\
    Ignatius & 17.818 / 0.623 / 0.256 & 17.351 / 0.599 / 0.272 & 17.247 / 0.603 / 0.288 \\
    Meetingroom & 15.902 / 0.570 / 0.388 & 16.502 / 0.601 / 0.361 & 17.685 / 0.636 / 0.350 \\
    Truck & 17.264 / 0.683 / 0.253 & 17.134 / 0.679 / 0.263 & 18.172 / 0.704 / 0.247 \\
    \hline
    \textbf{Average} & 15.261 / 0.537 / 0.378 & 15.849 / 0.572 / 0.353 & 16.200 / 0.579 / 0.357 \\
    \hline
     & $\mathbf{TV-SGS}_{\mathbf{P}}$ & $\mathbf{TV-SGS}_{\mathbf{F}}$ & $\mathbf{TV-SGS}_{\mathbf{V}}$ \\
     \hline
    Barn & 15.122 / 0.586 / 0.375 & 15.959 / 0.656 / 0.314 & 16.766 / 0.663 / 0.314 \\
    Caterpillar & 15.166 / 0.487 / 0.400 & 15.394 / 0.495 / 0.386 & 15.428 / 0.536 / 0.369 \\
    Courthouse & 10.582 / 0.304 / 0.586 & 11.714 / 0.385 / 0.545 & 11.631 / 0.350 / 0.558 \\
    Ignatius & 17.614 / 0.621 / 0.259 & 17.345 / 0.600 / 0.270 & 17.294 / 0.603 / 0.288 \\
    Meetingroom & 15.428 / 0.564 / 0.400 & 16.454 / 0.604 / 0.365 & 17.533 / 0.635 / 0.350 \\
    Truck & 17.044 / 0.677 / 0.257 & 17.042 / 0.680 / 0.259 & 18.366 / 0.705 / 0.246 \\
    \hline
    \textbf{Average} & 15.159 / 0.540 / 0.380 & 15.651 / 0.570 / 0.356 & 16.170 / 0.582 / 0.354 \\
    \hline
  \end{tabular}
  }
  \caption{Quantitative results on \textbf{NVS} on the \textbf{TnT} dataset. TV-SGS maintains or improves upon the baselines rendering quality, showing that we are able to improve geometry while maintaining rendering quality. P/S/L refers to PSNR/SSIM/LPIPS respectively}
  \label{tab:sparse_tnt_nvs_20_views}
\end{table}

%% file: tab/tab_TnT_ablations.tex
\begin{table}[t]
\centering
\begin{tabular}{@{}l|ccc|c@{}}
\toprule
& Whiten & $\mathcal{L}_{tvn}$ & $\mathcal{L}_{tvp}$ & F1 $\uparrow$ \\ \hline
FatesGS Backbone&    &    &    &    0.359 \\
\midrule 
$\mathcal{L}_{tvn}$ only&   &   \usym{1F5F8}    &   &    0.366     \\
$\mathcal{L}_{tvp}$ only&   &    &  \usym{1F5F8}  &    0.352     \\
$\mathcal{L}_{tvn} + \mathcal{L}_{tvp}$&    &    \usym{1F5F8}    &    \usym{1F5F8}     &    0.366   \\
whiten + $\mathcal{L}_{tvp}$& \usym{1F5F8}  &    &  \usym{1F5F8}   &    0.358   \\
whiten + $\mathcal{L}_{tvn}$& \usym{1F5F8} &\usym{1F5F8} &    & 0.374   \\ \midrule
TV-SGS\textsubscript{F}& \usym{1F5F8} & \usym{1F5F8}    &    \usym{1F5F8}   &   0.385    \\
\bottomrule
\end{tabular}

\caption{Ablation on TnT dataset averaged on \textbf{4 scenes} excluding Courthouse and Meetingroom. 
}
\label{tab:abl_tnt_20_views}
\end{table}

%% file: sec/conclusion.tex
\section{Conclusion}
\label{sec:conclusion}

We have presented TV-SGS, an approach for Gaussian splatting under sparse views based on a re-formulation of Tensor Voting. TV-SGS embodies both algorithmic novelty and practical value. It is successful in enhancing geometric accuracy due to two new 3D losses presented in this paper. These losses, which are complementary to those already available in the literature, rely on geometric information computed by TV to supervise splat optimization while maintaining rendering quality. We have demonstrated TV-SGS's capabilities by integrating it with multiple recent backbones without having to modify TV-specific hyperparameters, demonstrating the robustness of our approach. We have also evaluated it with different initialization schemes, on multiple sparse-view settings as well as on the standard dense-view setting. (See also the Supplement.)

The current implementation of TV-SGS can serve as the foundation for further progress along several directions. First, tensor voting is differentiable and its parameters, such as the scale of voting, can be optimized at inference time. Second, we will investigate ways to couple TV with GS more tightly by harmonizing the eigenvalues of the tangent space of the covariance matrix $\boldsymbol{\Sigma}$ and the tensor $\boldsymbol{T}$. 
Our software will be open-sourced upon acceptance.

%% file: sec_supp/implementation_details.tex
\section{Implementation Details}
\label{sec_supp:impl_details}

This section includes more details on: the calculation of the constant $c$, which controls the attenuation of vote strength due to curvature (Section~\ref{subsec_supp:vote_strength_attn}); the way the scale of voting is set for each voter (Section~\ref{subsec_supp:vote_scale}); the placement of virtual receivers (Section~\ref{subsec_supp:vir_rec}); the implementation of the whitening transform (Section~\ref{subsec_supp:whiten_impl});  backbone-specific settings (Section \ref{subsec_supp:backbone_impl}).

\subsection{Vote Strength Attenuation}
\label{subsec_supp:vote_strength_attn}

The constant $c$ balances the attenuation of vote strength due to distance and curvature, as shown in (2), and reproduced here for completeness:

\begin{align}
    &V(\boldsymbol{d},\theta)  = e^{-(\frac{s^2 +
     c\kappa^2}{\tau^2})}, \;
     \text{where }
      s =\frac{\theta ||\boldsymbol{d}||}{\sin(\theta)}, \;  \kappa =\frac{2\sin(\theta)}{||\boldsymbol{d}||}
    \label{eq_supp:decay_func}
\end{align} 

We set $c$ by requiring that, when $\theta$ is $45^\circ$, the strength of a vote is 10\% of what it would be if $\theta$ was $0$. This yields:

\begin{equation}
c = -\frac{16\ln(0.1)\tau^2}{\pi^2} \label{eq:c}
\end{equation}

\subsection{Scale of Voting}
\label{subsec_supp:vote_scale}
For each voter, we determine the scale of voting $\tau$ using the median Mahalanobis distance $d'_{med}$ of its $k$ nearest neighbors (where k=100) in the whitened space out of the nearest $2k$ neighbors in Euclidean space. We set the scale such that the strength of the vote will be at 10\% of the  maximum at $d'_{med}
= \text{median}
\left\Vert \mathbf{d'_{k}} \right\Vert$ distance away in whitened space:

\begin{align}
0.1 &= e^{-\frac{ d'^2_{med}}{\tau^2}} \nonumber \\
\tau &= \sqrt{- \frac{d'^2_{med}}{\ln(0.1)}} 
\label{eq_supp:scale_per_pair}
\end{align}

\subsection{Virtual Receivers}
\label{subsec_supp:vir_rec}

We sample the virtual receivers starting from the center of each splat $\mu_i$ along its estimated normal using Tensor Voting as a set of 5 equally spaced points on either side of the $\mu_i$ (totaling 10 virtual receivers) per splat. We use the largest eigenvalue of the scaling matrix $\mathbf{S}$ to determine the spacing between virtual receivers. After the voting process, we choose the virtual, or the actual receiver, that has the highest surface saliency to be the target position $\mathbf{p_{tv}}$ in the TV position loss. This is illustrated in Fig.~\ref{fig_supp:vir_rec_3D}.

As mentioned in Section 3.5 of the main paper, we do not apply the TV position loss to all splats. One reason for this is to reduce the influence of outliers and the other is to avoid using the TV position loss at locations where the surface normal is ambiguous, such as near surface intersections, where multiple normals co-exist.

Specifically, we identify splats whose neighborhoods exhibit high surface normal variability together with high vote strength, 
which indicates the presence of multiple strong surface normals. We compute the surface normal variability according to (\ref{eq:ang_coh}):
\begin{align}
\label{eq:ang_coh}
&\theta_{ij} = \arccos(\mathopen|\boldsymbol{N}_i \cdot\boldsymbol{N}_j|\mathopen) \text{ where } g_j\in \mathcal{N}_i  \nonumber \\
&\theta_{i,rms} = \sqrt{(\sum_{j=0}^J\theta_{ij}^2)/J} 
\end{align}
where $\boldsymbol{N}_i$ and $\boldsymbol{N}_j$ are the surface normals, estimated by Tensor Voting, of splats $g_i$ and $g_j$ respectively, and $\mathcal{N}_i$ is the neighborhood of $g_i$ which contains $J$ neighbors. We compute the vote strength of each splat by taking the root mean square of the strength of the votes it receives.
We then mask out splats that are in the highest $5^{\text{th}}$ percentile of surface normal variability and the highest $5^{\text{th}}$ percentile of vote strength simultaneously.

\input{figs_supp/vir_rec}

\subsection{Whitening Transform}
\label{subsec_supp:whiten_impl}

FatesGS uses 2D splats to represent the scene, with a 2D scaling matrix $\mathbf{S}$. We therefore compute the whitening transform in TV-SGS\textsubscript{F} by setting the third scale equal to 0.1\% of the minimum scale of the 2D Gaussian. This is needed because FatesGS optimizes only 2 scales, whereas a 3D whitening transform is needed for tensor voting.

In addition, we add a small amount of noise to the 3D covariance matrices of the splats equal to 0.01\% of their average trace before computing the whitening transform in all three implementations of TV-SGS. This is to ensure that a degenerate splat will not cause numerical instabilities. 

\subsection{Backbone-Specific Settings}
\label{subsec_supp:backbone_impl}
The TV-SGS\textsubscript{P} implementation follows the same training schedule as the official implementation of PGSR, where the min-scale loss is applied from the beginning of the optimization process, and the single-view and multi-view losses are started after 7000 iterations. The densification and pruning schemes are also kept the same. For the sparse view setting, we only run TV-SGS\textsubscript{P} and PGSR for 15,000 iterations, since we found the improvement in geometry and appearance from 15,000 to 30,000 to be negligible.

The TV-SGS\textsubscript{F} implementation follows the same training schedule as the official implementation of FatesGS, where we train for 15000 iterations. FatesGS uses the depth distortion loss from 2DGS~\cite{Huang_2024_2DGS} starting at iteration 3000, and the rendered normal consistency loss starting at iteration 7000. The RGB loss, cross view feature consistency loss, depth ranking loss, and depth smoothness loss start from the first iteration. The weights for the RGB loss, depth distortion loss, rendered normal consistency loss, cross view feature consistency loss, depth ranking loss and depth smoothness loss are 0.2, 10000 (DTU) and 100 (TnT), 0.05, 1.5, 10.0, 1.0 respectively.

The TV-SGS\textsubscript{V} implementation follows the same training schedule as the official implementation of VGGS and is run for 3000 iterations. The depth loss, single view normal loss, relative depth consistency (RDC) loss and a normal prior loss are started at iteration 1000. The RGB loss and min-scale loss are run from the first iteration. The weights for RGB loss, min-scale loss, single-view normal loss, RDC rank loss, mono-normal prior loss and depth loss are 0.2, 100.0, 0.015, 2.0, 0.045, 0.5 respectively.

As mentioned in the main paper, we start our TV-SGS losses (i.e. TV normal and TV position losses) from the first iteration and vote every 10 iterations with neighborhood updates happening every 100 iterations. We set the weights $w_{tvn}$ and $w_{tvp}$ to 0.5 and 1.0, common to all backbone integrations. All statistics and images generated by the backbones are based on our executions of the authors' code for a fair comparison with the respective TV-SGS implementations, using the same initialization for all methods.

%% file: figs_supp/vir_rec.tex
\begin{figure}[!hb]
  \centering
    \includegraphics[width=0.8\linewidth]{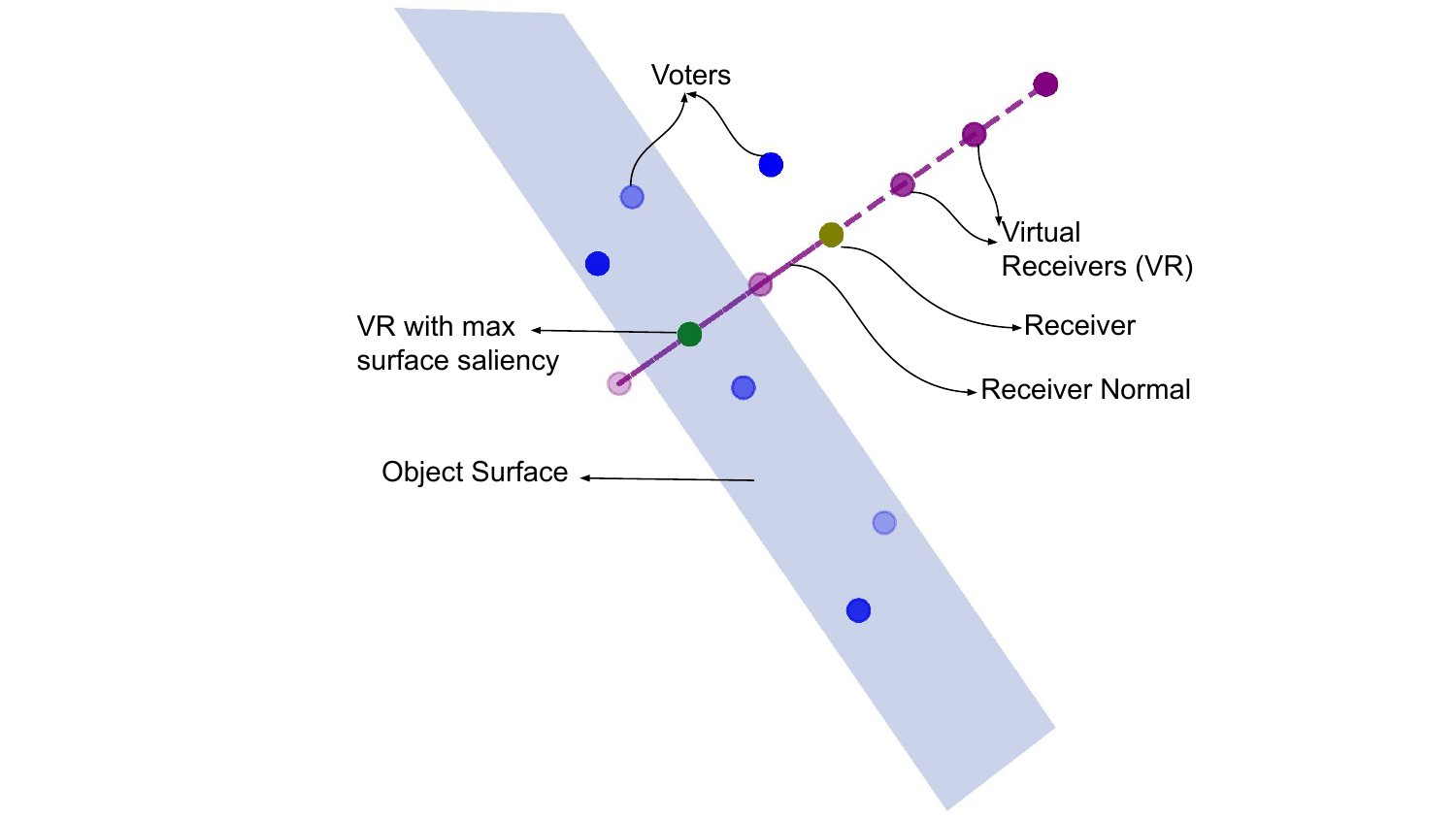}
    \caption{Illustration of virtual receivers in 3D. Virtual points are sampled along a receiver's normal, shown in purple. The virtual point with the highest surface saliency (green point) is chosen, among the real and virtual receivers, as the target position for the position loss $\mathcal{L}_{tvp}$.}
    \label{fig_supp:vir_rec_3D}
\end{figure}

%% file: figs_supp/dtu_small_large_overlap_example.tex
\renewcommand\cellwt{0.30\linewidth} 
\renewcommand\hgapt{0.5pt}             
\renewcommand\vgapt{0.5pt}             

\begin{figure}[t!]
\centering
\setlength{\tabcolsep}{\hgapt}

\begin{tabular}{@{}lccc@{}}
\rotatebox{90}{Large-overlap} &
\includegraphics[width=\cellwt]{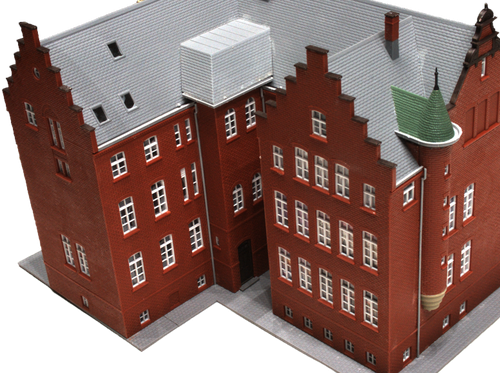} &
\includegraphics[width=\cellwt]{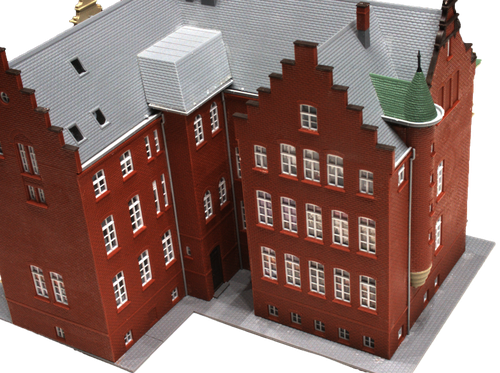} &
\includegraphics[width=\cellwt]{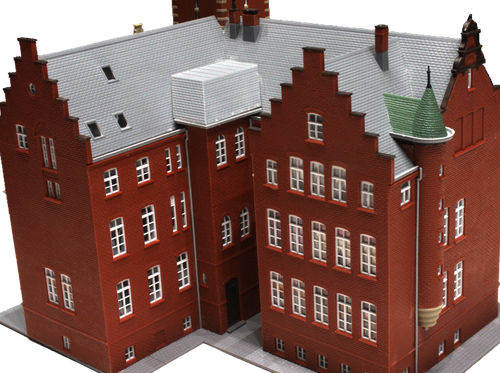} \\[\vgapt]
\rotatebox{90}{Small-overlap} &
\includegraphics[width=\cellwt]{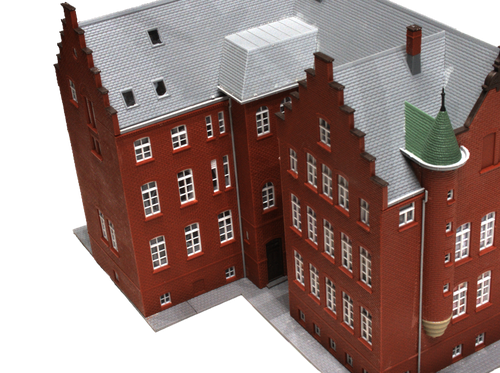} &
\includegraphics[width=\cellwt]{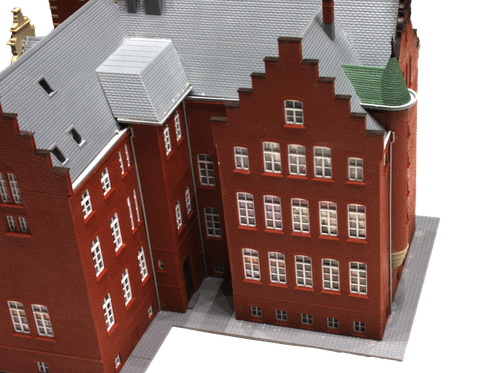} &
\includegraphics[width=\cellwt]{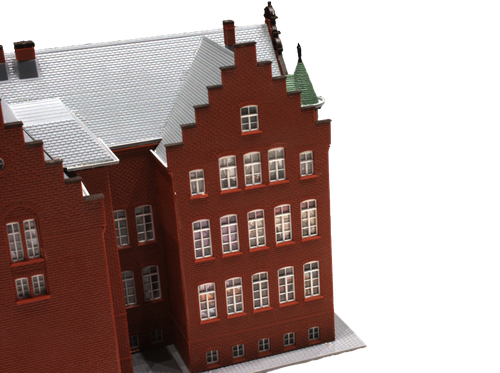}
\end{tabular}

\caption{Ground truth images of Scan24 scene of the large and small-overlap splits for the DTU dataset shown here for illustrating the viewing angles. Notice the small-overlap split has very large baselines.}
\label{fig:dtu_large_small}
\end{figure}

%% file: sec_supp/add_quant_results.tex
\input{tabs_supp/tnt/tab_tnt_20_view_m3_init_splat_centers}
\input{tabs_supp/tnt/tab_tnt_20_view_mvsany_init_mesh}

\section{Additional Quantitative Results}
\label{sec_supp:add_quant_results}
Here we present the per-scene Chamfer distances for the DTU large and small-overlap splits (see Table \ref{tab:dtu_large_small_mesh_merged}) evaluated on the points sampled from the extracted mesh, as an extension of Table 1 of the main paper. Please see Fig.~\ref{fig:dtu_large_small} for an example of the viewing angles of the large and small-overlap splits of the DTU dataset.

We also present statistics on the centers of the splats on the TnT and DTU datasets in addition to the mesh metrics already presented in the main paper (see Tables \ref{tab:tnt_20_views_pcd}, \ref{tab:dtu_small_large_splat_center}). 
These results capture the geometric properties of the splats themselves, before they undergo TSDF fusion, which has drastic effects on the representation of the scene. (Specifically, TSDF fusion begins by rendering depth-maps for the training views from the optimized splats, followed by a volumetric step in which the signed distance is computed for each voxel, before the  marching cubes algorithm extracts the triangle mesh at the zero-level of the distance function.) Measuring Chamfer distances from the splat centers and the provided ground truth allows us to directly measure how well the various GS algorithms reconstruct geometry.

It should be noted that since we are only using splat centers for these evaluations, ignoring their extent, accuracy/precision statistics are more relevant than completeness/recall. Nevertheless, we include both types of statistics in the tables along with the aggregate statistic, F1 for TnT and the average of accuracy and completeness for DTU.

\input{tabs_supp/tnt/tab_tnt_20_view_mvsany_init_pcd}
\input{tabs_supp/dtu/dtu_splat_center_sparse}

Tables \ref{tab:tnt_20_views_mvsany_init_mesh} and \ref{tab:tnt_20_views_mvsany_init_pcd} show the F1 scores for the TnT dataset evaluated on the mesh and splat centers initialized with pointclouds from MVSAnywhere~\cite{izquierdo2025mvsanywhere}. This shows that TV-SGS is robust to different initializations.

%% file: tabs_supp/tnt/tab_tnt_20_view_m3_init_splat_centers.tex
\begin{table}[t!]
  \centering
  \setlength{\tabcolsep}{3pt}
  \resizebox{\columnwidth}{!}{%
  \begin{tabular}{c|c||c||c}
    \hline
     & $\mathbf{P}$ / $\mathbf{R}$ / $\mathbf{F1}$ & $\mathbf{P}$ / $\mathbf{R}$ / $\mathbf{F1}$ & $\mathbf{P}$ / $\mathbf{R}$ / $\mathbf{F1}$\\
    \hline
    \textbf{Scenes} & $\mathbf{PGSR}$ & $\mathbf{FatesGS}$ & $\mathbf{VGGS}$ \\
    \hline
    Barn & 0.113 / 0.018 / 0.032 & 0.317 / 0.058 / 0.098 & 0.241 / 0.075 / 0.115 \\
    Caterpillar & 0.100 / 0.018 / 0.030 & 0.175 / 0.027 / 0.046 & 0.136 / 0.049 / 0.072 \\
    Courthouse & \textbf{0.014} / 0.002 / 0.004 & 0.020 / 0.002 / 0.003 & 0.015 / 0.002 / 0.004 \\
    Ignatius & 0.388 / 0.057 / 0.099 & 0.408 / 0.053 / 0.093 & 0.253 / 0.080 / 0.121 \\
    Meetingroom & 0.114 / 0.013 / 0.024 & 0.125 / 0.014 / 0.025 & 0.218 / 0.041 / 0.069 \\
    Truck & 0.406 / 0.095 / 0.154 & 0.406 / 0.090 / 0.148 & 0.361 / 0.168 / 0.229 \\
    \hline
    Average & 0.189 / 0.034 / 0.057 & 0.242 / 0.041 / 0.069 & 0.204 / 0.069 / 0.102 \\
    \hline
     & $\mathbf{TV-SGS}_{\mathbf{P}}$ & $\mathbf{TV-SGS}_{\mathbf{F}}$ & $\mathbf{TV-SGS}_{\mathbf{V}}$ \\
     \hline
    Barn & \textbf{0.153} / \textbf{0.027} / \textbf{0.045} & \textbf{0.386} / \textbf{0.073} / \textbf{0.123} & 0.282 / 0.093 / 0.140 \\
    Caterpillar & \textbf{0.161} / \textbf{0.029} / \textbf{0.049} & \textbf{0.182} / \textbf{0.029} / \textbf{0.050} & 0.136 / 0.051 / 0.074 \\
    Courthouse & 0.013 / \textbf{0.002} / \textbf{0.004} & \textbf{0.021} / \textbf{0.002} / \textbf{0.003} & 0.015 / 0.002 / 0.004 \\
    Ignatius & \textbf{0.434} / \textbf{0.065} / \textbf{0.113} & \textbf{0.411} / \textbf{0.055} / \textbf{0.097} & \textbf{0.283} / \textbf{0.088} / \textbf{0.135} \\
    Meetingroom & \textbf{0.131} / \textbf{0.016} / \textbf{0.028} & \textbf{0.128} / \textbf{0.015} / \textbf{0.026} & \textbf{0.235} / \textbf{0.045} / \textbf{0.076} \\
    Truck & \textbf{0.434} / \textbf{0.108} / \textbf{0.173} & \textbf{0.449} / \textbf{0.105} / \textbf{0.170} & \textbf{0.377} / \textbf{0.182} / \textbf{0.245} \\
    \hline
    Average & \textbf{0.221} / \textbf{0.041} / \textbf{0.069} & \textbf{0.263} / \textbf{0.046} / \textbf{0.078} & \textbf{0.221} / \textbf{0.077} / \textbf{0.112} \\
    \hline
  \end{tabular}
  }
  \caption{Quantitative results on geometry evaluated directly on the \textbf{splat centers} on the \textbf{TnT} dataset using 20 training views, using \textbf{Mast3r}~\cite{duisterhof2025mast3r} pointcloud initialization. P/R/F1 refers to precision, recall and F1 scores. Note the precision improvement of the splat centers.}
  \label{tab:tnt_20_views_pcd}
\end{table}

%% file: tabs_supp/tnt/tab_tnt_20_view_mvsany_init_mesh.tex
\begin{table}[t!]
  \centering
  \setlength{\tabcolsep}{3pt}
  \resizebox{\columnwidth}{!}{%
  \begin{tabular}{c|c||c||c}
    \hline
     & $\mathbf{P}$ / $\mathbf{R}$ / $\mathbf{F1}$ & $\mathbf{P}$ / $\mathbf{R}$ / $\mathbf{F1}$ & $\mathbf{P}$ / $\mathbf{R}$ / $\mathbf{F1}$\\
    \hline
    \textbf{Scenes} & $\mathbf{PGSR}$ & $\mathbf{FatesGS}$ & $\mathbf{VGGS}$ \\
    \hline
    Barn & 0.210 / 0.156 / 0.179 & 0.357 / 0.346 / 0.352 & 0.193 / 0.177 / 0.185 \\
    Caterpillar & 0.196 / 0.280 / 0.231 & 0.147 / 0.225 / 0.178 & 0.114 / 0.179 / 0.139 \\
    Courthouse & 0.034 / 0.006 / 0.011 & 0.107 / 0.055 / 0.072 & 0.083 / 0.086 / 0.085 \\
    Ignatius & 0.552 / 0.582 / 0.567 & 0.484 / 0.510 / 0.497 & 0.249 / 0.268 / 0.258 \\
    Meetingroom & 0.066 / 0.008 / 0.014 & 0.043 / 0.013 / 0.020 & 0.145 / 0.067 / 0.091 \\
    Truck & 0.413 / 0.455 / 0.433 & 0.340 / 0.374 / 0.356 & 0.273 / 0.316 / 0.293 \\
    \hline
    Average & 0.245 / 0.248 / 0.239 & 0.246 / 0.254 / 0.246 & 0.176 / 0.182 / 0.175 \\
    \hline
     & TV-SGS\_P & TV-SGS\_F & TV-SGS\_V \\
     \hline 
    Barn & 0.244 / 0.187 / 0.212 & 0.368 / 0.367 / 0.368 & 0.230 / 0.208 / 0.219 \\
    Caterpillar & 0.200 / 0.298 / 0.239 & 0.166 / 0.244 / 0.198 & 0.106 / 0.155 / 0.126 \\
    Courthouse & 0.079 / 0.013 / 0.022 & 0.157 / 0.072 / 0.098 & 0.114 / 0.085 / 0.097 \\
    Ignatius & 0.537 / 0.563 / 0.549 & 0.488 / 0.524 / 0.505 & 0.273 / 0.288 / 0.280 \\
    Meetingroom & 0.070 / 0.014 / 0.023 & 0.053 / 0.016 / 0.025 & 0.122 / 0.065 / 0.085 \\
    Truck & 0.453 / 0.476 / 0.464 & 0.387 / 0.415 / 0.400 & 0.247 / 0.307 / 0.274 \\
    \hline
    Average & 0.264 / 0.258 / 0.252 & 0.270 / 0.273 / 0.266 & 0.182 / 0.185 / 0.180 \\
    \hline
  \end{tabular}
  }
  \caption{Quantitative results on geometry evaluated directly on the \textbf{mesh} on the \textbf{TnT} dataset using 20 training views, using \textbf{MVSAnywhere}~\cite{izquierdo2025mvsanywhere} pointcloud initialization. P/R/F1 refers to precision, recall and F1 scores. TV-SGS improves upon all three backbones.}
  \label{tab:tnt_20_views_mvsany_init_mesh}
\end{table}

%% file: tabs_supp/tnt/tab_tnt_20_view_mvsany_init_pcd.tex
\begin{table}[h!]
  \centering
  \setlength{\tabcolsep}{3pt}
  \resizebox{\columnwidth}{!}{%
  \begin{tabular}{c|c||c||c}
    \hline
     & $\mathbf{P}$ / $\mathbf{R}$ / $\mathbf{F1}$ & $\mathbf{P}$ / $\mathbf{R}$ / $\mathbf{F1}$ & $\mathbf{P}$ / $\mathbf{R}$ / $\mathbf{F1}$\\
    \hline
    \textbf{Scenes} & $\mathbf{PGSR}$ & $\mathbf{FatesGS}$ & $\mathbf{VGGS}$ \\
    \hline
    Barn & 0.118 / 0.018 / 0.032 & 0.340 / 0.040 / 0.072 & 0.148 / 0.023 / 0.040 \\
    Caterpillar & 0.177 / 0.030 / 0.051 & 0.196 / 0.026 / 0.046 & 0.093 / 0.020 / 0.032 \\
    Courthouse & 0.023 / 0.004 / 0.006 & 0.191 / 0.014 / 0.026 & 0.158 / 0.017 / 0.030 \\
    Ignatius & 0.399 / 0.052 / 0.093 & 0.409 / 0.041 / 0.075 & 0.235 / 0.040 / 0.068 \\
    Meetingroom & 0.055 / 0.006 / 0.011 & 0.052 / 0.005 / 0.009 & 0.124 / 0.018 / 0.031 \\
    Truck & 0.375 / 0.086 / 0.139 & 0.415 / 0.063 / 0.109 & 0.238 / 0.053 / 0.087 \\
    \hline
    Average & 0.191 / 0.033 / 0.055 & 0.267 / 0.032 / 0.056 & 0.166 / 0.028 / 0.048 \\
    \hline
     & \textbf{TV-SGS\_P} & \textbf{TV-SGS\_F} & \textbf{TV-SGS\_V} \\
     \hline
    Barn & 0.139 / 0.022 / 0.038 & 0.424 / 0.050 / 0.090 & 0.170 / 0.028 / 0.049 \\
    Caterpillar & 0.164 / 0.030 / 0.051 & 0.203 / 0.026 / 0.046 & 0.103 / 0.022 / 0.037 \\
    Courthouse & 0.027 / 0.004 / 0.007 & 0.208 / 0.015 / 0.028 & 0.165 / 0.017 / 0.030 \\
    Ignatius & 0.412 / 0.057 / 0.100 & 0.422 / 0.042 / 0.076 & 0.255 / 0.044 / 0.075 \\
    Meetingroom & 0.056 / 0.007 / 0.013 & 0.054 / 0.005 / 0.009 & 0.137 / 0.020 / 0.034 \\
    Truck & 0.422 / 0.098 / 0.159 & 0.430 / 0.068 / 0.117 & 0.238 / 0.059 / 0.094 \\
    \hline
    Average & 0.203 / 0.037 / 0.062 & 0.290 / 0.034 / 0.061 & 0.178 / 0.032 / 0.053 \\
    \hline
  \end{tabular}
  }
  \caption{Quantitative results on geometry evaluated directly on the \textbf{splat centers} on the \textbf{TnT} dataset using 20 training views, using \textbf{MVSAnywhere}~\cite{izquierdo2025mvsanywhere} pointcloud initialization. P/R/F1 refers to precision, recall and F1 scores.  Note the precision improvement of the splat centers.}
  \label{tab:tnt_20_views_mvsany_init_pcd}
\end{table}

%% file: tabs_supp/dtu/dtu_splat_center_sparse.tex
\begin{table}[htbp]
  \centering
  \label{tab:mytable}
  \begin{tabular}{c|c||c}
  \toprule
    {} & \textbf{Small-overlap} & \textbf{Large-overlap} \\
    \hline
    \textbf{Method} & \textbf{A / C / CD} & \textbf{A / C / CD} \\
    \hline
    $\mathrm{PGSR}$ & 2.37 / 2.29 / 2.33 & 1.41 / 1.80 / 1.61 \\
    $\mathbf{TV-SGS}_{\mathbf{P}}$ & 1.68 / 2.09 / 1.88 & 1.11 / 1.68 / 1.40 \\
    \hline
    $\mathrm{FatesGS}$ & 1.34 / 1.26 / 1.30 & 1.08 / 1.45 / 1.26 \\
    $\mathbf{TV-SGS}_{\mathbf{F}}$ & 1.36 / 1.21 / 1.29 & 1.08 / 1.37 / 1.22 \\
    \hline
    $\mathrm{VGGS}$ & 1.40 / 1.34 / 1.37 & 1.13 / 1.56 / 1.34 \\
    $\mathbf{TV-SGS}_{\mathbf{V}}$ & 1.18 / 1.30 / 1.24 & 1.03 / 1.52 / 1.28 \\
    \hline
  \end{tabular}
  \caption{Geometric evaluation on the \textbf{splat centers} for \textbf{DTU} using three views with \textbf{small} and \textbf{large} overlap, using \textbf{Mast3r}~\cite{duisterhof2025mast3r} initialization. We report the average accuracy, average completeness and average Chamfer distances over all the scenes for compactness. A / C / CD stands for average accuracy / average completeness / average Chamfer distance.}
  \label{tab:dtu_small_large_splat_center}
\end{table}

%% file: sec_supp/add_qual_results.tex
\section{Additional Qualitative Results}
\label{sec_supp:add_qual_results}

In this section, we show how the TV position loss helps move floaters with the precision heatmap of the splat centers, as shown in Fig.~\ref{fig_supp:TnT_ptcloud_qualitative}. The splat centers are color-coded according to their precision with respect to ground truth pointcloud, where brighter colors mean higher precision. We show a zoomed in view of a surface patch to illustrate the effects of the TV position loss guiding the floaters towards the most salient surface.

We also show qualitative comparisons of rendered RGB images from held out test views and also the meshes from the same test views in Figures \ref{fig_supp:TnT_compare_qualitative} and \ref{fig_supp:TnT_compare_qualitative_mesh} for the TnT dataset and Figures \ref{fig_supp:dtu_compare_qualitative} and \ref{fig_supp:dtu_compare_qualitative_mesh} for the DTU dataset, illustrating that TV-SGS does not sacrifice rendering quality to obtain better geometry.

\input{tabs_supp/dtu/dtu_large_small_mesh_merged}

\clearpage

\input{figs_supp/tnt_splat_prec}
\input{figs_supp/tnt_compare_supp}
\input{figs_supp/tnt_compare_mesh}
\input{figs_supp/dtu_compare_supp}
\input{figs_supp/dtu_compare_mesh_supp}

%% file: tabs_supp/dtu/dtu_large_small_mesh_merged.tex
\begin{table*}[ht!]
  \centering
  \setlength{\tabcolsep}{5pt}
  \begin{tabular}{@{}c@{}}
  \resizebox{\linewidth}{!}{%
  \begin{tabular}{@{}lccccccccccccccccc@{}}
  \toprule
  Scan ID & 24 & 37 & 40 & 55 & 63 & 65 & 69 & 83 & 97 & 105 & 106 & 110 & 114 & 118 & 122 & Mean CD $\downarrow$ \\
    \midrule
    PGSR \cite{Chen_2024_PGSR} & 1.40 & 3.61 & 1.67 & 1.54 & 4.12 & 1.87 & 1.32 & 1.94 & 1.63 & 1.06 & 1.34 & 0.90 & \textbf{0.72} & 1.40 & \textbf{1.60} & 1.74 \\
    $\mathbf{TV-SGS}_{\mathbf{P}}$ & \textbf{1.36} & \textbf{3.56} & \textbf{1.33} & \textbf{1.41} & \textbf{3.69} & \textbf{1.74} & \textbf{1.12} & \textbf{1.71} & \textbf{1.44} & \textbf{1.02} & \textbf{1.13} & \textbf{0.80} & 0.74 & \textbf{1.28} & 1.65 & \textbf{1.60} \\
    \midrule
    $\mathrm{SparseNeuS_{ft}}$ \cite{long2022sparseneus} & 1.29 & 2.27 & 1.57 & 0.88 & 1.61 & 1.86 & 1.06 & 1.27 & 1.42 & 1.07 & 0.99 & 0.87 & 0.54 & 1.15 & 1.18 & 1.27 \\
    $\mathrm{VolRecon}$ \cite{Ren_2023_VolRecon} & 1.20 & 2.59 & 1.56 & 1.08 & 1.43 & 1.92 & 1.11 & 1.48 & 1.42 & 1.05 & 1.19 & 1.38 & 0.74 & 1.23 & 1.27 & 1.38 \\
    $\mathrm{GenS_{ft}}$ \cite{peng2023gens} & 0.91 & 2.33 & 1.46 & 0.75 & 1.02 & 1.58 & 0.74 & 1.16 & 1.05 & 0.77 & 0.88 & 0.56 & 0.49 & 0.78 & 0.93 & 1.03 \\
    $\mathrm{ReTR}$ \cite{liang2023retr} & 1.05 & 2.31 & 1.44 & 0.98 & 1.18 & 1.52 & 0.88 & 1.35 & 1.30 & 0.87 & 1.07 & 0.77 & 0.59 & 1.05 & 1.12 & 1.17 \\
    $\mathrm{UFORecon}$ \cite{na2024uforecon} & 0.76 & 2.05 & 1.31 & 0.82 & 1.12 & 1.18 & 0.74 & 1.17 & 1.11 & 0.71 & 0.88 & 0.58 & 0.54 & 0.86 & 0.99 & 0.99 \\
    \midrule
    $\mathrm{MonoSDF}$ \cite{yu2022monosdf} & 2.85 & 3.91 & 2.26 & 1.22 & 3.37 & 1.95 & 1.95 & 5.53 & 5.77 & 1.10 & 5.99 & 2.28 & 0.65 & 2.65 & 2.44 & 2.93 \\
    $\mathrm{NeuSurf}$ \cite{huang2024neusurf} & 0.78 & 2.35 & 1.55 & 0.75 & 1.04 & 1.68 & 0.60 & 1.14 & 0.98 & 0.70 & 0.74 & 0.49 & 0.39 & 0.75 & 0.86 & 0.99 \\
    $\mathrm{SparseCraft}$ \cite{younes2024sparsecraft} & 1.17 & 1.74 & 1.80 & 0.70 & 1.19 & 1.53 & 0.83 & 1.05 & 1.42 & 0.78 & 0.80 & 0.56 & 0.44 & 0.77 & 0.84 & 1.04 \\
    \hline
    $\mathrm{Sparse2DGS}$ \cite{wu2025sparse2dgs} & 1.05 & 2.35 & 1.38 & 0.83 & 1.37 & 1.45 & 0.84 & 1.16 & 1.43 & 0.74 & 0.85 & 0.84 & 0.57 & 0.95 & 1.01 & 1.13 \\
    $\mathrm{FatesGS}$ \cite{huang2025fatesgs} & 0.71 & 2.18 & 1.17 & 0.80 & \textbf{1.05} & 1.33 & 0.65 & 1.15 & 1.05 & \textbf{0.66} & 0.80 & 0.50 & 0.41 & \textbf{0.76} & 0.96 & 0.95 \\
    $\mathbf{TV-SGS}_{\mathbf{F}}$ & \textbf{0.69} & \textbf{2.10} & \textbf{1.15} & \textbf{0.79} & 1.07 & \textbf{1.25} & \textbf{0.64} & \textbf{1.14} & \textbf{1.02} & 0.67 & \textbf{0.78} & \textbf{0.49} & \textbf{0.40} & 0.77 & \textbf{0.95} & \textbf{0.92} \\
    $\mathrm{VGGS}$ \cite{xiang2026vggs} & \textbf{0.90} & 2.08 & 1.49 & 1.07 & 2.30 & \textbf{1.43} & \textbf{0.92} & 1.44 & 1.81 & 1.11 & 0.98 & \textbf{0.81} & \textbf{0.56} & 1.25 & \textbf{1.14} & 1.29 \\
    $\mathbf{TV-SGS}_{\mathbf{V}}$ & 0.99 & \textbf{2.07} & \textbf{1.44} & \textbf{1.06} & \textbf{1.51} & 1.44 & 0.95 & \textbf{1.31} & \textbf{1.48} & \textbf{1.08} & \textbf{0.93} & \textbf{0.81} & 0.57 & \textbf{1.17} & 1.15 & \textbf{1.20} \\
    \bottomrule
  \end{tabular}
  } \\
  \vspace{2pt}
  {\small (a) Geometric evaluation on \textbf{DTU} using three views with \textbf{large} overlap, using \textbf{Mast3r}~\cite{duisterhof2025mast3r} initialization, evaluated on the extracted \textbf{mesh}.\par} \\
  \vspace{10pt}
  \resizebox{\linewidth}{!}{%
  \begin{tabular}{@{}lccccccccccccccccc@{}}
  \toprule
    Scan ID & 24 & 37 & 40 & 55 & 63 & 65 & 69 & 83 & 97 & 105 & 106 & 110 & 114 & 118 & 122 & Mean CD $\downarrow$ \\
    \midrule
    $\mathrm{PGSR}$\cite{Chen_2024_PGSR} & 2.47 & 4.41 & 3.53 & 1.63 & 2.86 & 2.97 & 1.76 & 2.43 & 2.36 & 2.22 & 2.62 & 2.60 & \textbf{1.36} & 2.43 & 2.77 & 2.56 \\
    $\mathbf{TV-SGS}_{\mathbf{P}}$ & \textbf{2.00} & \textbf{3.34} & \textbf{2.76} & \textbf{1.62} & \textbf{2.75} & \textbf{2.30} & \textbf{1.58} & \textbf{2.10} & \textbf{2.18} & \textbf{1.96} & \textbf{2.13} & \textbf{2.32} & 1.37 & \textbf{2.13} & \textbf{2.14} & \textbf{2.18} \\
    \midrule
    $\mathrm{MonoSDF}$ \cite{yu2022monosdf} & 3.47 & 3.61 & 2.10 & 1.05 & 2.37 & 1.38 & 1.41 & 1.85 & 1.74 & 1.10 & 1.46 & 2.28 & 1.25 & 1.44 & 1.45 & 1.86 \\
    $\mathrm{NeuSurf}$ \cite{huang2024neusurf} & 1.35 & 3.25 & 2.50 & 0.80 & 1.21 & 2.35 & 0.77 & 1.19 & 1.20 & 1.05 & 1.05 & 1.21 & 0.41 & 0.80 & 1.08 & 1.35 \\
    \hline
    $\mathrm{FatesGS}$ \cite{huang2025fatesgs} & \textbf{0.68} & 2.56 & \textbf{1.45} & \textbf{0.70} & 1.09 & 1.80 & 0.89 & \textbf{1.20} & 1.07 & 0.77 & \textbf{0.90} & 0.85 & 0.47 & 0.78 & 0.83 & 1.07\\
    $\mathbf{TV-SGS}_{\mathbf{F}}$ & \textbf{0.68} & \textbf{2.35} & 1.48 & \textbf{0.70} & \textbf{1.04} & \textbf{1.70} & \textbf{0.85} & 1.25 & \textbf{0.96} & \textbf{0.73} & 0.93 & \textbf{0.78} & \textbf{0.45} & \textbf{0.77} & \textbf{0.81} & \textbf{1.03} \\
    $\mathrm{VGGS}$ \cite{xiang2026vggs} & 0.73 & 1.71 & \textbf{1.02} & 0.73 & 1.56 & \textbf{1.21} & \textbf{0.71} & 1.28 & 0.92 & 0.78 & \textbf{0.79} & 2.23 & \textbf{0.46} & \textbf{0.80} & \textbf{0.90} & 1.06 \\
    $\mathbf{TV-SGS}_{\mathbf{V}}$ & \textbf{0.69} & \textbf{1.33} & 1.05 & \textbf{0.69} & \textbf{1.28} & 1.32 & 0.73 & \textbf{1.24} & \textbf{0.89} & \textbf{0.75} & 0.84 & \textbf{2.08} & \textbf{0.46} & 0.81 & 0.91 & \textbf{1.00}  \\
    \bottomrule
  \end{tabular}
  } \\
  \vspace{2pt}
  {\small (b) Geometric evaluation on \textbf{DTU} using three views with \textbf{small} overlap, using \textbf{Mast3r}~\cite{duisterhof2025mast3r} initialization, evaluated on the extracted \textbf{mesh}.\par}
  \end{tabular}
  \caption{The methods are divided into four categories for the large-overlap setting, from top to bottom: (1) dense-view reconstruction methods related to TV-SGS (but trained on three views), (2)  generalizable sparse-view reconstruction methods (not present in the lower table) (3) Implicit sparse-view inference time optimization methods, and (4) Explicit sparse-view inference time optimization methods. We mark with boldface the best result as compared to the respective backbones. The values of all methods are taken from the respective papers, except the backbones used by TV-SGS, for which we executed the authors’ code.}
  \label{tab:dtu_large_small_mesh_merged}
  \end{table*}

%% file: figs_supp/tnt_splat_prec.tex
\newcommand\quartlwrend{0.29\linewidth}
\renewcommand\rotvert{0.45cm}
\clearpage
\begin{figure*}[tb]
    \centering
    \begin{tabular}{@{}lccc@{}}
        {} & FatesGS & TV-SGS\textsubscript{F} & TV-SGS\textsubscript{F} whiten + $\mathcal{L}_{tvn}$ only \\
        \rotatebox{90}{\hspace{\rotvert} Barn zoom out} &
        \includegraphics[width=\quartlwrend]{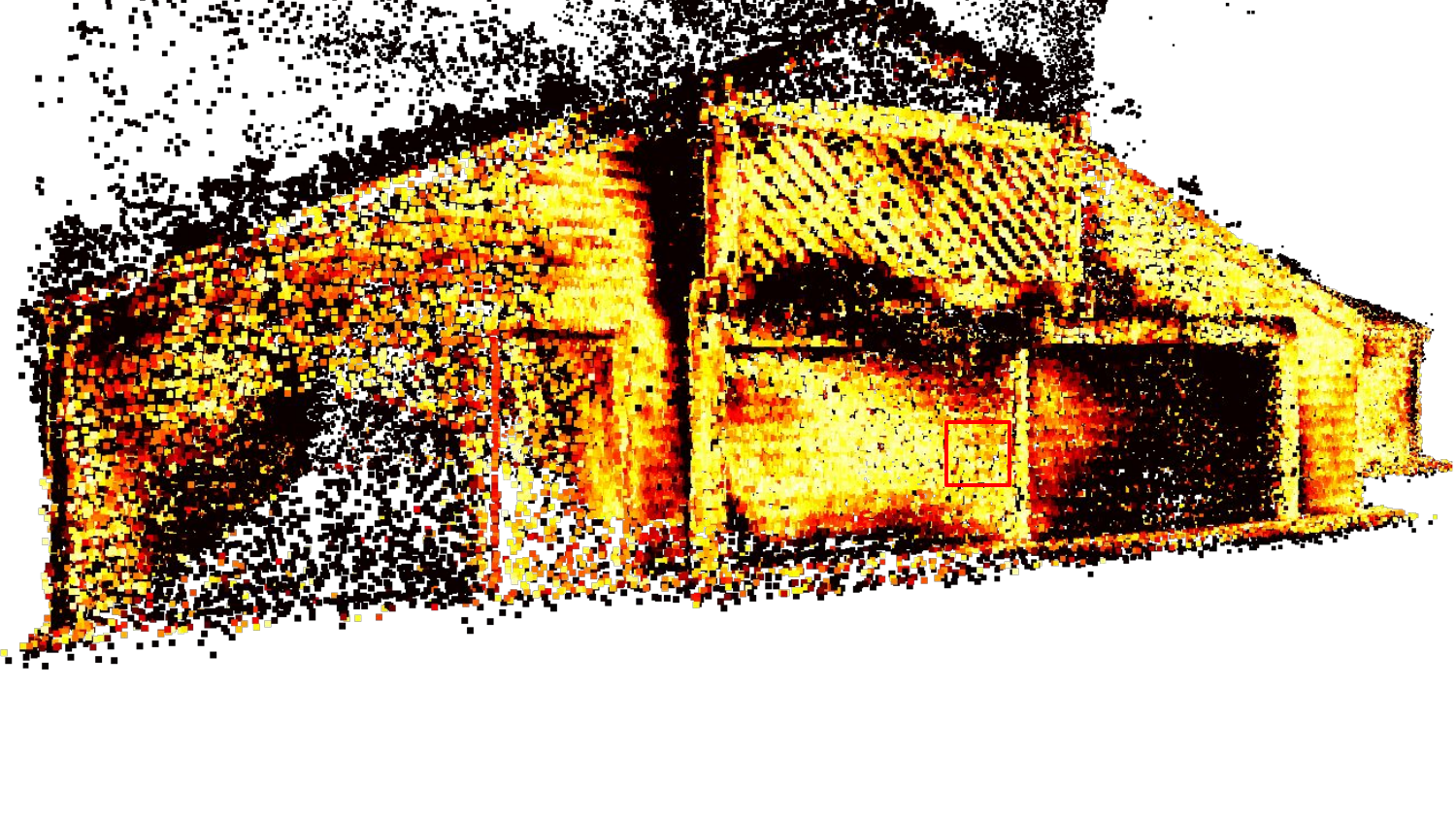} &
        \includegraphics[width=\quartlwrend]{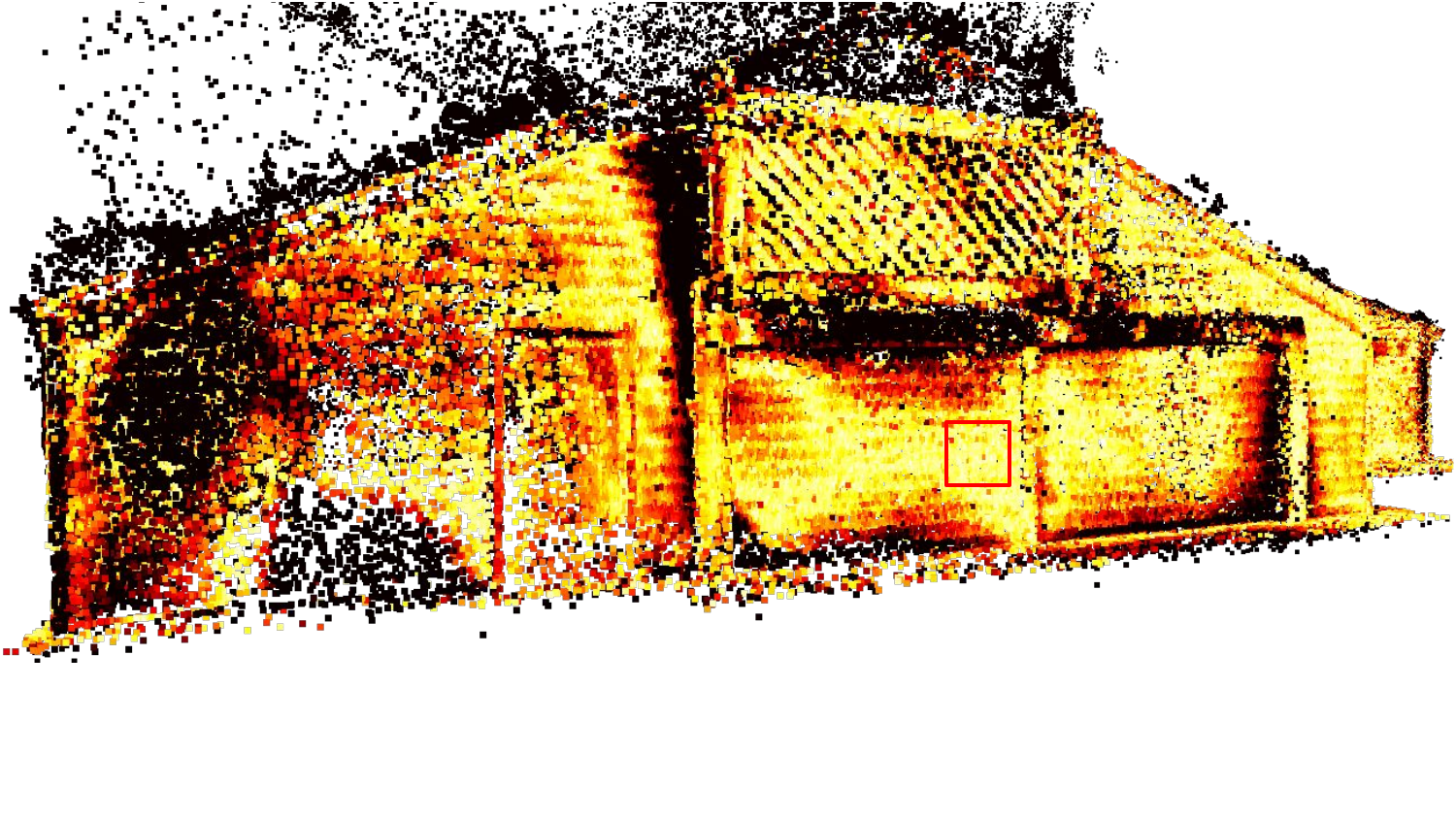} &
        \includegraphics[width=\quartlwrend]{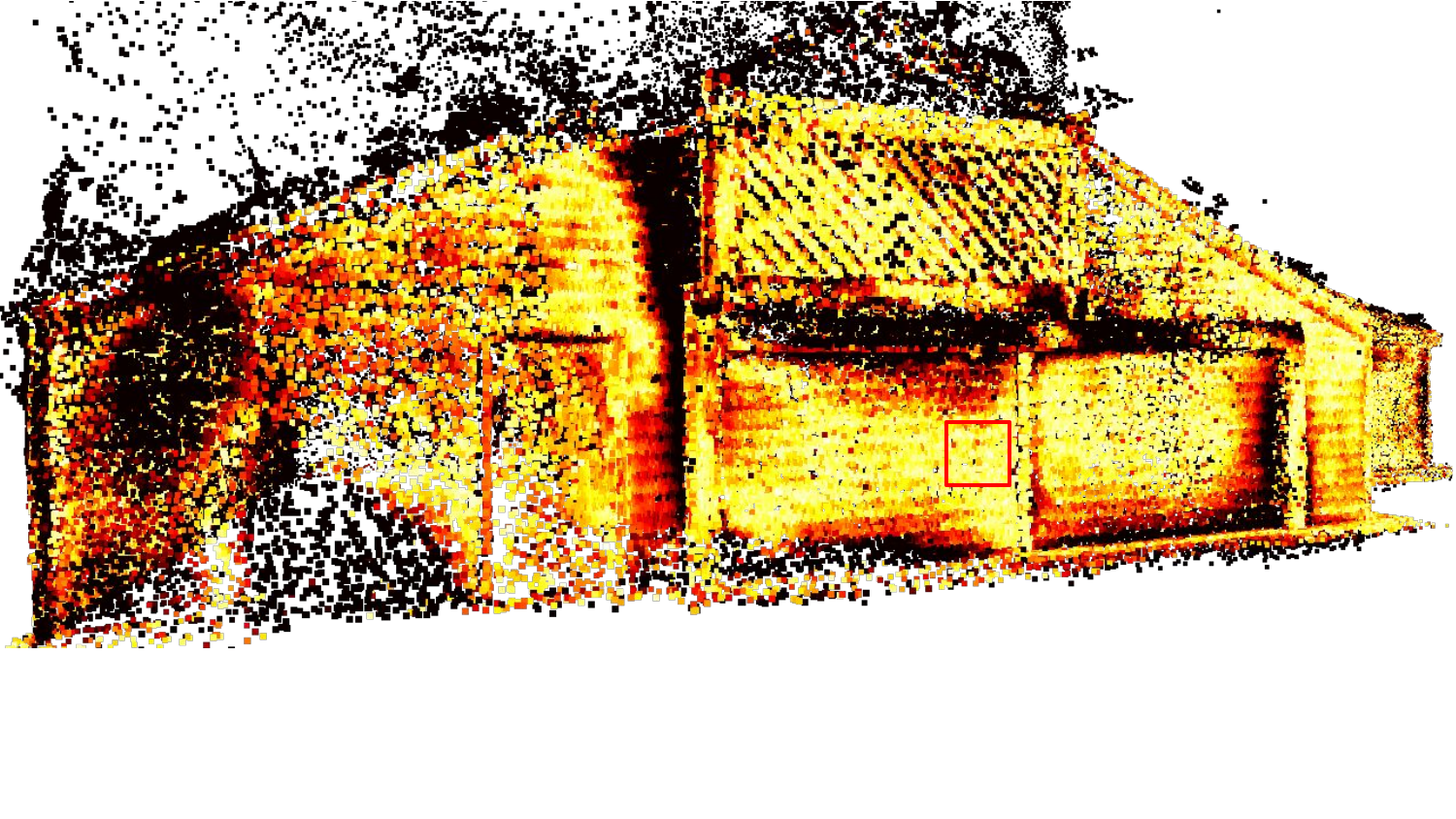} \\
        \rotatebox{90}{\hspace{\rotvert} Barn zoom in} &
        \includegraphics[width=\quartlwrend]{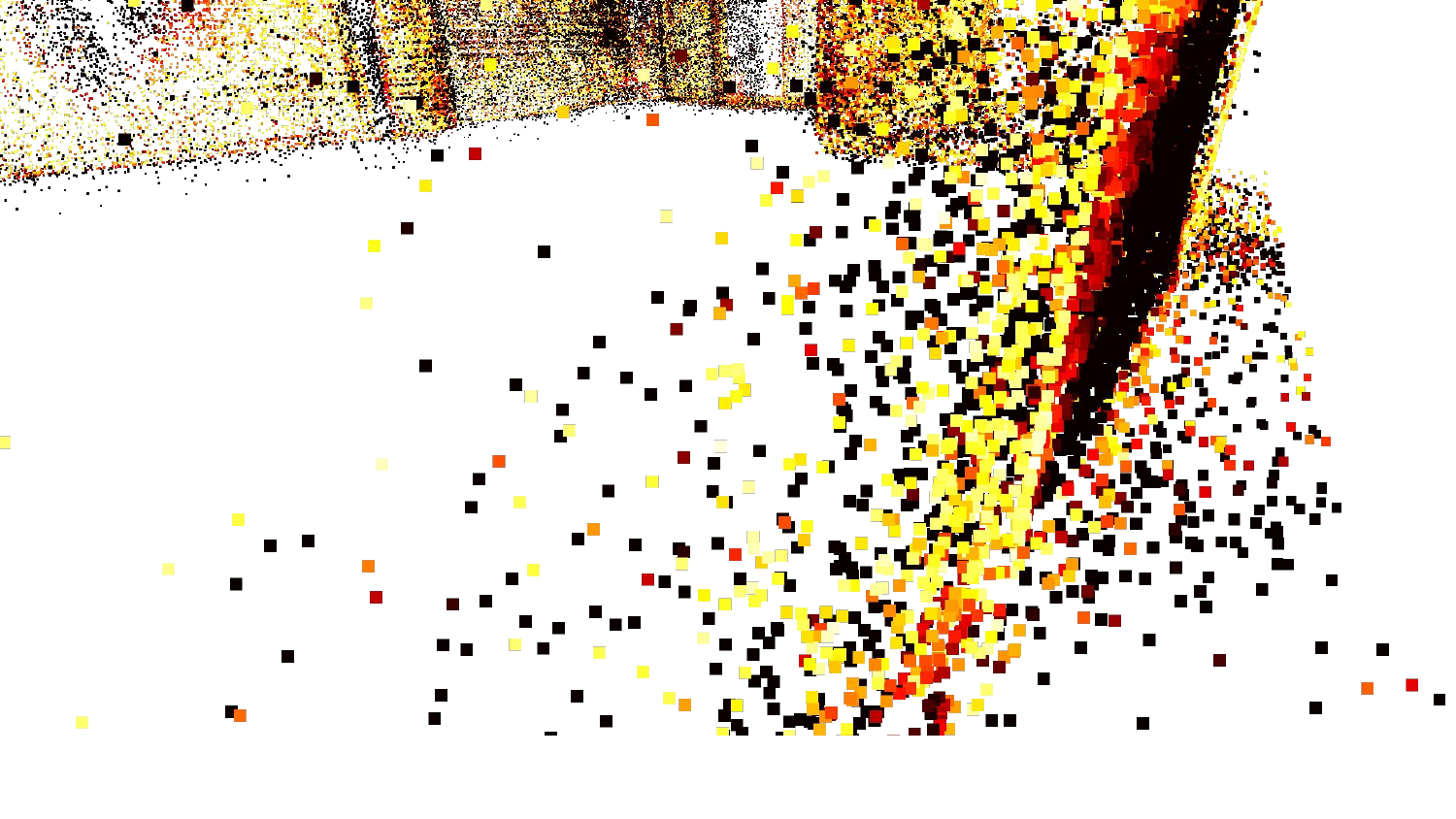} &
        \includegraphics[width=\quartlwrend]{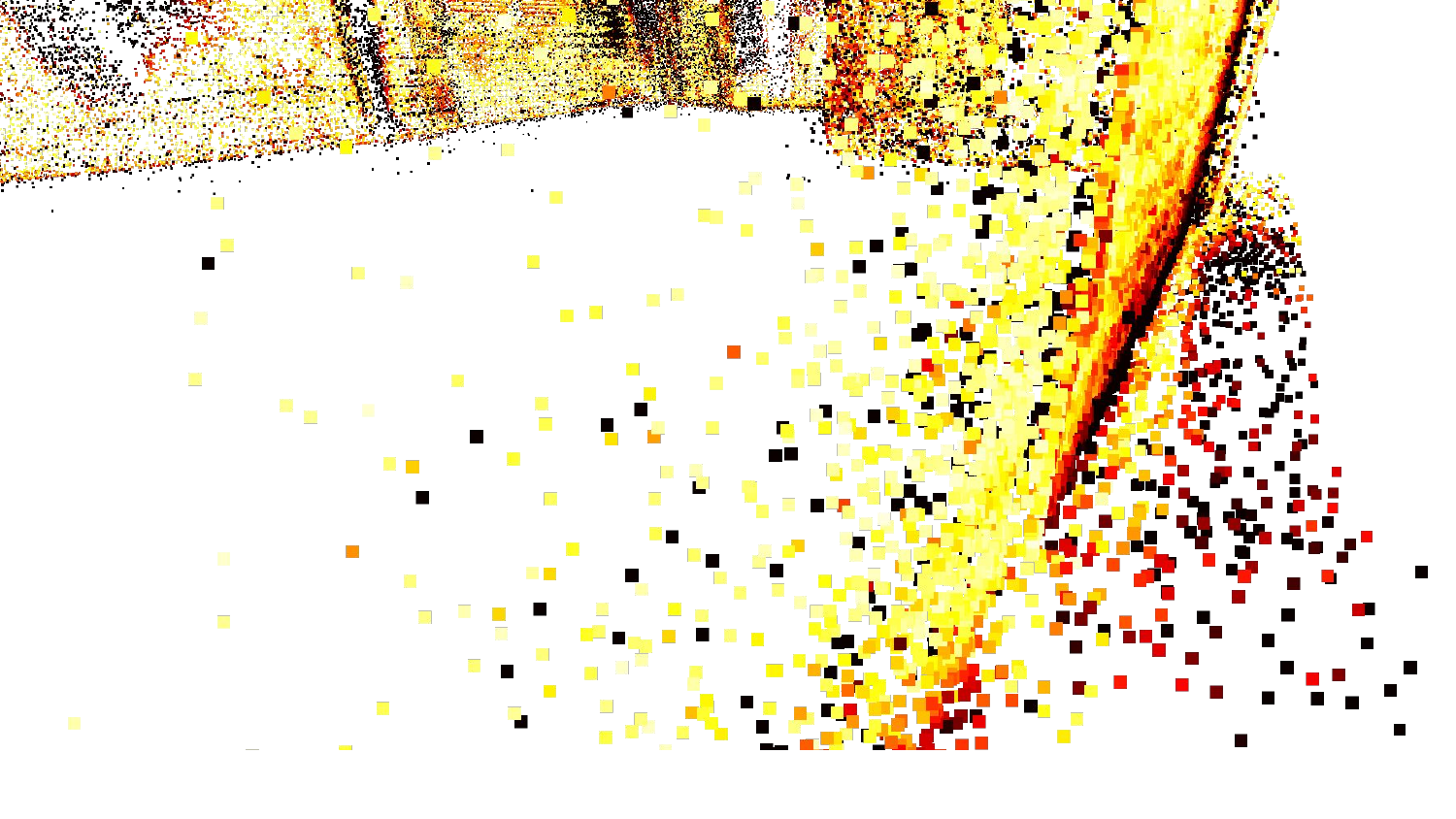} &
        \includegraphics[width=\quartlwrend]{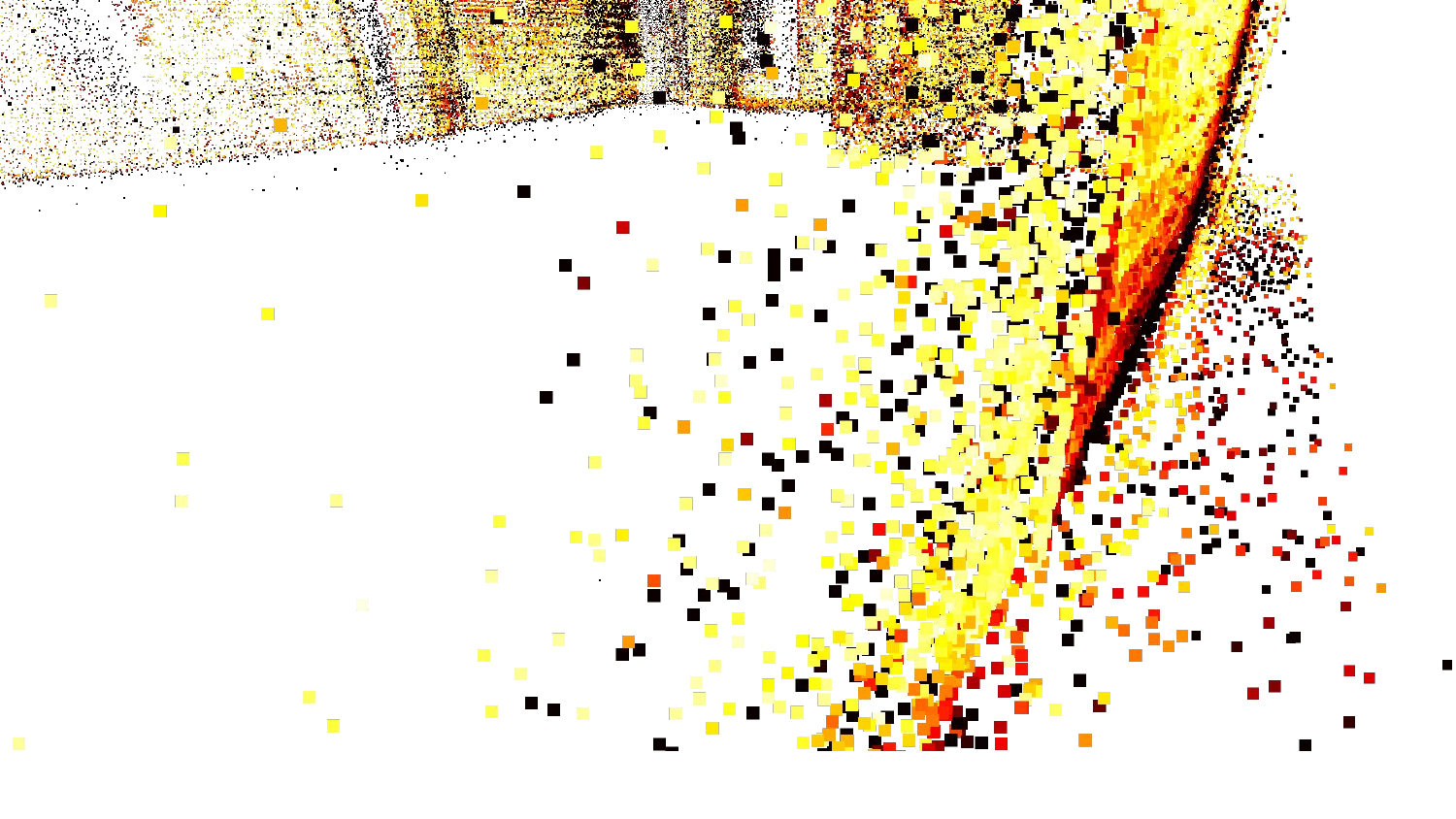} \\

        \rotatebox{90}{\hspace{0.30cm} Barn prec-recall curves} &
        \includegraphics[width=\quartlwrend]{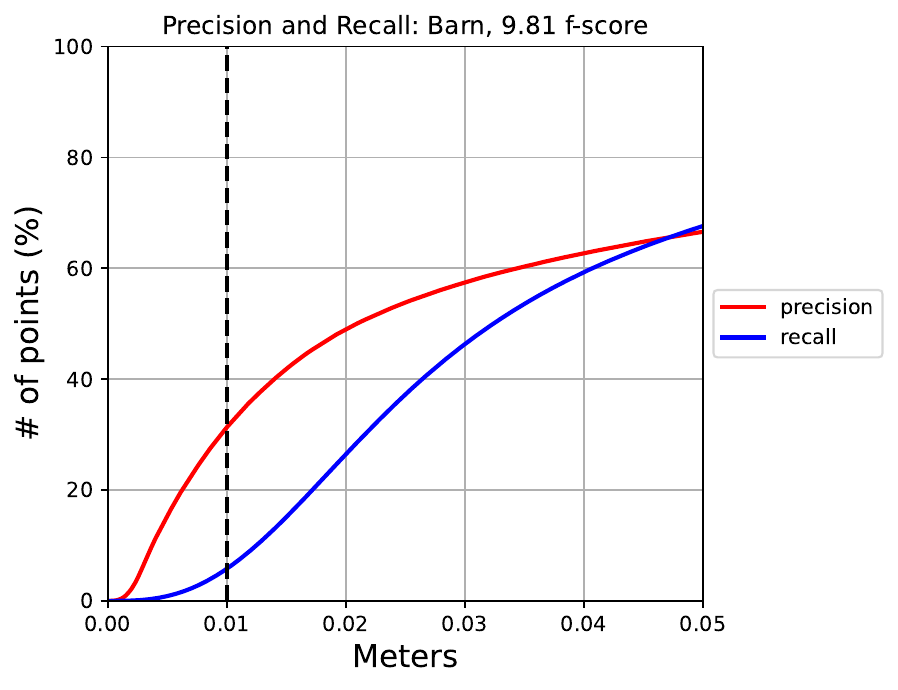} &
        \includegraphics[width=\quartlwrend]{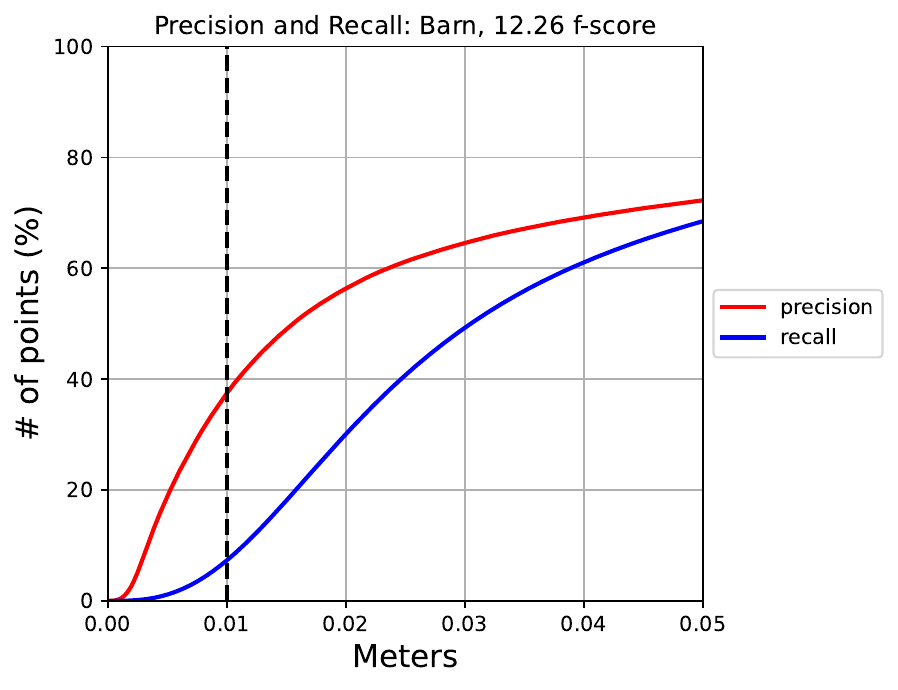} &
        \includegraphics[width=\quartlwrend]{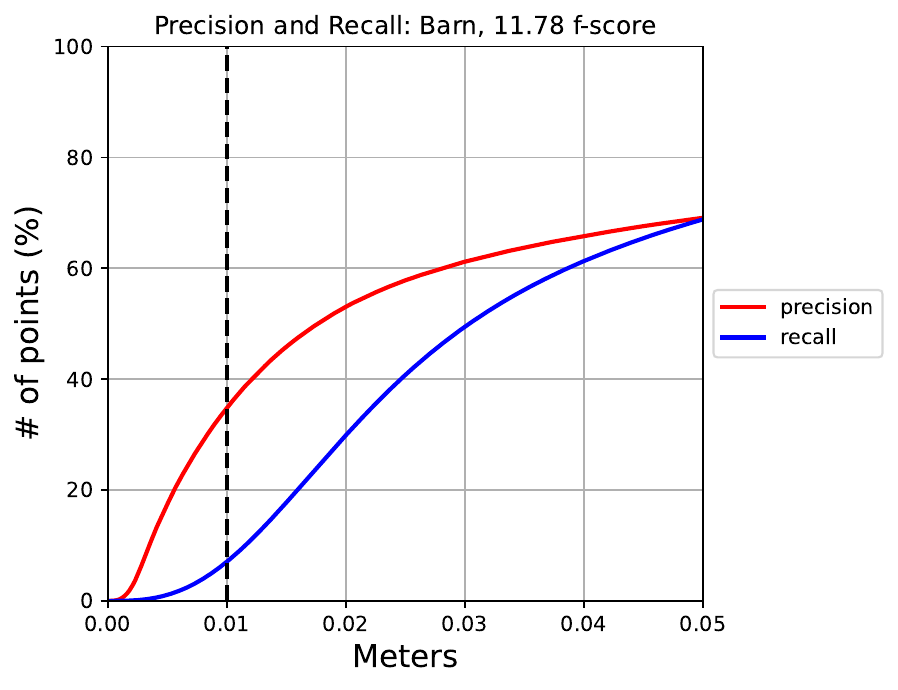} \\
        
    \end{tabular}
    \caption{Qualitative evaluation of splat center precision using FatesGS, TV-SGS\textsubscript{F} and TV-SGS\textsubscript{F} \textit{without} TV position loss on the Barn scene using \textbf{20 training views}, with \textbf{Mast3r}~\cite{duisterhof2025mast3r} initialization. The second row is the zoomed in view of the area highlighted by the red squares in the images in the top row. Brighter colors means higher precision. Notice the amount of floaters (colored black due to low precision), especially when viewed from inside out. The bottom row shows the precision-recall curves generated by the official TnT evaluation code.
    TV-SGS\textsubscript{F} has higher precision both inside and beyond the threshold. The official threshold for this scene is 0.01 meters. The precision metric for FatesGS is 0.317, for TV-SGS\textsubscript{F} without TV position loss it is 0.359, and for TV-SGS\textsubscript{F} full model precision is 0.386.}
    \label{fig_supp:TnT_ptcloud_qualitative}
\end{figure*}
\clearpage

%% file: figs_supp/tnt_compare_supp.tex
\renewcommand\quartlwrend{0.20\linewidth}
\renewcommand\rotvert{0.45cm}
\clearpage
\begin{figure*}[tb]
    \centering
    \textbf{Novel View Synthesis}\\[3pt]
    \setlength{\tabcolsep}{0.4pt}
    \renewcommand{\arraystretch}{0}
    \begin{tabular}{@{}lcccc@{}}
        
        \rotatebox{90}{\hspace{\rotvert} GT} &
        \includegraphics[width=\quartlwrend]{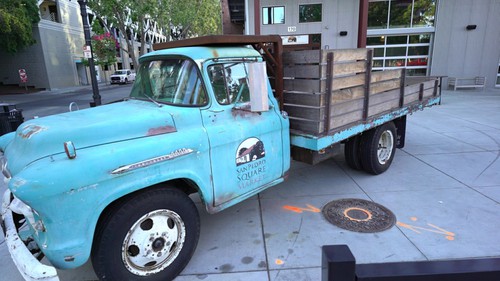} &
        \includegraphics[width=\quartlwrend]{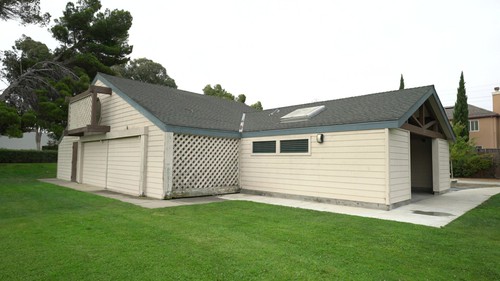} &
        \includegraphics[width=\quartlwrend]{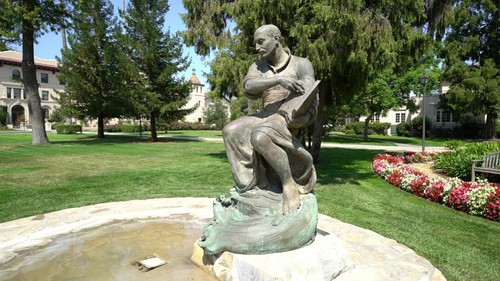} &
        \includegraphics[width=\quartlwrend]{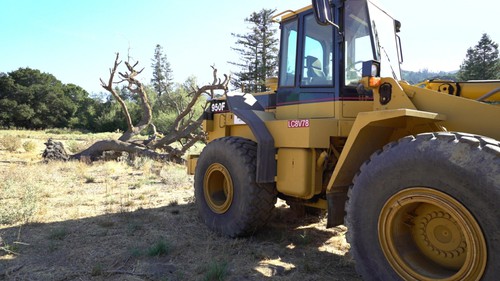} \\

        \rotatebox{90}{\hspace{\rotvert} PGSR} &
        \includegraphics[width=\quartlwrend]{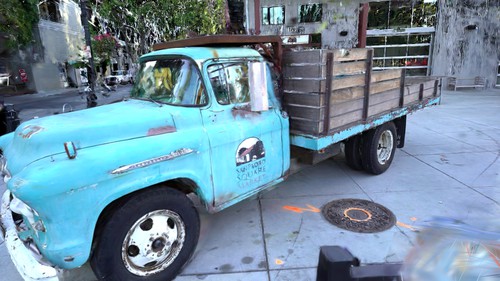} &
        \includegraphics[width=\quartlwrend]{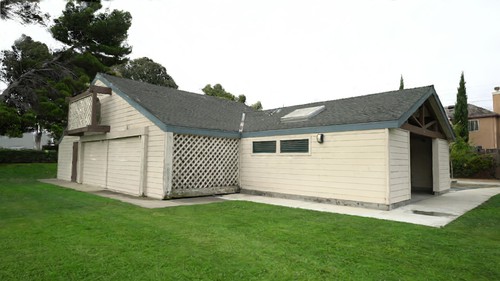} &
        \includegraphics[width=\quartlwrend]{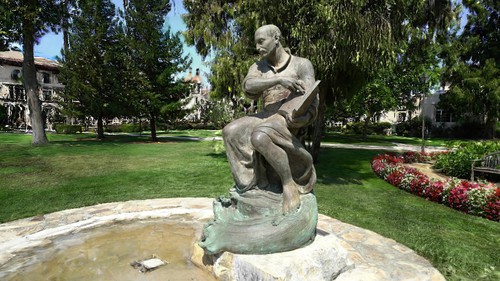} &
        \includegraphics[width=\quartlwrend]{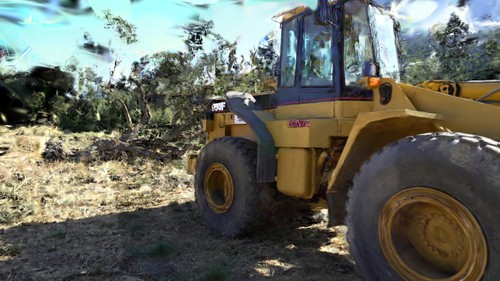} \\

        \rotatebox{90}{\hspace{\rotvert} TV-SGS\textsubscript{P}} &
        \includegraphics[width=\quartlwrend]{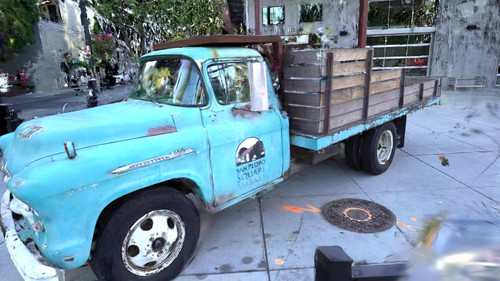} &
        \includegraphics[width=\quartlwrend]{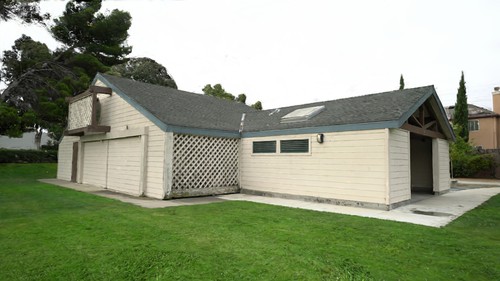} &
        \includegraphics[width=\quartlwrend]{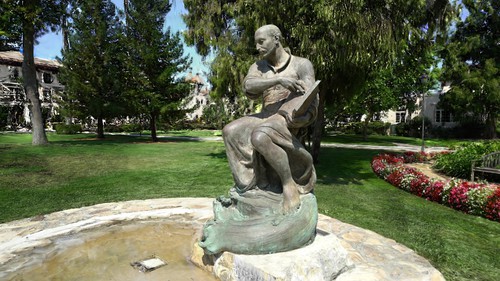} &
        \includegraphics[width=\quartlwrend]{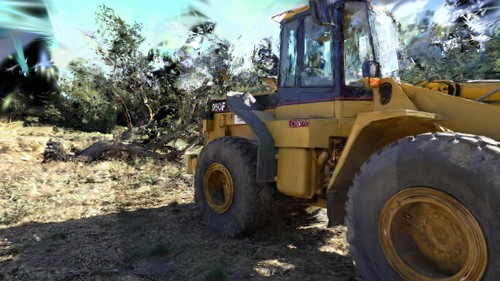} \\

        \rotatebox{90}{\hspace{\rotvert} FatesGS} &
        \includegraphics[width=\quartlwrend]{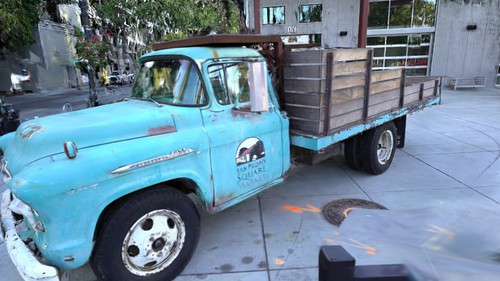} &
        \includegraphics[width=\quartlwrend]{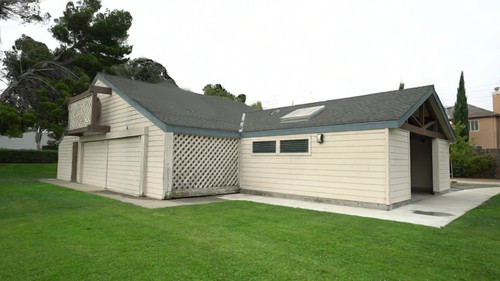} &
        \includegraphics[width=\quartlwrend]{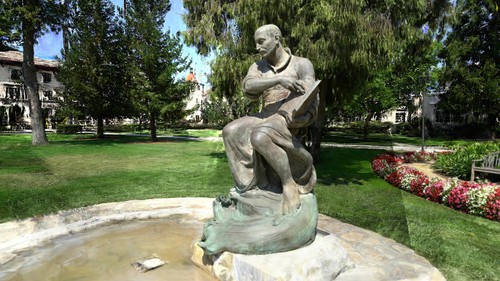} &
        \includegraphics[width=\quartlwrend]{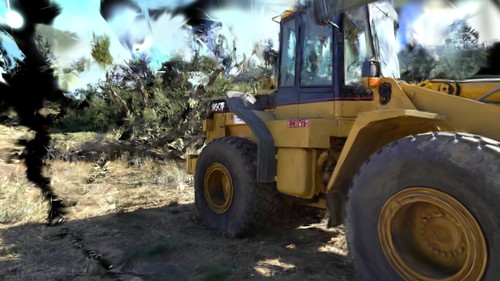} \\

        \rotatebox{90}{\hspace{\rotvert} TV-SGS\textsubscript{F}} &
        \includegraphics[width=\quartlwrend]{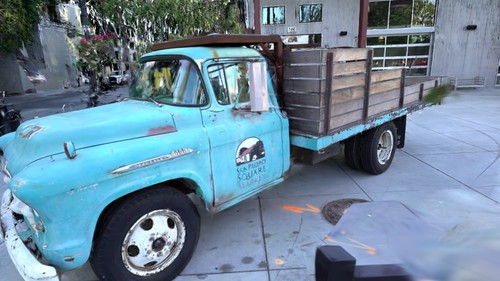} &
        \includegraphics[width=\quartlwrend]{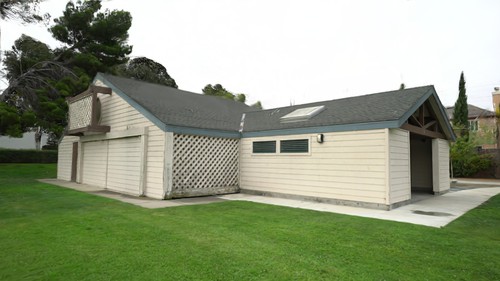} &
        \includegraphics[width=\quartlwrend]{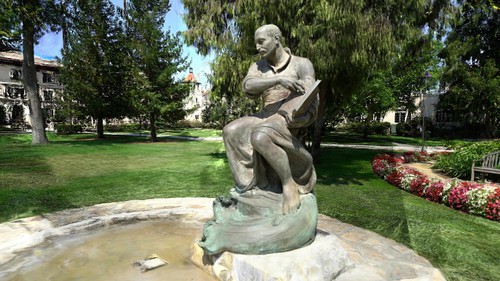} &
        \includegraphics[width=\quartlwrend]{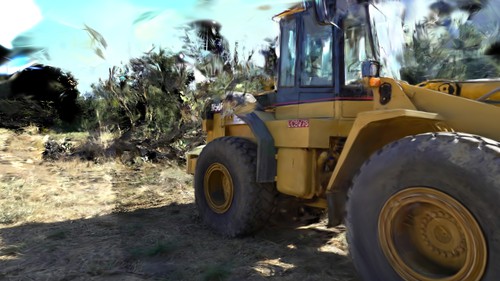} \\

        \rotatebox{90}{\hspace{\rotvert} VGGS} &
        \includegraphics[width=\quartlwrend]{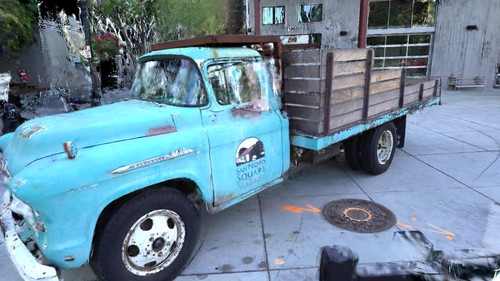} &
        \includegraphics[width=\quartlwrend]{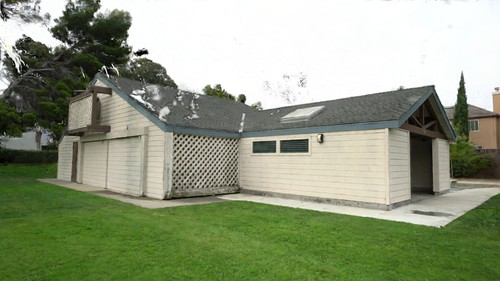} &
        \includegraphics[width=\quartlwrend]{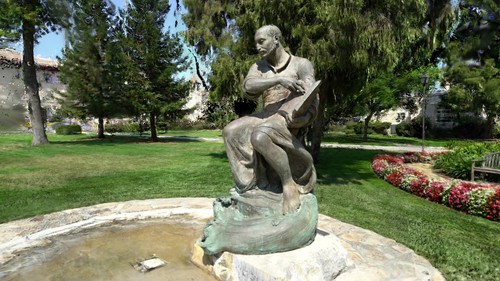} &
        \includegraphics[width=\quartlwrend]{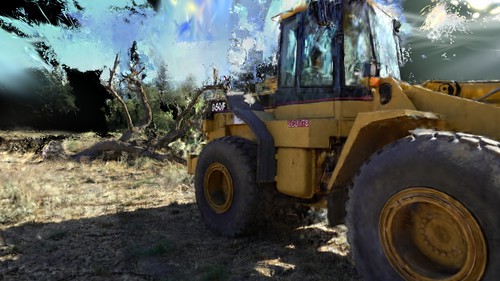} \\

        \rotatebox{90}{\hspace{\rotvert} TV-SGS\textsubscript{V}} &
        \includegraphics[width=\quartlwrend]{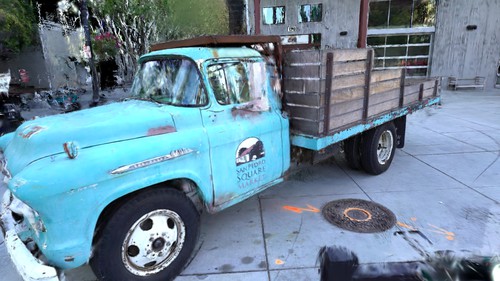} &
        \includegraphics[width=\quartlwrend]{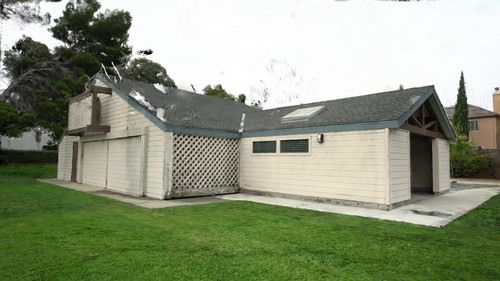} &
        \includegraphics[width=\quartlwrend]{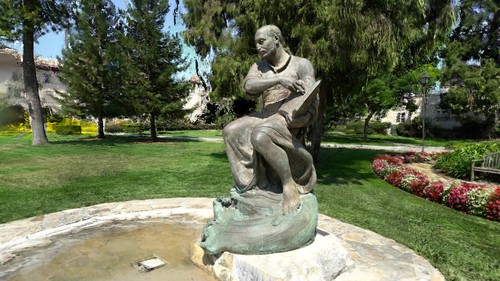} &
        \includegraphics[width=\quartlwrend]{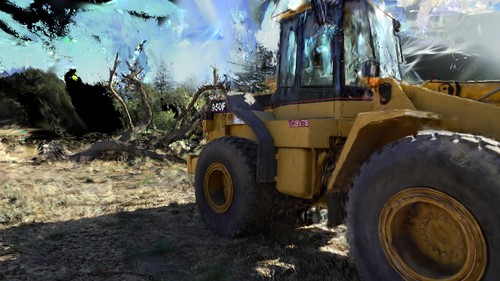} \\

        \end{tabular}
    \caption{Qualitative comparison of RGB renderings between TV-SGS and the different backbones on the TnT dataset using \textbf{only  20 training views}.}
    \label{fig_supp:TnT_compare_qualitative}
\end{figure*}
\clearpage

%% file: figs_supp/tnt_compare_mesh.tex
\renewcommand\quartlwrend{0.20\linewidth}
\renewcommand\rotvert{0.45cm}
\clearpage
\begin{figure*}[tb]
    \centering
    \textbf{Surface Reconstruction}\\[3pt]
    \setlength{\tabcolsep}{0.4pt}
    \renewcommand{\arraystretch}{0}
    \begin{tabular}{@{}lcccc@{}}
        
        \rotatebox{90}{\hspace{\rotvert} GT} &
        \includegraphics[width=\quartlwrend]{imgs_supp/tnt/Truck/view_000053/jpegs/gt.jpeg} &
        \includegraphics[width=\quartlwrend]{imgs_supp/tnt/Barn/view_000409/jpegs/gt.jpeg} &
        \includegraphics[width=\quartlwrend]{imgs_supp/tnt/Ignatius/view_000055/jpegs/gt.jpeg} &
        \includegraphics[width=\quartlwrend]{imgs_supp/tnt/Caterpillar/view_000255/jpegs/gt.jpeg} \\

        \rotatebox{90}{\hspace{\rotvert} PGSR} &
        \includegraphics[width=\quartlwrend]{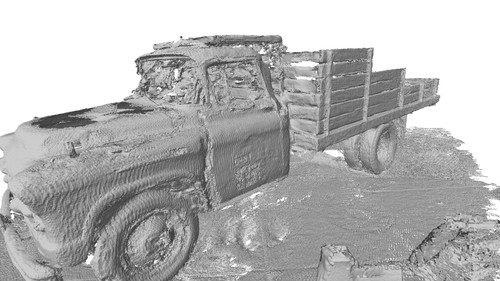} &
        \includegraphics[width=\quartlwrend]{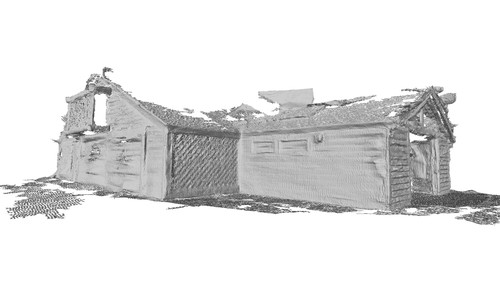} &
        \includegraphics[width=\quartlwrend]{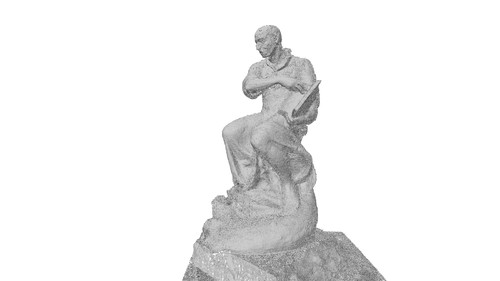} &
        \includegraphics[width=\quartlwrend]{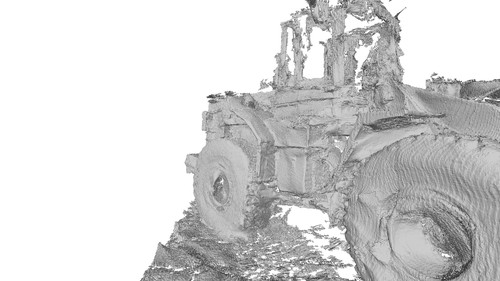} \\

        \rotatebox{90}{\hspace{\rotvert} TV-SGS\textsubscript{P}} &
        \includegraphics[width=\quartlwrend]{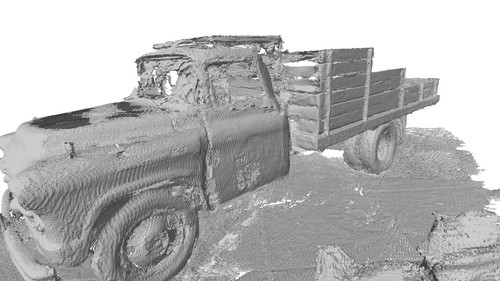} &
        \includegraphics[width=\quartlwrend]{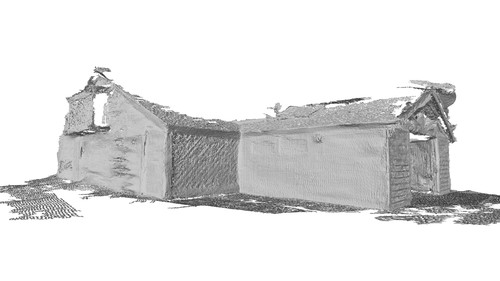} &
        \includegraphics[width=\quartlwrend]{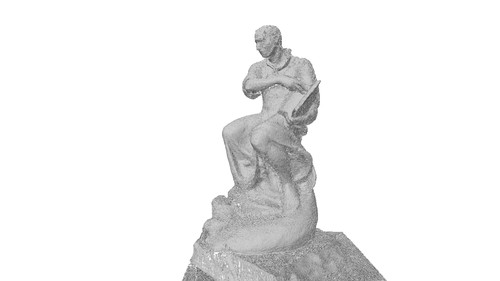} &
        \includegraphics[width=\quartlwrend]{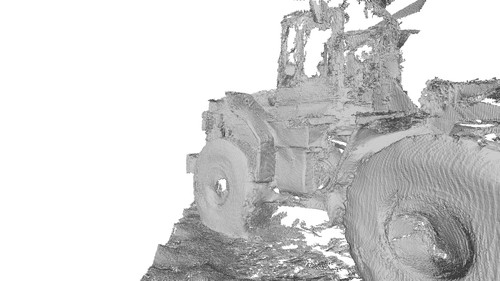} \\

        \rotatebox{90}{\hspace{\rotvert} FatesGS} &
        \includegraphics[width=\quartlwrend]{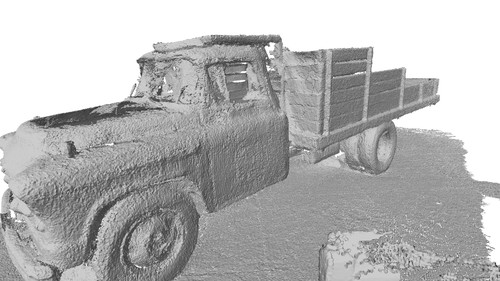} &
        \includegraphics[width=\quartlwrend]{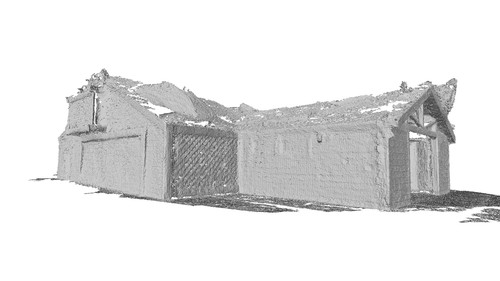} &
        \includegraphics[width=\quartlwrend]{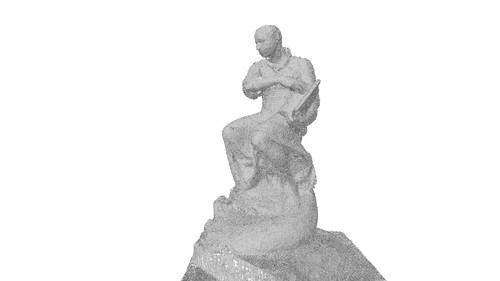} &
        \includegraphics[width=\quartlwrend]{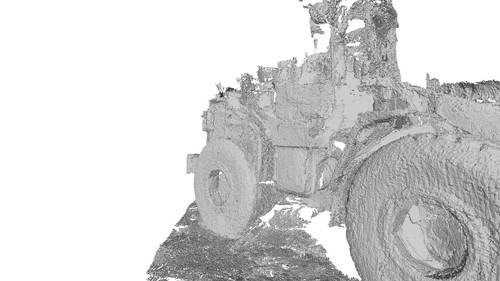} \\

        \rotatebox{90}{\hspace{\rotvert} TV-SGS\textsubscript{F}} &
        \includegraphics[width=\quartlwrend]{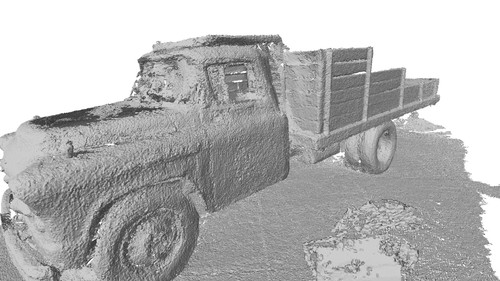} &
        \includegraphics[width=\quartlwrend]{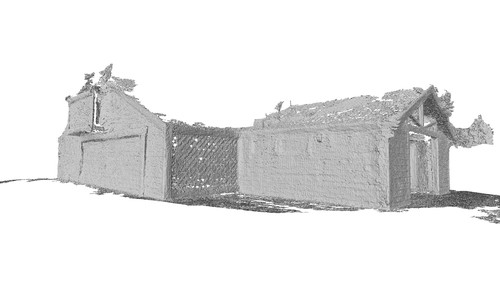} &
        \includegraphics[width=\quartlwrend]{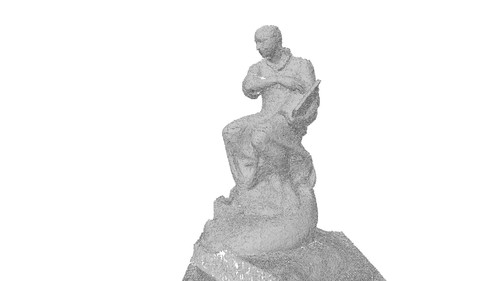} &
        \includegraphics[width=\quartlwrend]{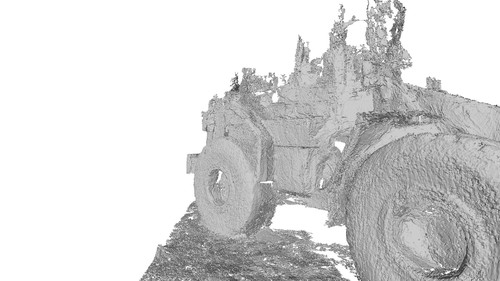} \\

        \rotatebox{90}{\hspace{\rotvert} VGGS} &
        \includegraphics[width=\quartlwrend]{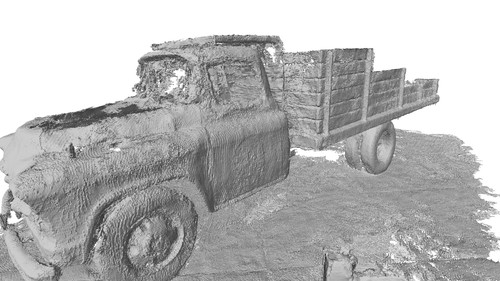} &
        \includegraphics[width=\quartlwrend]{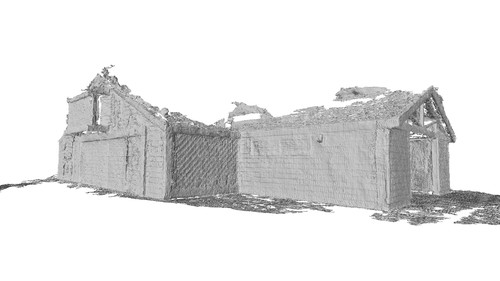} &
        \includegraphics[width=\quartlwrend]{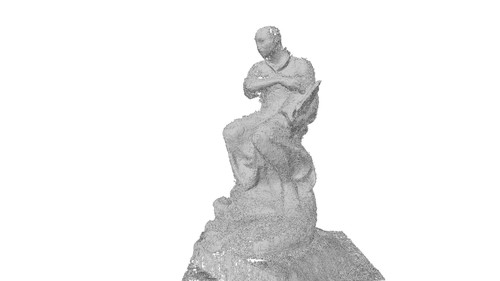} &
        \includegraphics[width=\quartlwrend]{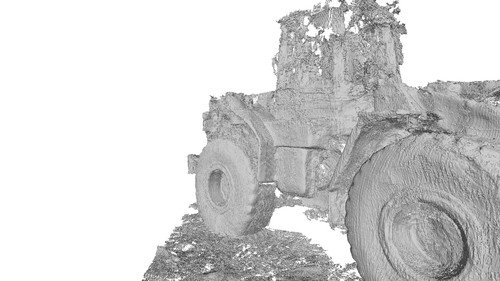} \\

        \rotatebox{90}{\hspace{\rotvert} TV-SGS\textsubscript{V}} &
        \includegraphics[width=\quartlwrend]{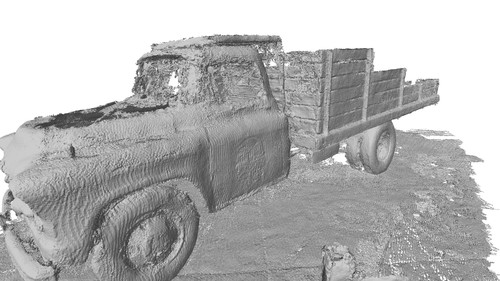} &
        \includegraphics[width=\quartlwrend]{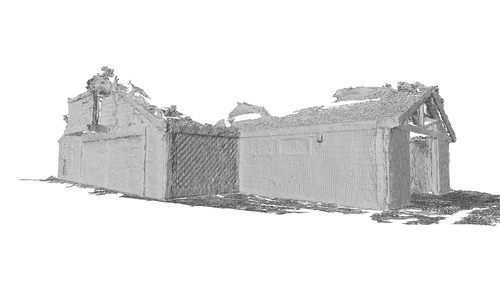} &
        \includegraphics[width=\quartlwrend]{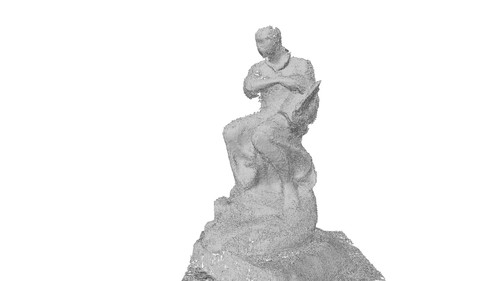} &
        \includegraphics[width=\quartlwrend]{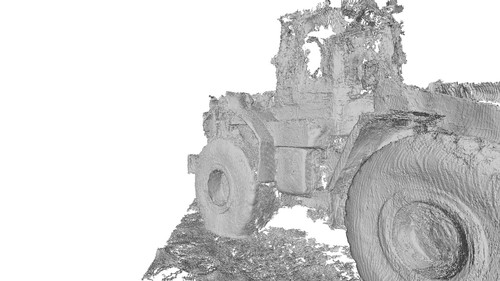} \\

        \end{tabular}
    \caption{Qualitative comparison of extracted meshes between TV-SGS and the different backbones on the TnT dataset using \textbf{only 20 training views}.}
    \label{fig_supp:TnT_compare_qualitative_mesh}
\end{figure*}
\clearpage

%% file: figs_supp/dtu_compare_supp.tex
\renewcommand\quartlwrend{0.20\linewidth}
\renewcommand\rotvert{0.45cm}
\clearpage
\begin{figure*}[tb]
    \centering
    \textbf{Novel View Synthesis}\\[3pt]
    \setlength{\tabcolsep}{0.4pt}
    \renewcommand{\arraystretch}{0}
    \begin{tabular}{@{}lccc@{}}
        \rotatebox{90}{\hspace{\rotvert} GT} &
        \includegraphics[width=\quartlwrend]{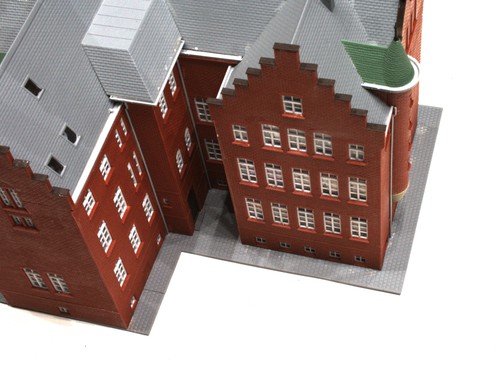} &
        \includegraphics[width=\quartlwrend]{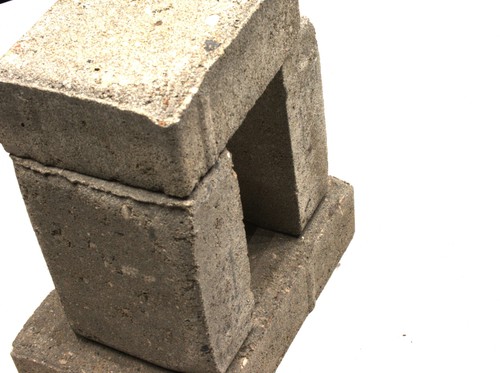} &
        \includegraphics[width=\quartlwrend]{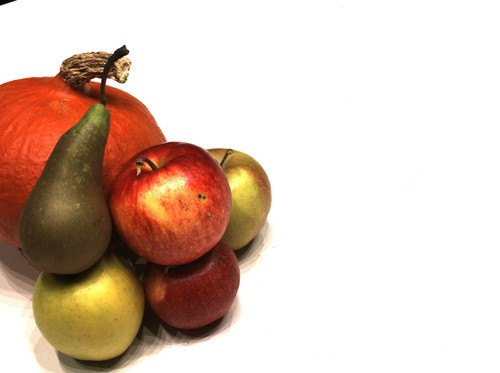} \\

        \rotatebox{90}{\hspace{\rotvert} PGSR} &
        \includegraphics[width=\quartlwrend]{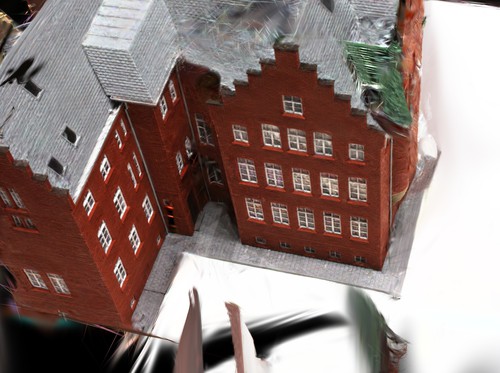} &
        \includegraphics[width=\quartlwrend]{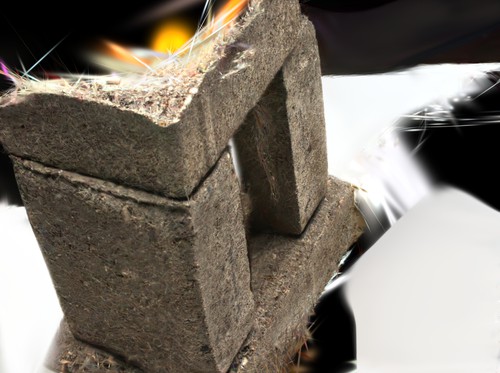} &
        \includegraphics[width=\quartlwrend]{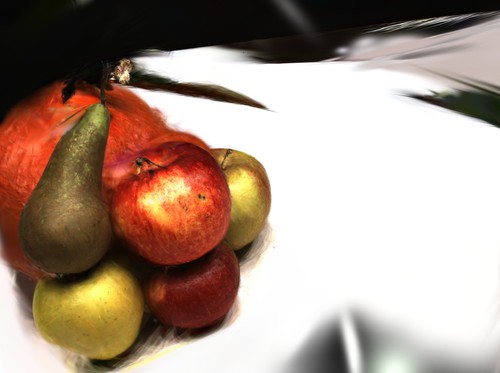} \\

        \rotatebox{90}{\hspace{\rotvert} TV-SGS\textsubscript{P}} &
        \includegraphics[width=\quartlwrend]{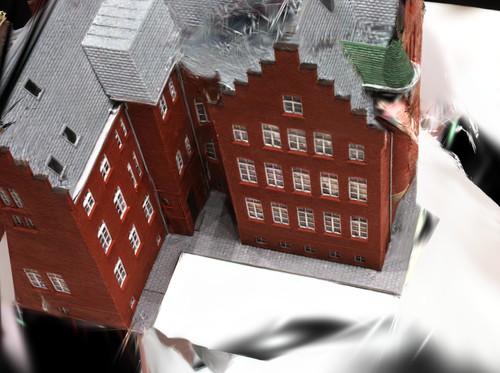} &
        \includegraphics[width=\quartlwrend]{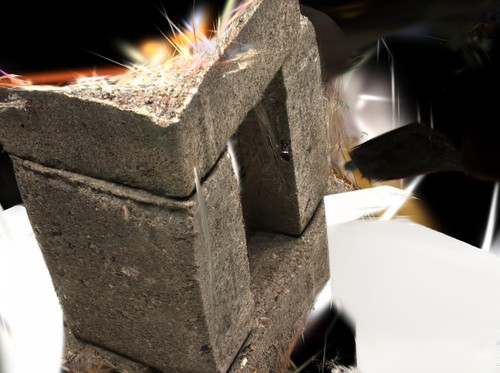} &
        \includegraphics[width=\quartlwrend]{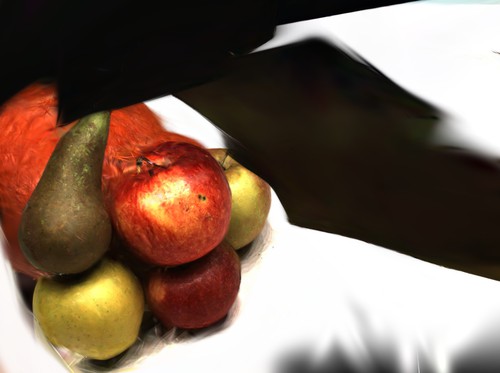} \\
        
        \rotatebox{90}{\hspace{\rotvert} FatesGS} &
        \includegraphics[width=\quartlwrend]{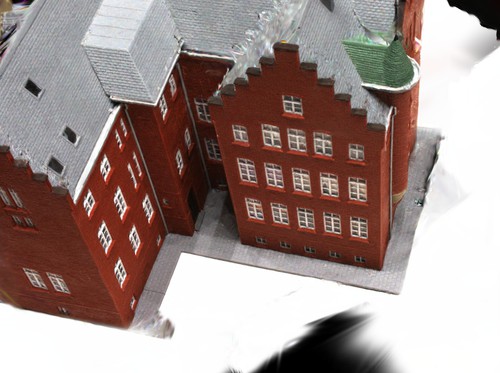} &
        \includegraphics[width=\quartlwrend]{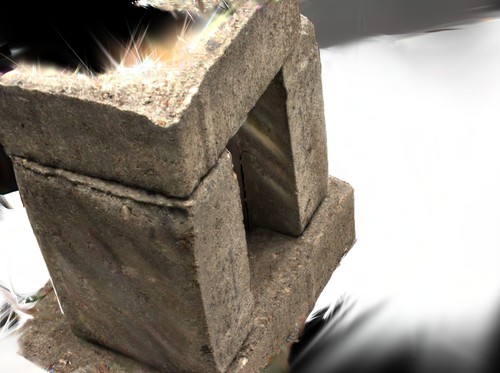} &
        \includegraphics[width=\quartlwrend]{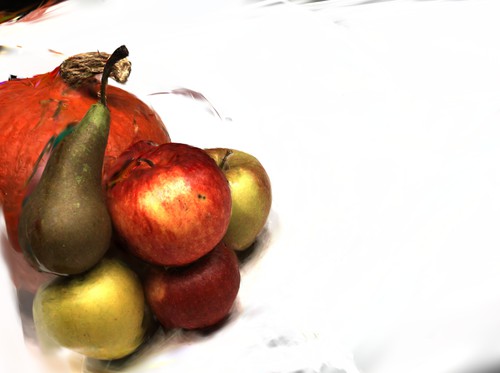} \\

        \rotatebox{90}{\hspace{\rotvert} TV-SGS\textsubscript{F}} &
        \includegraphics[width=\quartlwrend]{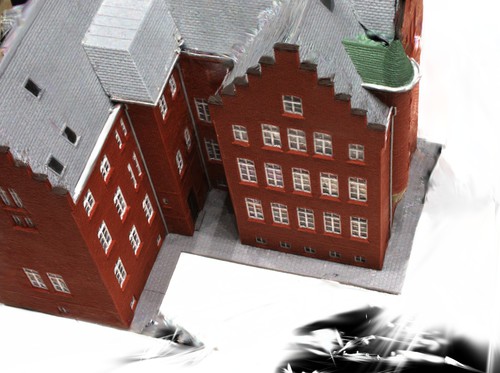} &
        \includegraphics[width=\quartlwrend]{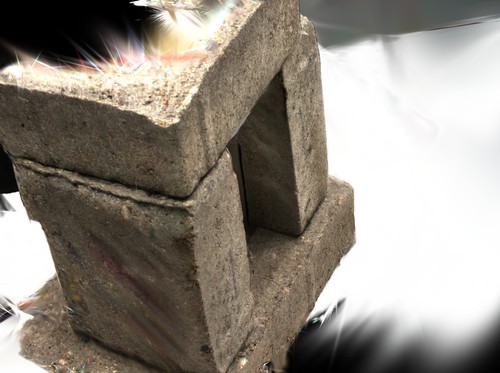} &
        \includegraphics[width=\quartlwrend]{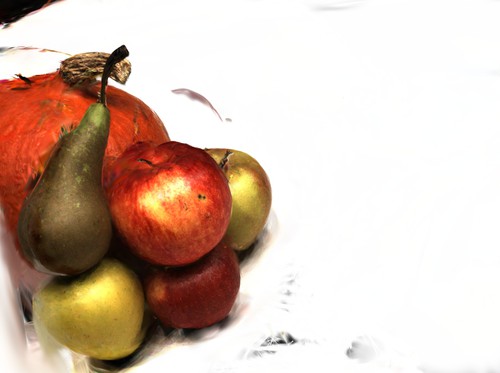} \\

        \rotatebox{90}{\hspace{\rotvert} VGGS} &
        \includegraphics[width=\quartlwrend]{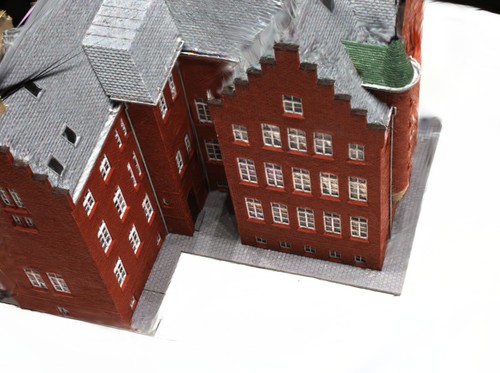} &
        \includegraphics[width=\quartlwrend]{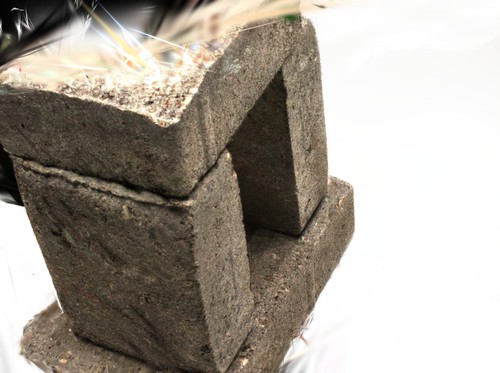} &
        \includegraphics[width=\quartlwrend]{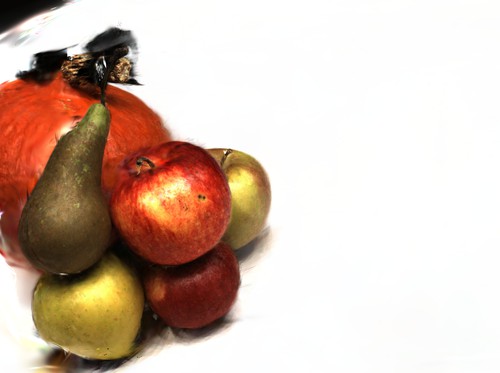} \\

        \rotatebox{90}{\hspace{\rotvert} TV-SGS\textsubscript{V}} &
        \includegraphics[width=\quartlwrend]{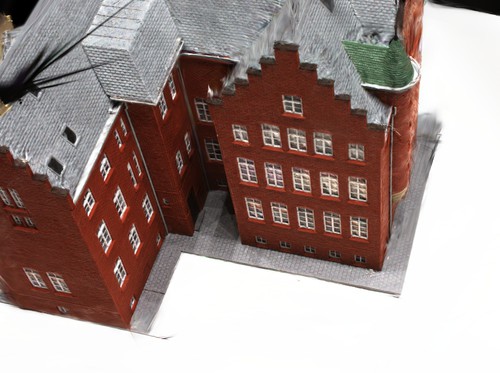} &
        \includegraphics[width=\quartlwrend]{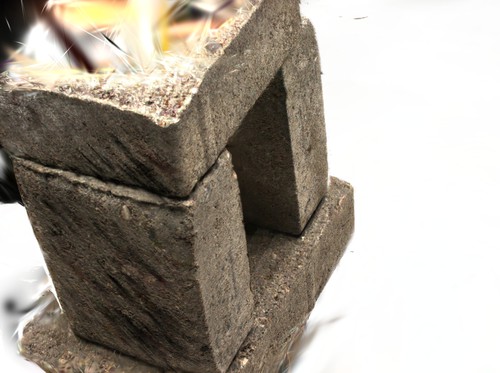} &
        \includegraphics[width=\quartlwrend]{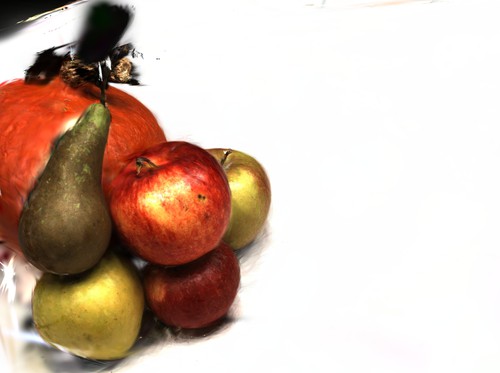} \\

        \end{tabular}
    \caption{Qualitative comparison of RGB renderings between TV-SGS and the different backbones on the DTU dataset, small-overlap split.}
    \label{fig_supp:dtu_compare_qualitative}
\end{figure*}
\clearpage

%% file: figs_supp/dtu_compare_mesh_supp.tex
\renewcommand\quartlwrend{0.20\linewidth}
\renewcommand\rotvert{0.45cm}
\clearpage
\begin{figure*}[tb]
    \centering
    \textbf{Surface Reconstruction}\\[3pt]
    \setlength{\tabcolsep}{0.4pt}
    \renewcommand{\arraystretch}{0}
    \begin{tabular}{@{}lccc@{}}
        \rotatebox{90}{\hspace{\rotvert} GT} &
        \includegraphics[width=\quartlwrend]{imgs_supp/dtu/small_overlap/scan24/jpegs/gt.jpeg} &
        \includegraphics[width=\quartlwrend]{imgs_supp/dtu/small_overlap/scan40/jpegs/gt.jpeg} &
        \includegraphics[width=\quartlwrend]{imgs_supp/dtu/small_overlap/scan63/jpegs/gt.jpeg} \\

        \rotatebox{90}{\hspace{\rotvert} PGSR} &
        \includegraphics[width=\quartlwrend]{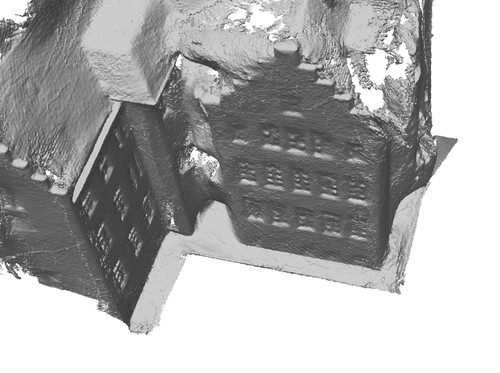} &
        \includegraphics[width=\quartlwrend]{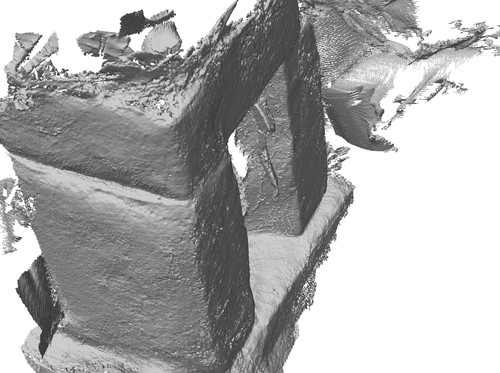} &
        \includegraphics[width=\quartlwrend]{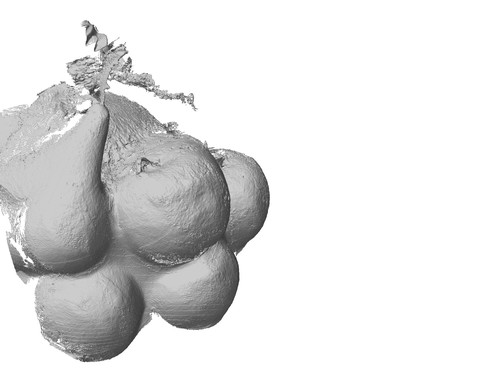} \\

        \rotatebox{90}{\hspace{\rotvert} TV-SGS\textsubscript{P}} &
        \includegraphics[width=\quartlwrend]{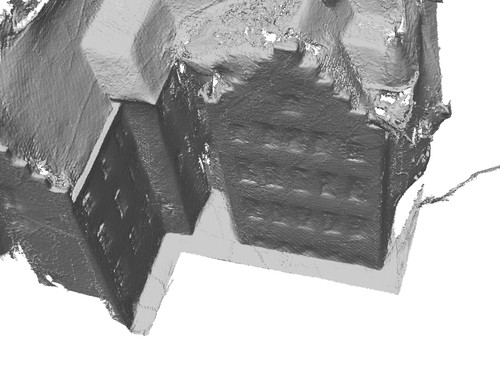} &
        \includegraphics[width=\quartlwrend]{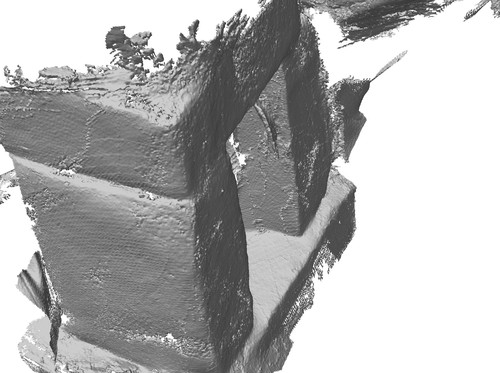} &
        \includegraphics[width=\quartlwrend]{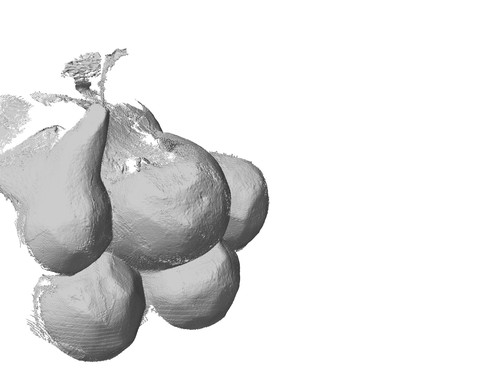} \\
        
        \rotatebox{90}{\hspace{\rotvert} FatesGS} &
        \includegraphics[width=\quartlwrend]{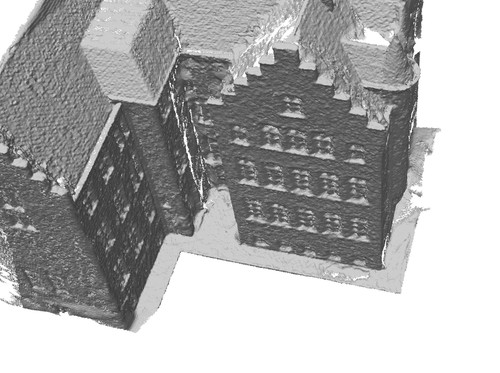} &
        \includegraphics[width=\quartlwrend]{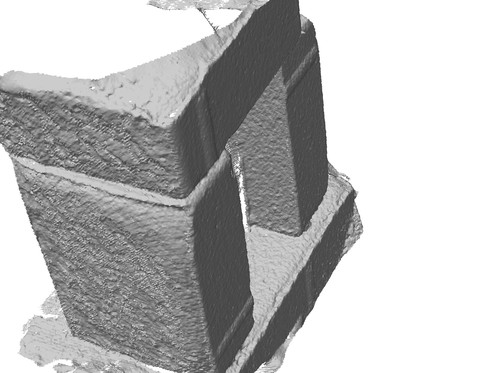} &
        \includegraphics[width=\quartlwrend]{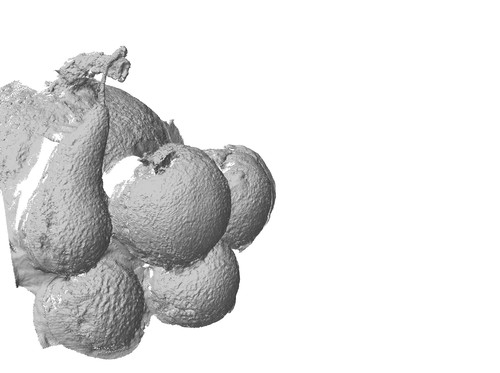} \\

        \rotatebox{90}{\hspace{\rotvert} TV-SGS\textsubscript{F}} &
        \includegraphics[width=\quartlwrend]{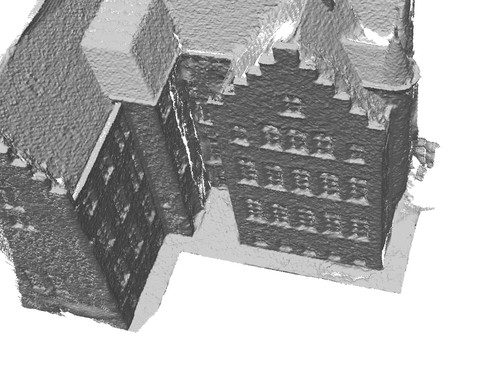} &
        \includegraphics[width=\quartlwrend]{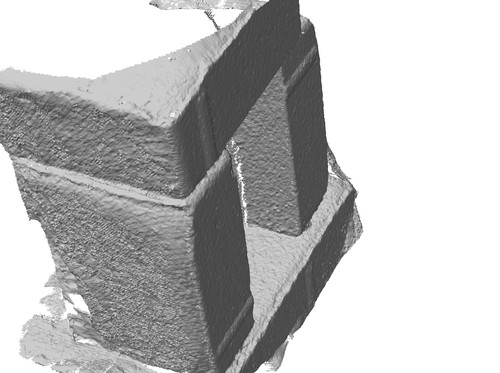} &
        \includegraphics[width=\quartlwrend]{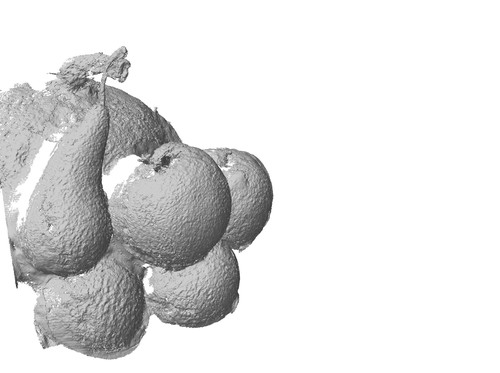} \\

        \rotatebox{90}{\hspace{\rotvert} VGGS} &
        \includegraphics[width=\quartlwrend]{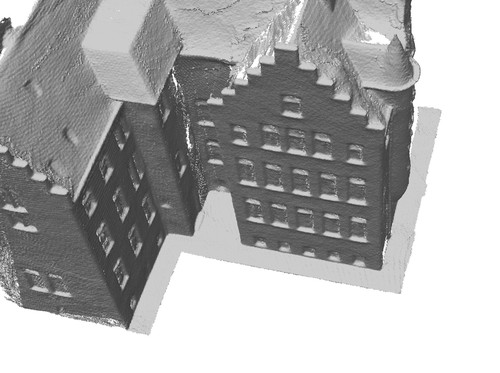} &
        \includegraphics[width=\quartlwrend]{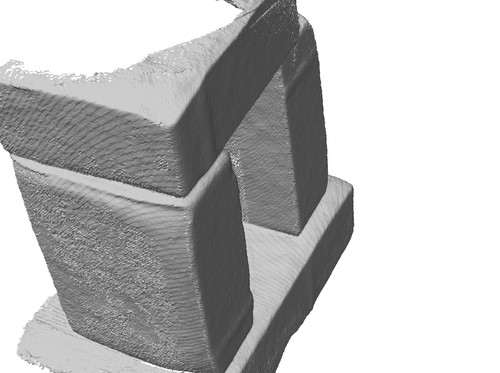} &
        \includegraphics[width=\quartlwrend]{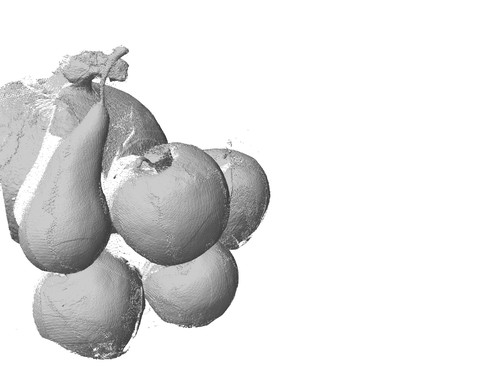} \\

        \rotatebox{90}{\hspace{\rotvert} TV-SGS\textsubscript{V}} &
        \includegraphics[width=\quartlwrend]{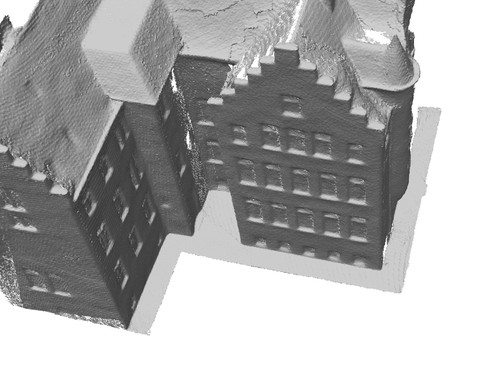} &
        \includegraphics[width=\quartlwrend]{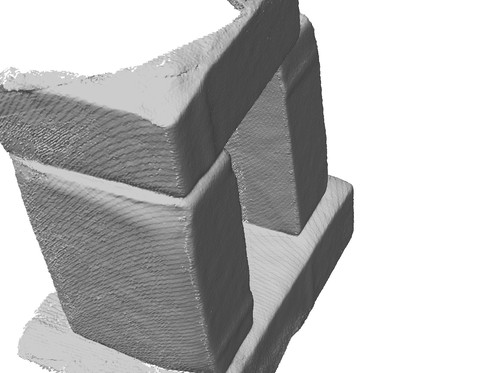} &
        \includegraphics[width=\quartlwrend]{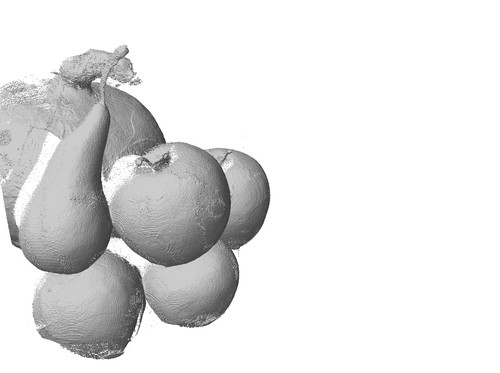} \\

        \end{tabular}
    \caption{Qualitative comparison of meshes extracted between TV-SGS and the different backbones on the DTU dataset small-overlap split.}
    \label{fig_supp:dtu_compare_qualitative_mesh}
\end{figure*}
\clearpage

%% file: sec_supp/dense_views.tex
\section{Dense Views}
\label{sec_supp:dense_views}

Here, we present results on the TnT dataset using \textit{all training views}, that is seven out of every eight views, leaving the eighth view for testing, which is the standard train-test split in the literature using TnT.
We integrate TV-SGS with three popular state of the art dense-view backbones with different types of GS optimizations: PGSR~\cite{Chen_2024_PGSR} which is the state of the art backbone in dense views and uses MVS based losses, RaDe-GS~\cite{Zhang_2026_RaDe} which uses an efficient closed form planar depth estimation, and 2DGS~\cite{Huang_2024_2DGS}, which uses 2D ellipses instead of the standard 3D Gaussians to represent the scene. The respective TV-SGS integrations are called TV-SGS\textsubscript{P}, TV-SGS\textsubscript{R} and TV-SGS\textsubscript{2D}. 

We use sparse SfM points obtained from COLMAP~\cite{schonberger2016structure} which is the standard initialization procedure for dense views. The initial SfM points are too sparse for Tensor Voting, so we start Tensor Voting at iteration 20,000 with the TV normal loss, while we activate the TV position loss at iteration 25,000. We perform voting every 10 iterations, and update each splat's neighbors every 500 iterations with the number of neighbors $k$ set to 100.
Meshes are extracted following their respective backbones, i.e. using Marching Tetrahedra for TV-SGS\textsubscript{R} and TSDF depth fusion for TV-SGS\textsubscript{2D} and TV-SGS\textsubscript{P}.

The F1 scores are evaluated both on the mesh and splat centers. We also show some qualitative results on TnT and DTU scenes using dense training views in Figures~\ref{fig:tnt_mesh_all_dense} and~\ref{fig_supp:DTU_all_mesh_dense}.

\input{tabs_supp/tnt/tnt_mesh_dense_views}
\input{tabs_supp/tnt/tnt_pcd_dense_views}

\subsection{Implementation Details for Dense-View Backbones}
The TV-SGS\textsubscript{R} implementation, which uses RaDe-GS as backbone does not include the depth distortion loss, even though it is part of the original RaDe-GS algorithm. This is because the authors removed the loss in their official implementation citing better performance without it. Benchmarking against baseline RaDe-GS is also done without this loss. We start the rendered normal loss of RaDe-GS at iteration 15,000, and set this loss weight to 0.05, same as in the original paper. We also make use of the 3D filter first proposed in Mip-Splatting~\cite{yu_2024_mip}, same as in the original RaDe-GS implementation. Densification and pruning strategies are also carried over from RaDe-GS.

The TV-SGS\textsubscript{2D} implementation follows the same training, densification and pruning schedule as the original 2DGS algorithm. Depth distortion loss starts at iteration 3000, and the rendered normal loss starts at iteration 7000. The weight for the rendered normal loss is set to 0.05, and the weight for the depth distortion loss is scene specific, and is set according to 2DGS' official implementation.

The TV-SGS\textsubscript{P} implementation follows the same training schedule as the official implementation of PGSR, where the min-scale loss is applied from the beginning of the optimization process, and the single view and multi-view losses are started after 7000 iterations. The densification and pruning schemes are also kept the same, and the total iterations for dense views is 30,000.

\input{tabs_supp/tnt/time_and_memory}

Note that we evaluate the dense view and sparse view backbones ourselves for a fair comparison with the respective TV-SGS implementations, using the same initialization for all.

\subsection{Quantitative and Qualitative Results}
\label{sec_supp:dense_views_qual_quant}
As can be seen from Table \ref{tab:tnt_mesh_dense_views}, TV-SGS\textsubscript{2D} and TV-SGS\textsubscript{R} improve upon the backbones by about 8.5\% and 3\% respectively, while TV-SGS\textsubscript{P} maintains the performance of PGSR backbone. This is because PGSR's multi-view losses provide adequate geometry supervision under dense views, making TV-SGS\textsubscript{P} redundant for rendered geometry and also for mesh extraction. Table \ref{tab:tnt_pcd_dense_views} gives the F1 scores evaluated on the splat centers, while Figures \ref{fig:tnt_mesh_all_dense} and \ref{fig_supp:DTU_all_mesh_dense} show some qualitative results of the extracted surfaces.

We also show the optimization runtime (wall time) and peak GPU memory usage on an Nvidia A6000 GPU for sparse and dense TnT runs in Table \ref{tab:runtime_and_memory}. Please note the code is not optimized for speed or memory footprint. Also note that we start TV-SGS module right from iteration 1 for sparse views and from iteration 20,000 for dense views, and that for sparse views we initialize the TnT scene with about 1.5 million points from Mast3r/MVSAnywhere, while the COLMAP SfM initialization starts off from about 150,000 points for dense views, which affects the GPU memory and runtimes. PGSR and TV-SGS\textsubscript{P} for sparse views were run for 15,000 iterations and 30,000 iterations for dense views, as mentioned previously in implementation details.

\input{figs_supp/tnt_mesh_dense}
\input{figs_supp/dtu_dense_mesh}

%% file: tabs_supp/tnt/tnt_mesh_dense_views.tex
\begin{table}[h!]
  \centering
  \setlength{\tabcolsep}{3pt}
  \resizebox{\columnwidth}{!}{%
  \begin{tabular}{c|c||c||c}
  \hline
     & \textbf{P} / \textbf{R} / \textbf{F1} & \textbf{P} / \textbf{R} / \textbf{F1} & \textbf{P} / \textbf{R} / \textbf{F1} \\
     \hline
    \textbf{Scenes} & \textbf{PGSR} & \textbf{RaDe-GS} & \textbf{2DGS} \\
    \hline
    \textbf{Barn} & 0.646 / 0.653 / 0.649 & 0.590 / 0.446 / 0.508 & 0.376 / 0.486 / 0.424 \\
    \textbf{Caterpillar} & 0.323 / 0.649 / 0.432 & 0.344 / 0.298 / 0.319 & 0.146 / 0.400 / 0.214 \\
    \textbf{Courthouse} & 0.170 / 0.304 / 0.218 & 0.389 / 0.178 / 0.244 & 0.074 / 0.114 / 0.090 \\
    \textbf{Ignatius} & 0.789 / 0.759 / 0.774 & 0.689 / 0.539 / 0.605 & 0.366 / 0.501 / 0.423 \\
    \textbf{Meetingroom} & 0.332 / 0.319 / 0.325 & 0.353 / 0.185 / 0.243 & 0.183 / 0.168 / 0.175 \\
    \textbf{Truck} & 0.568 / 0.682 / 0.620 & 0.565 / 0.548 / 0.556 & 0.374 / 0.539 / 0.441 \\
    \hline
    \textbf{Average} & 0.471 / 0.561 / 0.503 & 0.488 / 0.366 / 0.413 & 0.253 / 0.368 / 0.294 \\
    \hline
     & \textbf{TV-SGS\textsubscript{P}} & \textbf{TV-SGS\textsubscript{R}} & \textbf{TV-SGS\textsubscript{2D}} \\
     \hline
    \textbf{Barn} & 0.641 / 0.641 / 0.641 & 0.616 / 0.471 / 0.534 & 0.421 / 0.502 / 0.458 \\
    \textbf{Caterpillar} & 0.315 / 0.630 / 0.420 & 0.355 / 0.316 / 0.335 & 0.163 / 0.414 / 0.234 \\
    \textbf{Courthouse} & 0.142 / 0.281 / 0.189 & 0.372 / 0.181 / 0.243 & 0.120 / 0.153 / 0.134 \\
    \textbf{Ignatius} & 0.784 / 0.752 / 0.768 & 0.687 / 0.543 / 0.607 & 0.385 / 0.519 / 0.442 \\
    \textbf{Meetingroom} & 0.336 / 0.316 / 0.325 & 0.369 / 0.191 / 0.252 & 0.186 / 0.173 / 0.179 \\
    \textbf{Truck} & 0.579 / 0.692 / 0.631 & 0.588 / 0.573 / 0.581 & 0.404 / 0.561 / 0.469 \\
    \hline
    \textbf{Average} & 0.466 / 0.552 / 0.496 & 0.498 / 0.379 / 0.425 & 0.280 / 0.387 / 0.319 \\
  \end{tabular}
  }
  \caption{Quantitative results on geometry on the \textbf{TnT} dataset for \textbf{dense views} using the standard train-test split for TnT and initialized with \textbf{COLMAP}~\cite{schonberger2016structure}. The geometry is evaluated on the \textbf{mesh}. We evaluate the TnT dataset with three backbones: 2DGS, PGSR and RaDe-GS, represented as TV-SGS\textsubscript{2D}, TV-SGS\textsubscript{P} and TV-SGS\textsubscript{R} respectively. TV-SGS improves upon RaDe-GS and 2DGS, while TV-SGS\textsubscript{P} matches PGSR. P/R/F1 refers to precision, recall and F1 scores}
  \label{tab:tnt_mesh_dense_views}
\end{table}

%% file: tabs_supp/tnt/tnt_pcd_dense_views.tex
\begin{table}[hb!]
  \centering
  \setlength{\tabcolsep}{3pt}
  \resizebox{\columnwidth}{!}{%
  \begin{tabular}{c|c||c||c}
  \hline
     & \textbf{P} / \textbf{R} / \textbf{F1} & \textbf{P} / \textbf{R} / \textbf{F1} & \textbf{P} / \textbf{R} / \textbf{F1} \\
     \hline
    \textbf{Scenes} &\textbf{PGSR} & \textbf{RaDe-GS} & \textbf{2DGS} \\
    \hline
    \textbf{Barn} & 0.421 / 0.113 / 0.178 & 0.534 / 0.091 / 0.155 & 0.591 / 0.042 / 0.078 \\
    \textbf{Caterpillar} & 0.417 / 0.122 / 0.189 & 0.376 / 0.066 / 0.112 & 0.376 / 0.039 / 0.070 \\
    \textbf{Courthouse} & 0.315 / 0.056 / 0.095 & 0.354 / 0.027 / 0.050 & 0.448 / 0.021 / 0.040 \\
    \textbf{Ignatius} & 0.555 / 0.131 / 0.212 & 0.576 / 0.105 / 0.177 & 0.529 / 0.055 / 0.099 \\
    \textbf{Meetingroom} & 0.401 / 0.092 / 0.150 & 0.327 / 0.045 / 0.079 & 0.157 / 0.014 / 0.025 \\
    \textbf{Truck} & 0.524 / 0.170 / 0.256 & 0.576 / 0.139 / 0.223 & 0.600 / 0.074 / 0.131 \\
    \hline
    \textbf{Average} & 0.439 / 0.114 / 0.180 & 0.457 / 0.079 / 0.133 & 0.450 / 0.041 / 0.074 \\
    \hline
     & \textbf{TV-SGS\textsubscript{P}} & \textbf{TV-SGS\textsubscript{R}} & \textbf{TV-SGS\textsubscript{2D}} \\
     \hline
    \textbf{Barn} & 0.623 / 0.132 / 0.219 & 0.601 / 0.097 / 0.168 & 0.609 / 0.067 / 0.120 \\
    \textbf{Caterpillar} & 0.430 / 0.095 / 0.156 & 0.415 / 0.071 / 0.121 & 0.396 / 0.051 / 0.090 \\
    \textbf{Courthouse} & 0.403 / 0.060 / 0.104 & 0.393 / 0.028 / 0.053 & 0.421 / 0.019 / 0.037 \\
    \textbf{Ignatius} & 0.682 / 0.122 / 0.207 & 0.668 / 0.120 / 0.204 & 0.535 / 0.064 / 0.114 \\
    \textbf{Meetingroom} & 0.408 / 0.075 / 0.126 & 0.359 / 0.045 / 0.081 & 0.347 / 0.031 / 0.056 \\
    \textbf{Truck} & 0.625 / 0.167 / 0.264 & 0.627 / 0.147 / 0.239 & 0.611 / 0.096 / 0.165 \\
    \hline
    \textbf{Average} & 0.529 / 0.109 / 0.179 & 0.510 / 0.085 / 0.144 & 0.486 / 0.055 / 0.097 \\
  \end{tabular}
  }
  \caption{Quantitative results on the geometry of the \textbf{TnT} dataset for \textbf{dense views} evaluated on the \textbf{splat centers}, and initialized with \textbf{COLMAP}~\cite{schonberger2016structure}. These results highlight that TV-SGS is able to move the floater splat centers towards the most salient surface. TV-SGS significantly improves the position of the splats.}
  \label{tab:tnt_pcd_dense_views}
\end{table}

%% file: tabs_supp/tnt/time_and_memory.tex
\begin{table*}[h!]
  \centering
  \setlength{\tabcolsep}{5pt}
  \begin{tabular}{c|cc||cc||cc}
  \hline
    \textbf{} & \textbf{PGSR} & $\mathbf{TV-SGS_P}$ & \textbf{FatesGS} & $\mathbf{TV-SGS_F}$ & \textbf{VGGS} & $\mathbf{TV-SGS_{V}}$ \\
    \hline
    \textbf{Runtime (minutes)} & 31.40 & 55.0 & 19.80 & 35.30 & 3.10 & 12.12 \\
    \textbf{GPU Memory (GB)} & 4.07 & 13.20 & 4.60 & 7.20 & 8.05 & 14.88 \\
  \hline
    \textbf{} & \textbf{PGSR} & $\mathbf{TV-SGS_P}$ & \textbf{RaDe-GS} & $\mathbf{TV-SGS_R}$ & \textbf{2DGS} & $\mathbf{TV-SGS_{2D}}$ \\
    \hline
    \textbf{Runtime (minutes)} & 80.42 & 93.80 & 19.31 & 31.26 & 12.88 & 17.47 \\
    \textbf{GPU Memory (GB)} & 5.45 & 6.48 & 3.44 & 4.54 & 0.64 & 1.17 \\

  \end{tabular}%
  \caption{Top: Sparse views, Bottom: Dense views runtime and memory analysis on the TnT dataset showing the average runtime and peak GPU memory usage.}
  \label{tab:runtime_and_memory}
\end{table*}

%% file: figs_supp/tnt_mesh_dense.tex
\begin{figure*}[t]
  \centering
    \includegraphics[width=\linewidth]{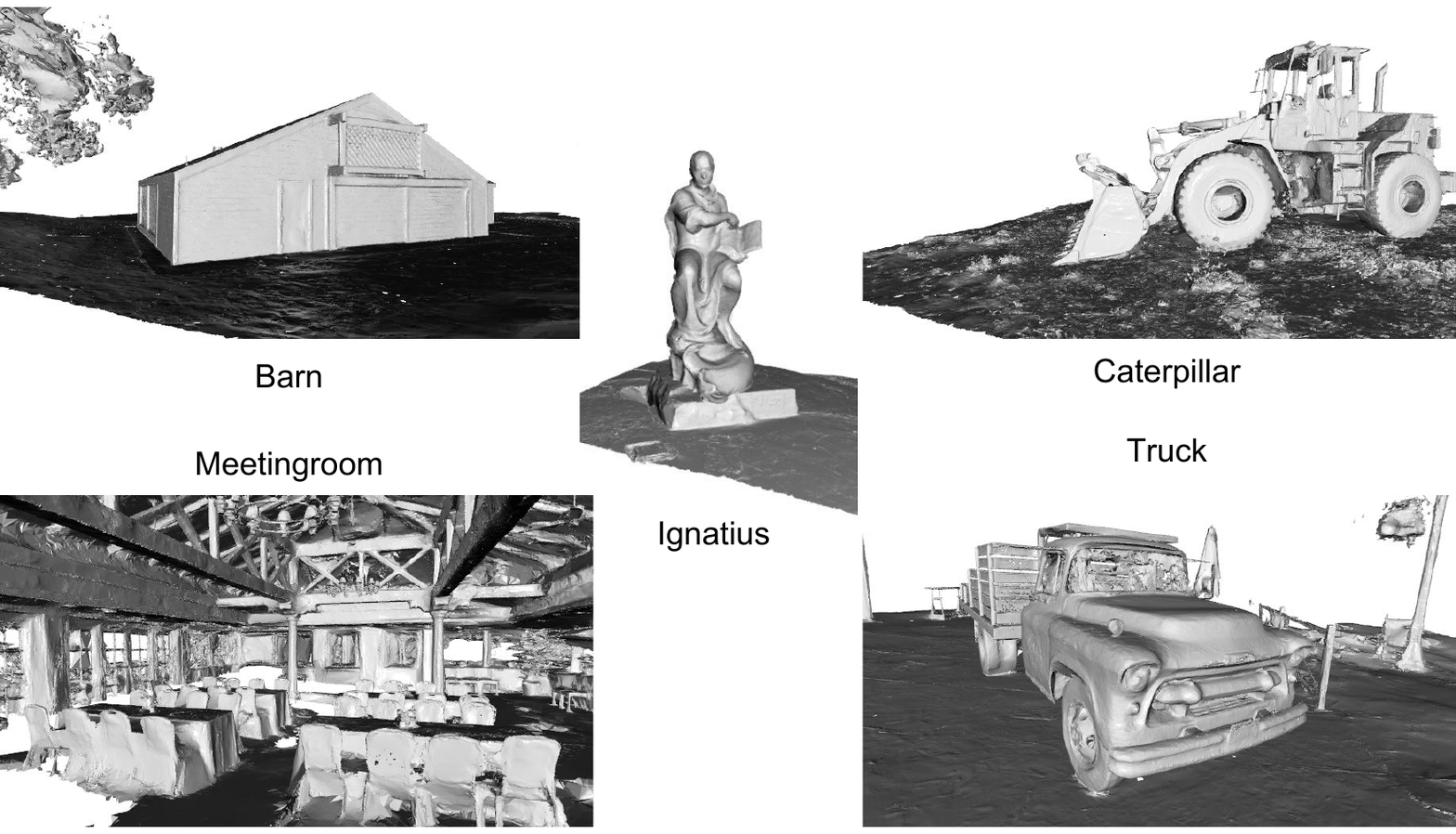}
  \caption{Qualitative results of the extracted meshes by TV-SGS\textsubscript{R} on the TnT dataset for dense views.}
  \label{fig:tnt_mesh_all_dense}
\end{figure*}

%% file: figs_supp/dtu_dense_mesh.tex
\renewcommand\quartlwrend{0.17\linewidth}
\begin{figure*}[tb]
    \centering
    \begin{tabular}{ccccc}
        \includegraphics[width=\quartlwrend]{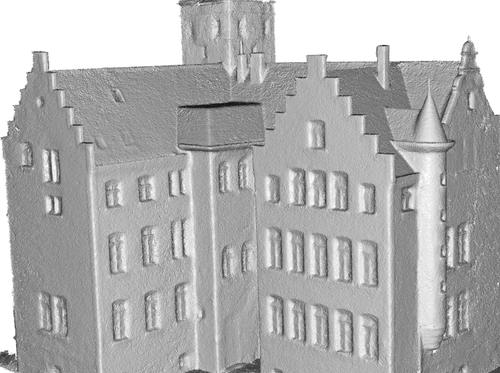} &
        \includegraphics[width=\quartlwrend]{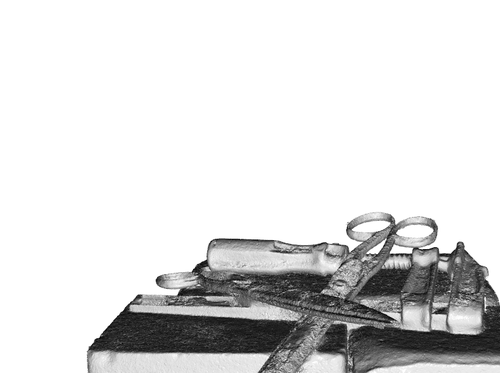} &
        \includegraphics[width=\quartlwrend]{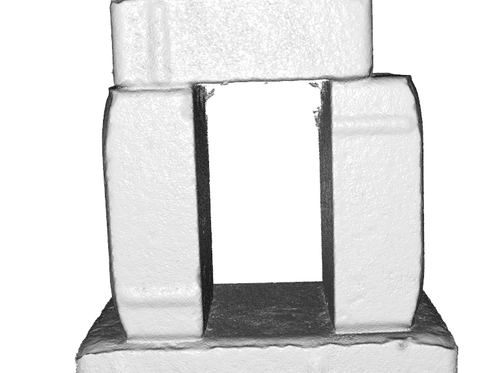} &
        \includegraphics[width=\quartlwrend]{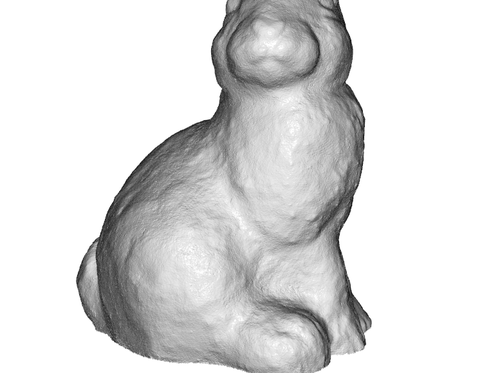} &
        \includegraphics[width=\quartlwrend]{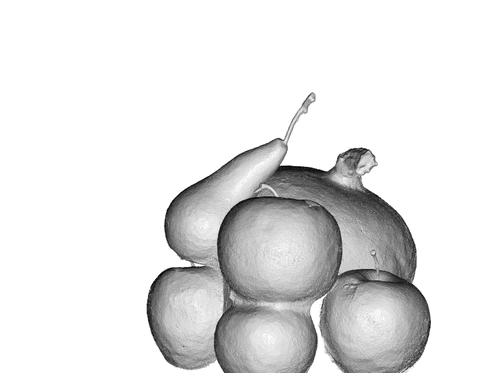} \\
        Scan24 & Scan37 & Scan40 & Scan55 & Scan63 \\
        \includegraphics[width=\quartlwrend]{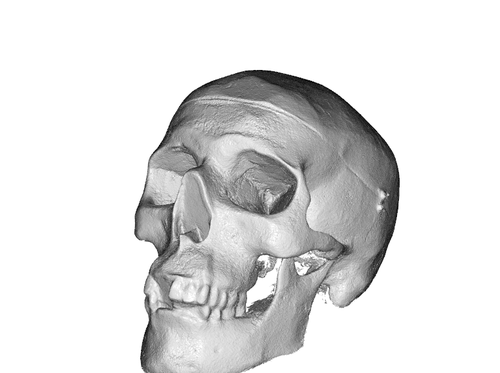} &
        \includegraphics[width=\quartlwrend]{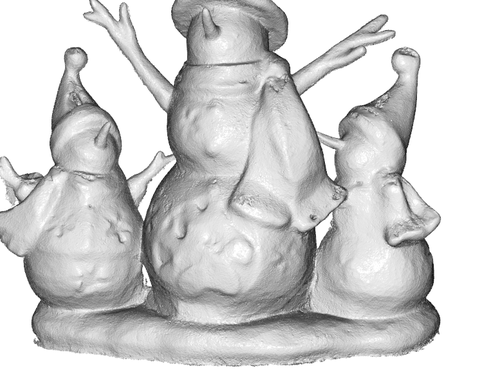} &
        \includegraphics[width=\quartlwrend]{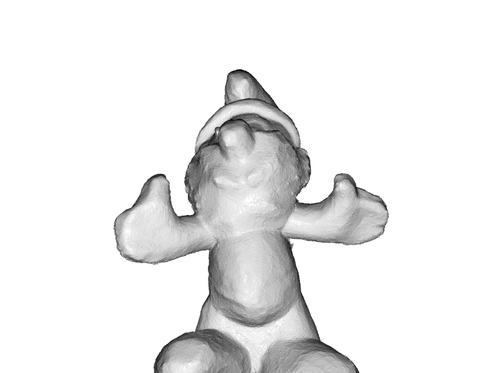} &
        \includegraphics[width=\quartlwrend]{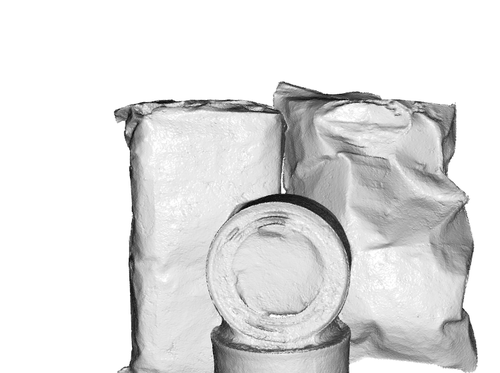} &
        \includegraphics[width=\quartlwrend]{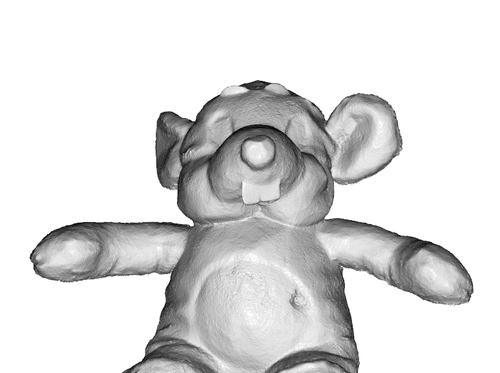} \\
        Scan65 & Scan69 & Scan83 & Scan97 & Scan105 \\
        \includegraphics[width=\quartlwrend]{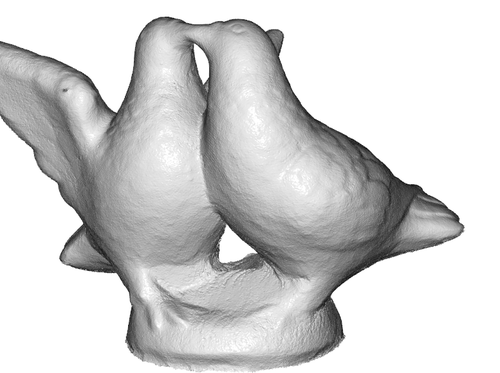} &
        \includegraphics[width=\quartlwrend]{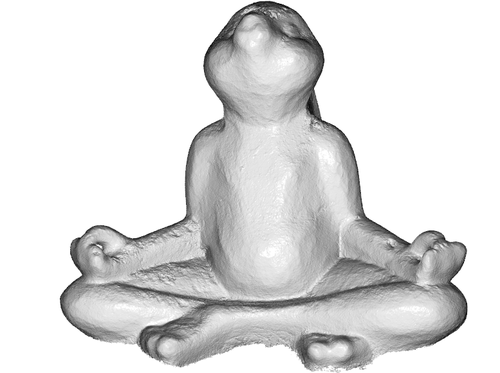} &
        \includegraphics[width=\quartlwrend]{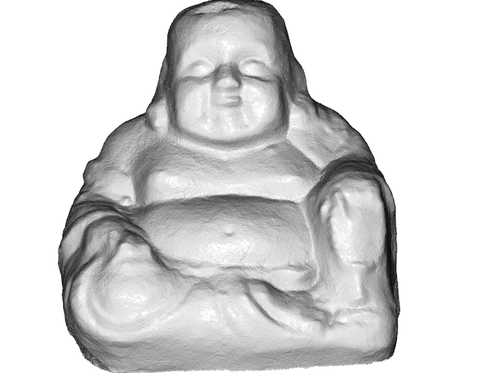} &
        \includegraphics[width=\quartlwrend]{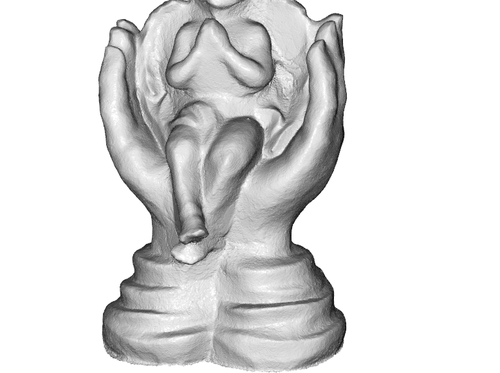} &
        \includegraphics[width=\quartlwrend]{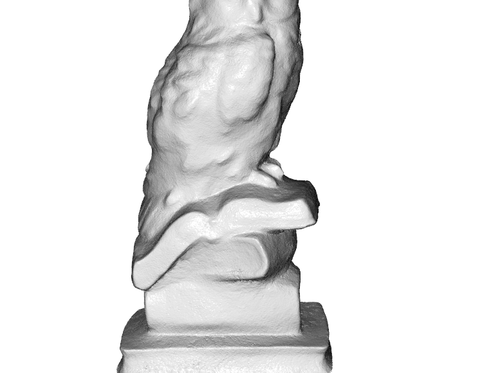} \\
        Scan106 & Scan110 & Scan114 & Scan118 & Scan122 \\
\end{tabular}
    \caption{Qualitative evaluation of surface reconstruction using TV-SGS\textsubscript{R} on the DTU dataset with \textbf{dense views}.}
    \label{fig_supp:DTU_all_mesh_dense}
\end{figure*}